\documentclass[a4paper,fleqn]{cas-dc}

\usepackage[numbers,sort&compress]{natbib}
\usepackage{algpseudocode}
\usepackage{algorithmicx,algorithm}
\usepackage{makecell}
\usepackage{float}
\usepackage{graphicx}
\usepackage[caption=false,font=footnotesize,labelfont=sf,textfont=sf]{subfig}   
\usepackage{subfig}
\usepackage{threeparttable}
\usepackage{stfloats}
\usepackage{caption}

\def\tsc#1{\csdef{#1}{\textsc{\lowercase{#1}}\xspace}}
\tsc{WGM}
\tsc{QE}

\begin{document}
\let\WriteBookmarks\relax
\def\floatpagepagefraction{1}
\def\textpagefraction{.001}

\shorttitle{Discovering PINNs model for Solving PDEs through Evolutionary Computation}    

\shortauthors{Bo Zhang et al.}  

\title [mode = title]{Discovering Physics-Informed Neural Networks Model for Solving Partial Differential Equations through Evolutionary Computation}

%\author[<aff no>]{<author name>}[<options>]
\author[1,2]{Bo\ Zhang} %[ORCID:0000-0003-3567-5748]

\ead{zhang_bo@pku.edu.cn}

\credit{Conceptualization, Investigation, Methodology, Software, Writing – original draft, Funding acquisition}

\affiliation[1]{organization={National Engineering Laboratory for Big Data Analysis and Applications, Peking University},
            city={Beijing},
            postcode={100871}, 
            country={China}}

\affiliation[2]{organization={School of Mathematical Sciences, Peking University}, 
            city={Beijing},
            postcode={100871}, 
            country={China}}

\affiliation[3]{organization={PKU-Changsha Institute for Computing and Digital Economy},
            city={Changsha},
            postcode={410205}, 
            country={China}}

%\author[<aff no>]{<author name>}[<options>]
\author[1,2,3]{Chao\ Yang}%[ORCID:0000-0001-7426-6248]

% Email id of the second author
\ead{chao_yang@pku.edu.cn}

%\cormark[<corr mark no>]
\cormark[1]

\cortext[1]{Corresponding author}

\credit{Conceptualization, Investigation, Methodology, Writing – review and editing, Funding acquisition}

\begin{abstract}
In recent years, the researches about solving partial differential equations (PDEs) based on artificial neural network have attracted considerable attention. In these researches, the neural network models are usually designed depend on human experience or trial and error. Despite the emergence of several model searching methods, these methods primarily concentrate on optimizing the hyperparameters of fully connected neural network model based on the framework of physics-informed neural networks (PINNs), and the corresponding search spaces are relatively restricted, thereby limiting the exploration of superior models. This article proposes an evolutionary computation method aimed at discovering the PINNs model with higher approximation accuracy and faster convergence rate. In addition to searching the numbers of layers and neurons per hidden layer, this method concurrently explores the optimal shortcut connections between the layers and the novel parametric activation functions expressed by the binary trees. In evolution, the strategy about dynamic population size and training epochs (DPSTE) is adopted, which significantly increases the number of models to be explored and facilitates the discovery of models with fast convergence rate. In experiments, the performance of different models that are searched through Bayesian optimization, random search and evolution is compared in solving Klein-Gordon, Burgers, and Lamé equations. The experimental results affirm that the models discovered by the proposed evolutionary computation method generally exhibit superior approximation accuracy and convergence rate, and these models also show commendable generalization performance with respect to the source term, initial and boundary conditions, equation coefficient and computational domain. The corresponding code is available at https://github.com/MathBon/Discover-PINNs-Model.

\end{abstract}

% Keywords

\maketitle

% Main text
\section{Introduction}

In recent years, the researches about solving partial differential equations (PDEs) utilizing artificial neural networks have surged dramatically. In these researches, the pivotal issue is how to effectively combine neural network with physical knowledge or data. Compared to the traditional numerical methods for solving PDEs such as finite element, finite difference, and finite volume, the neural network methods are usually mesh-free and have more powerful in dealing with the high-dimensional problems. Besides, the experimental or observational data can be seamlessly integrated into the loss function of neural network to improve the solution accuracy. With these characteristics, the burgeoning neural network methods have attracted much attention.

The mainstream researches about neural network methods for solving PDEs involve several aspects. In terms of the computational form of PDEs, the strong form is the most widely used \cite{lagaris1998artificial, raissi2019physics, lyu2022mim}, with physics-informed neural networks (PINNs) being a representative research \cite{raissi2019physics}. The variational form \cite{sheng2021pfnn, yu2018deep} and weak form \cite{sheng2022pfnn2, kharazmi2021hp, khodayi2020varnet} are also adopted in some studies, which reduce the regularity requirement about the solution space and decrease the computational cost of derivative. In solving PDEs, the initial and boundary conditions are provided for determining the solutions, which are indispensable. For making full use of the given conditions, different formulations of neural network surrogate solutions are delicately constructed so that the surrogate solutions automatically satisfy the initial and boundary conditions \cite{lagaris1998artificial, mcfall2009artificial, lyu2020enforcing}, especially in complex geometries \cite{berg2018unified, sheng2021pfnn, sheng2022pfnn2}. Additionally, the sampling strategies \cite{wight2021solving, nabian2021efficient, tang2023pinns}, loss terms balancing methods \cite{wang2021understanding, bischof2021multi, wang2022and}, domain decomposition approaches \cite{shukla2021parallel, jagtap2020extended, sheng2022pfnn2} are all the research hotspots. With the rapid development of neural network methods for solving PDEs, several software libraries \cite{lu2021deepxde, hennigh2021nvidia, peng2021idrlnet} are developed for promoting the corresponding applications.

In this paper, designing the PINNs model for solving PDEs is mainly concerned, which directly impacts the solution accuracy and convergence rate of training. In constructing the structure of PINNs model, in addition to the common fully connected neural network (FcNet), FcNet with residual blocks \cite{cheng2021deep} (FcResNet), gated neural network \cite{wang2021understanding}, convolutional neural network \cite{gao2021phygeonet}, recurrent neural network \cite{rodriguez2021physics}, and graph neural network \cite{jiang2023phygnnet} have been adopted with utilizing their own characteristics. Besides, activation function is the crucial component of the neural network, which influences the nonlinear representation ability of the neural network. Typical activation functions such as tanh and sin are widely used and several parametric activation functions are further studied \cite{jagtap2020adaptive, jagtap2020locally}. The corresponding experimental results demonstrate that the proper parametric activation function tailored to the given problem may improve the convergence rate as well as the approximation accuracy compared to the fixed activation function \cite{jagtap2020adaptive}.

Currently, the neural network models for solving PDEs are primarily designed depend on the human experience or trial and error. How to automatically design more favorable neural network is an interesting issue, and the related technique is known as neural architecture search (NAS) that belongs to automated machine learning (AutoML). In machine learning community, the search strategies in NAS mainly include random search \cite{li2020random, yu2019evaluating}, evolutionary computation \cite{real2019regularized, xie2017genetic}, reinforcement learning \cite{zoph2016neural, zhong2020blockqnn}, differentiable search \cite{liu2018darts, cai2018proxylessnas} and so on. The performance of many automatically designed neural networks approaches or even surpasses the hand-crafted ones. While, in most studies about NAS, the used activation functions are chosen from the existing ones. Therefore, discovering the novel effective activation functions had become an attractive topic. The early research about searching the activation functions is based on reinforcement learning \cite{ramachandran2017searching}, which trains a recurrent neural network (RNN) controller to generate each component of an activation function expressed by a binary tree. Subsequently, several evolutionary searching methods are proposed \cite{basirat2018quest, bingham2020evolutionary}, especially for discovering the parametric activation functions \cite{bingham2022discovering}. The above endeavors about automatic design of neural networks are mainly focus on the tasks about computer vision and natural language processing.

Recently, in the realm of solving PDEs, several works about automatically designing PINNs model to address the forward and inverse problems have emerged. The core of these works mainly comprises the search space, search strategy, and evaluation indicator. The search space is usually constructed by number of layers, number of neurons in each hidden layer and activation function. In search strategy, the grid search \cite{du2022autoke} is used due to its simpleness. However, its practicality diminishes as the number of sampling points in each search dimension increases. In view of computational efficiency, the Bayesian optimization \cite{guo2022stochastic, escapil2023hyper, bischof2021multi} is most widely adopted, which takes advantage of the information from the evaluated models to iteratively guide the sampling of new model for approximating the objective function so that it avoids evaluating a large number of models in the brute-force way. Furthermore, reinforcement learning based search strategy is also adopted \cite{wu2022autopinn}, which trains a RNN controller to generate the number of neurons and activation function at each layer. In order to compress the search space, the hyperparameters to be searched are decoupled \cite{wang2022auto} in searching. As for the evaluation indicator, in the forward problem, the relative $L_2$ error is taken as the evaluation indicator of the model \cite{guo2022stochastic}, and the minimum loss function value in the training process serves as the evaluation indicator when the exact solution is unknown \cite{wang2022auto}. In inverse problem about parameter identification of PDEs, mean absolute error (MAE) is introduced into evaluation indicator, and the model size is also considered so as to comply the given hardware constraints \cite{wu2022autopinn}.

In the current researches on automatically designing PINNs model to solve PDEs, the model structure primarily revolves around FcNet, with activation function usually searched from the commonly used options. In this paper, apart from searching the numbers of layers and neurons per hidden layer, we mainly focus on discovering the optimal shortcut connections of the structure as well as the novel activation function for the given problem, aiming at improving the approximation accuracy and convergence rate of PINNs model. The main contributions of this paper are summarized as follows:

(1) An evolutionary computation method about discovering superior model of PINNs to solve PDEs is proposed, which concurrently searches the structure and activation function of the model. 

(2) In addition to searching the numbers of layers and neurons per hidden layer, the effective shortcut connections between the layers and the novel parametric activation functions expressed by the binary trees are also explored. 

(3) The strategy about dynamic population size and training epochs (DPSTE) is adopted in evolution, which significantly increases the number of searching models and facilitates the discovery of models with fast convergence rate.

The remainder of this paper is organized as follows. In Section 2, preliminaries about PINNs and sensitivity analysis about search objectives are introduced. The evolutionary computation method is proposed in Section 3. The experiments are presented in Section 4. Eventually, the conclusion and outlook are given in Section 5.  

\section{Preliminaries}\label{}
\subsection{Physics-informed neural networks}
At present, the most famous neural network method for solving PDEs is physics-informed neural networks (PINNs)\cite{raissi2019physics}. And it has been expanded for solving integro-differential equations, fractional differential equations \cite{pang2019fpinns}, and stochastic differential equations \cite{zhang2019quantifying}. Consider a general form of PDE as
\begin{equation}
\label{eq:PDE}
\begin{aligned}
& \mathcal{N}[u(\boldsymbol{x}, t)]=0, && \quad \boldsymbol{x} \in \Omega, t \in (0, T], \\
& \mathcal{B}[u(\boldsymbol{x}, t)]=0, && \quad \boldsymbol{x} \in \partial \Omega, t \in [0, T], \\
& \mathcal{I}[u(\boldsymbol{x}, t)]=0, && \quad \boldsymbol{x} \in \overline\Omega, t = 0,
\end{aligned}
\end{equation}
where $\mathcal{N}[.]$ is a differential operator defined on $ \Omega \times(0, T] \subset \mathbb{R}^d \times \mathbb{R}_{+}$, $\mathcal{B}[.]$ and $\mathcal{I}[.]$ are the differential operators corresponding to the boundary and initial conditions, and $(\boldsymbol{x}, t)$ denotes spatio-temporal coordinates. In vanilla PINNs, a neural network $ \hat u(\boldsymbol{x}, t; \boldsymbol{\theta})$ is used to approximate the solution $ u(\boldsymbol{x}, t)$ of problem \eqref{eq:PDE}. 
The optimal network parameters $\boldsymbol{\theta}^*$ can be determined by minimizing the loss function $J(\boldsymbol{\theta})$ that is constructed from the residuals as
\begin{equation}
\begin{aligned} 
\label{eq:residual}
& J(\boldsymbol{\theta}) =  J_r(\boldsymbol{\theta})+\eta_b J_b(\boldsymbol{\theta}) + \eta_0 J_0(\boldsymbol{\theta}) , \\ 
& J_r(\boldsymbol{\theta}) =\int_{\Omega \times (0, T]}(\mathcal{N}[\hat u(\boldsymbol{x}, t; \boldsymbol{\theta})])^2  d\boldsymbol{x}dt, \\ 
& J_b(\boldsymbol{\theta}) =\int_{\partial \Omega \times (0, T]}(\mathcal{B}[\hat u(\boldsymbol{x}, t; \boldsymbol{\theta})])^2 d\boldsymbol{x}dt, \\
& J_0(\boldsymbol{\theta}) =\int_{\overline\Omega}(\mathcal{I}[\hat u(\boldsymbol{x}, 0; \boldsymbol{\theta})])^2 d\boldsymbol{x},  
\end{aligned}
\end{equation}
and $\boldsymbol{\theta}^*$ is expressed by
\begin{equation}
\begin{aligned} 
\boldsymbol{\theta}^*=\underset{\boldsymbol{\theta}}{\operatorname{argmin}} \ J(\boldsymbol{\theta}).
\end{aligned}
\end{equation}
In this expression, the problem of solving PDEs with initial and boundary constrains is transformed to an unconstrained optimization problem. In numerical calculating, $J(\boldsymbol{\theta})$ in Eq. \eqref{eq:residual} can be estimated by Monte Carlo approximation, and the computable loss function $\mathcal{L}(\boldsymbol{\theta}) \approx \frac{J(\boldsymbol{\theta})}{\left| \Omega \right| T } $ is defined by
\begin{equation}
\begin{aligned}
&\mathcal{L}(\boldsymbol{\theta}) =\mathcal{L}_r(\boldsymbol{\theta})+\lambda_b\mathcal{L}_b(\boldsymbol{\theta})+\lambda_0\mathcal{L}_0(\boldsymbol{\theta}), \\
&\mathcal{L}_r(\boldsymbol{\theta})  =\frac{1}{N_r} \sum_{i=1}^{N_r} |\mathcal{N}[\hat{u}(\boldsymbol{x}_r^i, t_r^i; \boldsymbol{\theta})]|^2, \\
&\mathcal{L}_b(\boldsymbol{\theta})  =\frac{1}{N_b} \sum_{i=1}^{N_b} |\mathcal{B}[\hat{u}(\boldsymbol{x}_b^i, t_b^i; \boldsymbol{\theta})]|^2, \\
&\mathcal{L}_0(\boldsymbol{\theta})  =\frac{1}{N_0} \sum_{i=1}^{N_0} |\mathcal{I}[\hat{u}(\boldsymbol{x}_0^i, 0; \boldsymbol{\theta})]|^2,
\end{aligned}
\end{equation}
where $\left\{\boldsymbol{x}_r^i, t_r^i\right\}_{i=1}^{N_r}$ denote the collocation points inside the computational domain, $\left\{\boldsymbol{x}_b^i, t_b^i\right\}_{i=1}^{N_b}$ and $\left\{\boldsymbol{x}_0^i, 0\right\}_{i=1}^{N_0}$ are the boundary and initial points, $\lambda_b= \frac{\left| \partial \Omega \right| \eta_b}{\left| \Omega \right|} $ and $\lambda_0 = \frac{\eta_0}{T} $ are the penalty coefficients for boundary loss term $\mathcal{L}_b(\boldsymbol{\theta})$ and initial loss term $\mathcal{L}_0(\boldsymbol{\theta})$. The partial derivatives about the output of neural network with respect to the spatial and temporal variables can be computed through automatic differentiation.

\subsection{Sensitivity analysis}  
\label{sec:sensitivity}
\begin{figure*}[!htb]
\centering
\subfloat[Case about solving Klein-Gordon equation]{\includegraphics[width=6.8in]{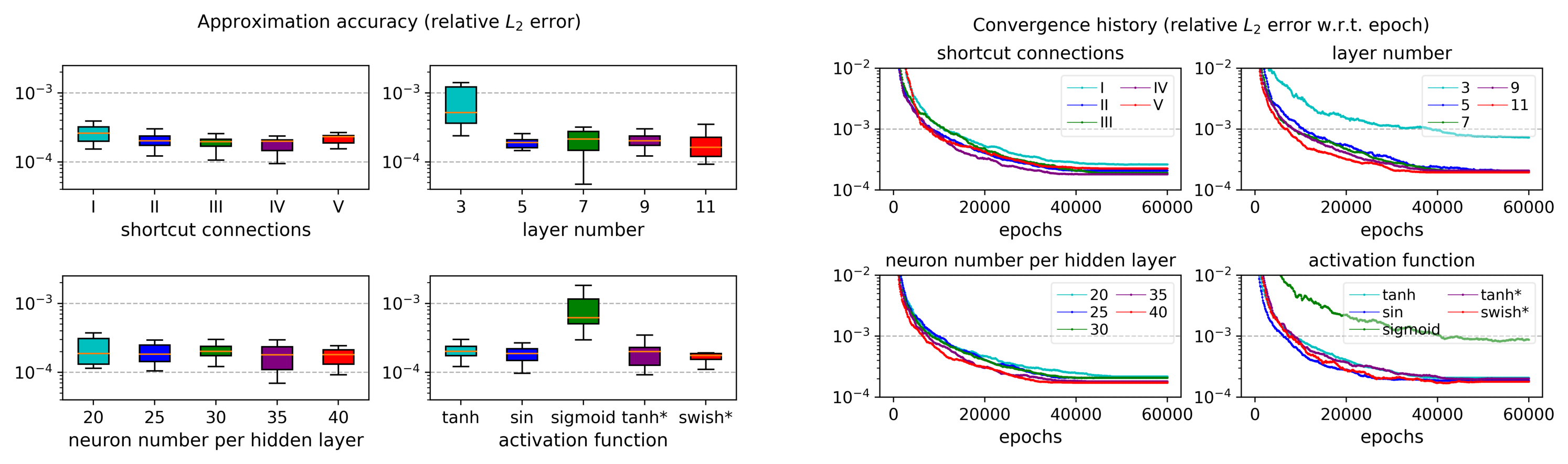} \label{fig:Klein-Gordon-sen}}  \\
\vspace{0mm}
\subfloat[Case about solving Burgers equation]{\includegraphics[width=6.8in]{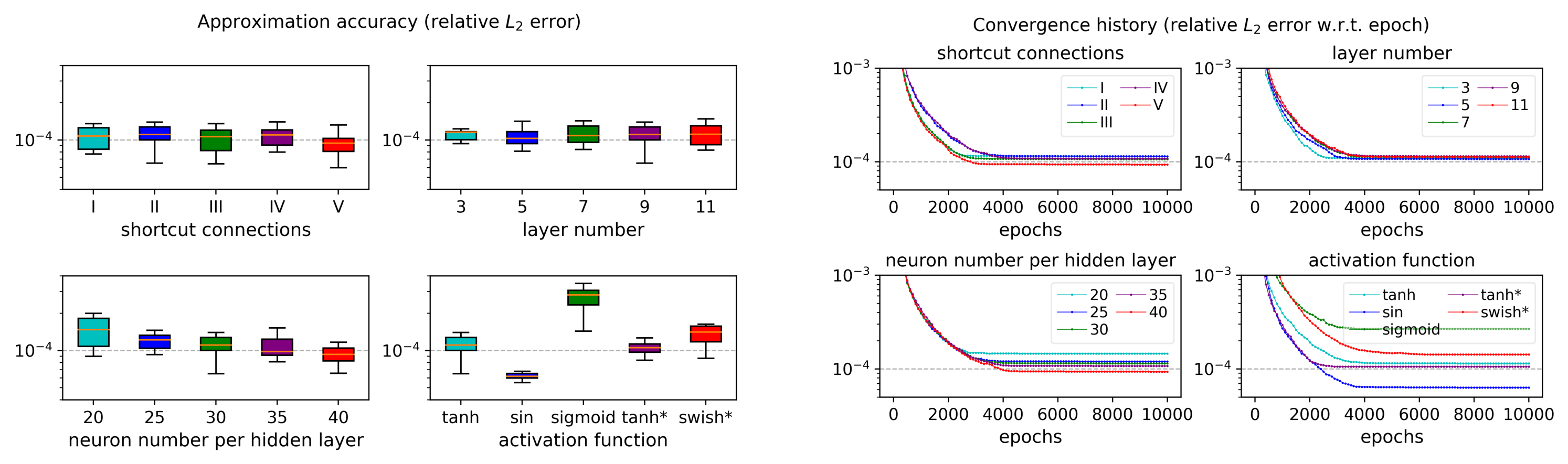} \label{fig:Burgers-sen}}  \\
\vspace{0mm}
\subfloat[Case about solving Lamé equations]{\includegraphics[width=6.8in]{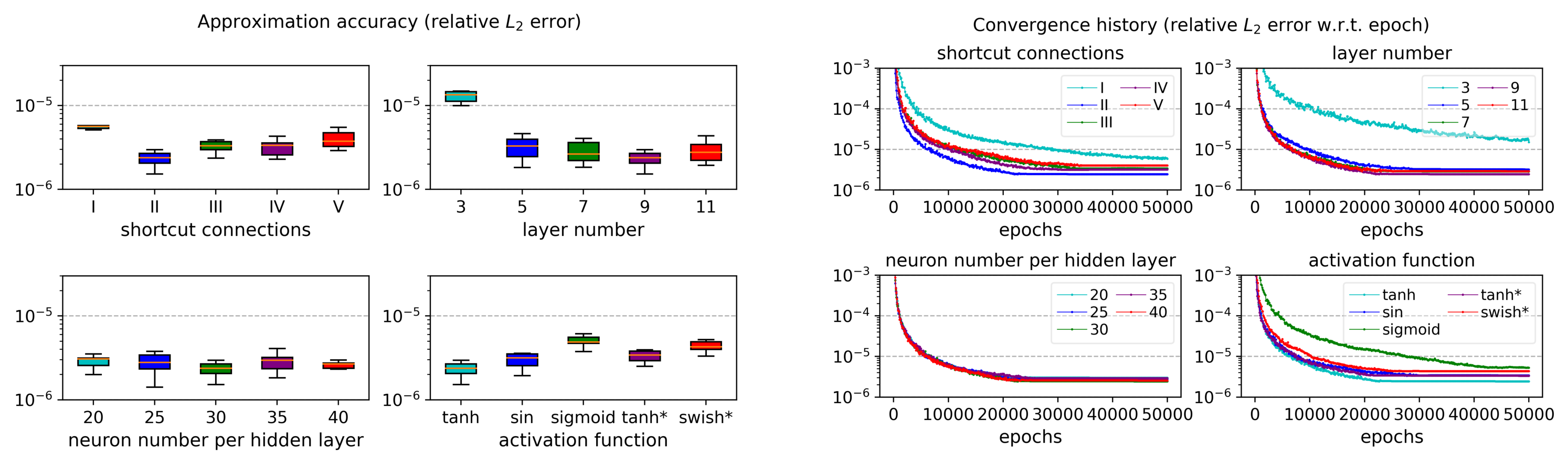} \label{fig:Lame-sen}}
\caption{Approximation accuracy and convergence history of relative $L_2$ error about different models in solving (a) Klein-Gordon, (b) Burgers, and (c) Lamé equations. Each curve of convergence history is the average result of ten independent trials. The error in solving Lamé equations is about $u$. The five ways of shortcut connections are labeled as: I:none, II:[0-2,2-4,4-6,6-8], III:[0-1,1-3,4-6,7-8], IV:[1-4,4-6,6-8], V:[0-3,4-7], where `0-2' indicates the connection between the inputs of first layer and third layer, and so forth. `tanh*' and `swish*' denote layer-wise parametric activation functions tanh($\alpha \cdot x$) and $x$ $\cdot $ sigmoid($\alpha \cdot x$).} 
\vspace{-3mm}
\label{fig:sensitivity}
\end{figure*}

For constructing a reasonable search space, the sensitivity of the performance of model to the search objectives is investigated. Besides the number of layers, number of neurons in each hidden layer and activation functions, the shortcut connections \cite{he2016deep} between the layers is also considered as the search objective. In this section, the sensitivities of the approximation accuracy and convergence rate of model to above four search objectives are analyzed by experiment. In analysis, the Klein-Gordon, Burgers, and Lamé equations are introduced and solved based on PINNs, and their specific expressions are exhibited in Section~\ref{sec:Experiments}. Initially, a baseline model is established, which has 9 layers with regular shortcut connections (each shortcut spans two fully connected layes), 30 neurons per hidden layer, and tanh activation function. And then each objective in this baseline model is independently changed for generating different models so as to investigate its sensitivity. The approximation accuracy and convergence history of relative $L_2$ error of each model are shown in Fig.~\ref{fig:sensitivity}. Ten times of independent trials are conducted about each model.

From Fig.~\ref{fig:sensitivity}, it is evident that these four search objectives affect the approximation accuracy or convergence rate to varying degrees, and their sensitivities vary across different equations. For example, the convergence rates are close over the different ways of shortcut connections in the case about Klein-Gordon equation (Fig.~\ref{fig:Klein-Gordon-sen}), but the discrepancy of convergence rates about the same objective is obviously larger in the case about Lamé equations (Fig.~\ref{fig:Lame-sen}). Moreover, the best choice for a objective may vary depending on different problems. For instance, the activation function sin exhibits higher accuracy compared to the other ones in solving Burgers equation (Fig.~\ref{fig:Burgers-sen}), however, it shows lower accuracy than tanh in solving Lamé equations (Fig.~\ref{fig:Lame-sen}). Therefore, the problem-oriented model searching is necessary. More detailed sensitivity analysis can be carried out by Morris method, extended Fourier amplitude sensitivity test (eFAST) method, etc. \cite{guo2022analysis}, which goes beyond the main scope of the present work. Sensitivity analysis can be used to exclude the search objectives in the search space, which have little impact on solving the given problem. Nevertheless, the sensitivity analysis is usually time-consuming, and in order to develop a general end-to-end model searching method across various problems, four objectives are all taken into account in the search space in this article.

\section{Evolutionary computation}
An evolutionary computation method for discovering the model of PINNs is proposed. And the search space, evolutionary operations, and key strategies in evolution are introduced in this section.

\subsection{Search space}
\begin{figure}[h]
\centering
\subfloat[Structure with shortcuts]{\includegraphics[width=1.60in]{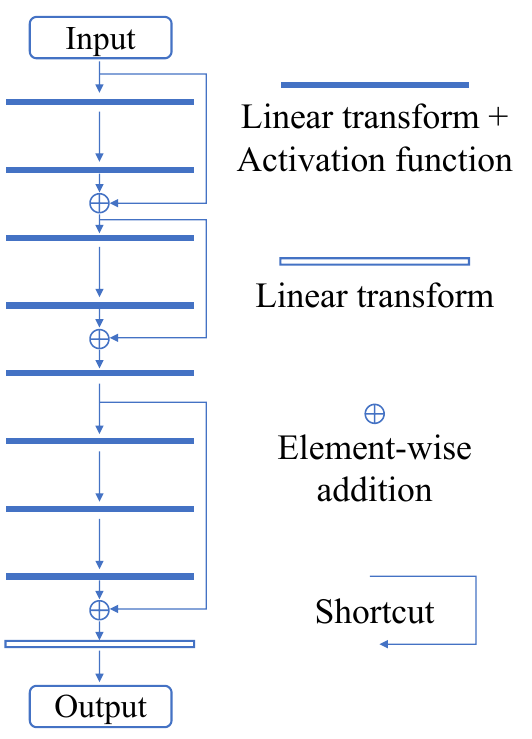}%
\label{fig:structure}}
\hspace{0.0mm}
\subfloat[Parametric activation function]{\includegraphics[width=1.65in]{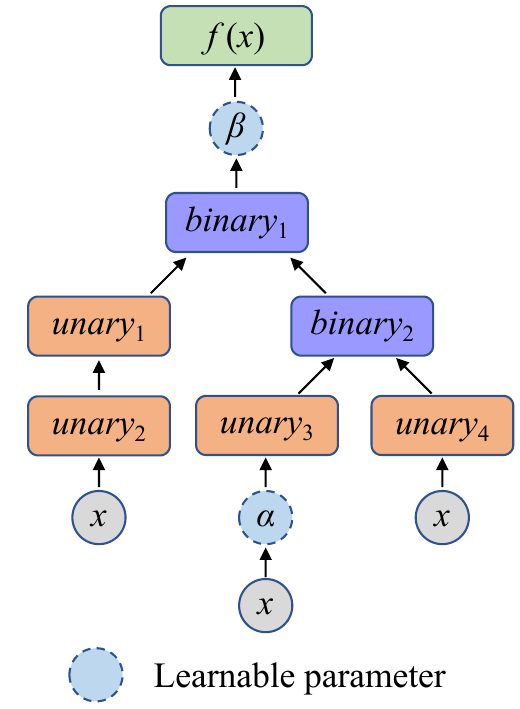}%
\label{fig:activation}}
\caption{The examples about the general forms of structure and activation function. The structure in (a) has 9 layers with shortcut connections: [0-2,2-4,5-8]. The unary and binary operators adopted in the activation function are listed in Table \ref{tab:operators}.} 
\label{fig:str_act}
\end{figure}

The search space of evolution is defined by potential combinations of the model’s structure and activation function. The general forms of the structure and activation function are illustrated in Fig.~\ref{fig:str_act}. The shortcut connections in the structure skip one or more layers without intersecting each other (Fig.~\ref{fig:structure}). The last layer is not contained in any shortcut connection. The computation graph of an activation function is expressed by a binary tree that contains several unary and binary operator nodes, and the layer-wise learnable parameters in the activation function are located on the edges of the binary tree as indicated in Fig.~\ref{fig:activation}. Because the structure and activation function are searched independently, the total number of the possible combinations ${C_{mod}}$ about a model is given as

\begin{equation}
\begin{aligned}
& {C_{mod}} = C_{str} \cdot C_{act},  \\
\end{aligned}
\end{equation}
where $C_{str}$ and $C_{act}$ are the combinatorial numbers of structure and activation function. $C_{str}$ can be calculated by
\begin{equation}
\begin{aligned}
& C_{str} = N_{neu} \cdot \sum\limits_{n}  Fib(2n-1),  \\
& Fib(n)= \frac{1}{\sqrt{5}} \left[\left(\frac{1+\sqrt{5}}{2}\right)^n-\left(\frac{1-\sqrt{5}}{2}\right)^n\right], \\
\end{aligned}
\end{equation}
where $N_{neu}$ is the number of alternative neuron number per hidden layer, $n$ is the number of layers, and $Fib(n)$ is the Fibonacci sequence. $C_{act}$ can be calculated by
\begin{equation}
\begin{aligned}
& C_{act} = \sum\limits_{m=1}^{M_N}  \sum\limits_{{b_n}=0}^{\lfloor(m-1)/2\rfloor}  \left( c({u_n},{b_n}) \cdot U^{u_n} \cdot B^{b_n} \cdot \sum\limits_{i=0}^{K_E}\begin{pmatrix}e_n \\ i \end{pmatrix} \right),\\ 
& u_{n} = m - b_{n}, \\
& e_{n} = 2b_{n} + u_{n} + 1, \\
& K_E = \mathrm{min}(e_{n}, M_E), 
\label{eq:C_act}
\end{aligned}
\end{equation}
where $m$ is the number of operator nodes in a binary tree, $M_N$ is the preset maximum number of operator nodes of a binary tree, $U$ and $B$ are the numbers of alternative unary and binary operators in the operator library (Table \ref{tab:operators}), $u_n$, $b_n$, and $e_n$ are the numbers of unary and binary operator nodes and edges in a binary tree, and $M_E$ is the preset maximum number of learnable parameters of an activation function. $c({u_n},{b_n})$ in Eq. \ref{eq:C_act} is the combinatorial number of a binary tree that has ${u_n}$ unary and ${b_n}$ binary operator nodes with not distinguishing the specific operations in unary and binary operator nodes, which is calculated by
\begin{equation}
\begin{aligned}
&c({u_n},{b_n}) = Cat(b_n) \cdot s({u_n},{b_n}) ,\\ 
&Cat(b_n) = \frac{1}{b_n+1} \cdot \begin{pmatrix}{2b_n} \\ b_n \end{pmatrix} ,\\ 
&s({u_n},{b_n}) = \begin{pmatrix}{u_n + b_n - 1} \\ 2b_n \end{pmatrix} ,\\ 
\end{aligned}
\end{equation}
where $Cat(b_n)$ is the Catalan number which is the combinatorial number of a strict binary tree with ${b_n}$ binary and ${b_n+1}$ unary operator nodes, and $s({u_n},{b_n})$ is the combinatorial number about inserting $u_n - (b_n + 1)$ unary operator nodes onto the $2b_n+1$ edges of the strict binary tree, where one edge can be inserted by more than one unary operator node. What needs clarification is that inserting the unary operator nodes on the input or output edge of an unary operator node of the strict binary tree is equivalent in calculating the combinatorial number, so the number of inserted edges is $2b_n+1$ rather than $3b_n+2$. The size of search space based on the used operator library ($U$=23, $B$=6) and $M_E=3$ is listed in Table \ref{tab:search_space}. Equivalent activation functions based on different expressions are contained in the search space. As a comparison, the size of search space about model in the existing searching methods \cite{bischof2021multi, wang2022auto, escapil2023hyper, du2022autoke, wu2022autopinn} is generally determined by the product of numbers of alternative neuron number per hidden layer, alternative layer number, and alternative activation functions. The order of magnitude about the size of search space without considering the hyperparameters such as learning rate in these searching methods is about $10^3$ or $10^4$, which is much smaller than the proposed evolutionary computation method. The large search space forms the foundation for discovering the superior models.

\begin{table}[!htb]
\renewcommand\arraystretch{1.0}
\caption{The alternative operators in activation function.}
    \centering
    \begin{footnotesize}
    \begin{tabular}{cc}
    \toprule    % & Operators   \\
    %\midrule 
    Unary  & \makecell[c]{$x$, $-x$, $x^{-1}$, $x^{2}$, $e^{x}$, $e^{x}+1$, $e^{-x}+1$, $e^{x}-1$, \\ $e^{x}+e^{-x}$, $e^{x}-e^{-x}$, sin($x$), sinh($x$), asinh($x$),  \\ cos($x$), cosh($x$), tanh($x$), atan($x$), erf($x$), erfc($x$), \\ sigmoid($x$), softsign($x$), fixed swish($x$), softplus($x$)  }  \\
                          
    \midrule %\hline 
    Binary  & \makecell[c]{$x_1+x_2$,  $x_1-x_2$,  $x_1 \cdot x_2$,  $x_1 / x_2$, \\ max($x_1, x_2$), min($x_1, x_2$)} \\
    \bottomrule
    \end{tabular}
    \end{footnotesize} %
    \label{tab:operators}
\end{table}

\begin{table}[H]
\renewcommand\arraystretch{1.0}
\setlength\tabcolsep{2pt}
\caption{The size of search space.}
    \centering
    \begin{footnotesize}
    \begin{threeparttable}
    \begin{tabular}{p{1.6cm}<{\centering}|p{4.0cm}<{\centering}|p{2.3cm}<{\centering}} 
    \toprule                      Structure & Activation function & Model  \\
    \midrule %\hline 
    \multirow{4}{*}[-2.2ex]{2.83e05\tnote{a}}     &    2.65e05 (m=3) &   7.51e10 \\
                           \cmidrule{2-3} %\hline 
                              &   9.97e08 (m=5) &   2.82e14  \\
                           \cmidrule{2-3} %\hline 
                             &   3.34e12 (m=7) &   9.47e17  \\
                           \cmidrule{2-3} %\hline 
                             &   3.40e12 (1 $\leqslant$ m $\leqslant$ 7) &    9.64e17   \\

    \bottomrule
    \end{tabular}

    \begin{tablenotes}
       \footnotesize
       \item[a] The number of layers $n$ is searched from 3 to 11, and $N_{neu}$ = 16.
     \end{tablenotes}
    \end{threeparttable}

    \end{footnotesize} %
    \label{tab:search_space}
\end{table}

\subsection{Evolutionary operations}

\subsubsection{Genetic encoding}
\begin{figure*}[!htb]
\centering
\includegraphics[width=6.5in]{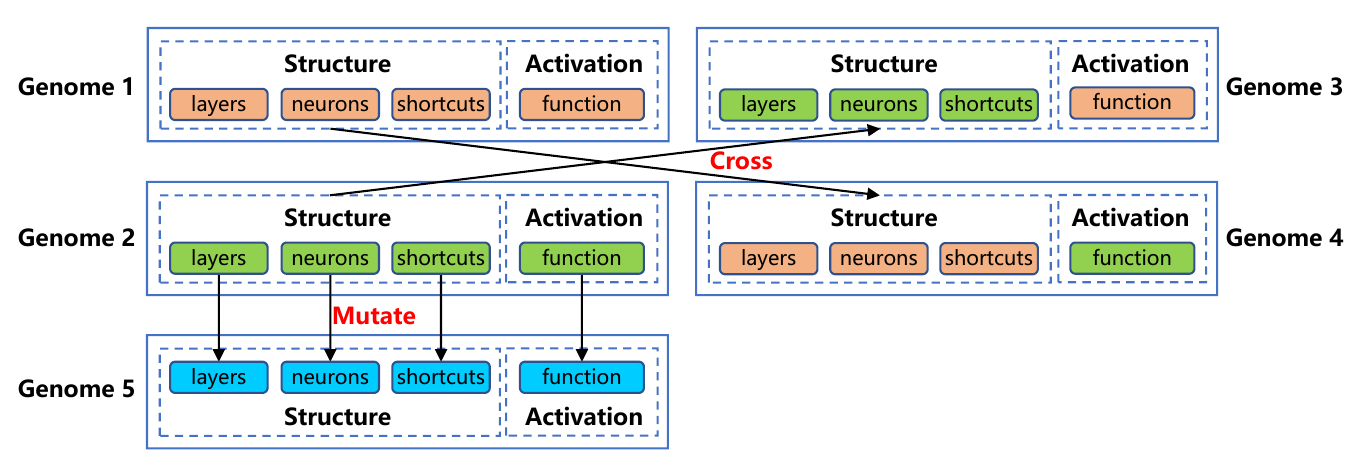}
\caption{Genetic encoding of the neural network model and the corresponding crossover and mutation operations.}
\label{fig:genome}
\end{figure*}

For simultaneously searching the structure and activation function, the genome of a model is encoded as illustrated in Fig.~\ref{fig:genome}, which is composed of the structure gene and activation function gene. In structure gene, three sub-genes that respectively represent the number of layers, number of neurons per hidden layer, and shortcut connections are contained. This way of genetic encoding is concise and it can be effectively combined with the crossover and mutation in evolution as shown in Fig.~\ref{fig:genome}. 

\subsubsection{Initialization}
In order to ensure the diversity of the initial population, which is conducive to avoiding the premature convergence of evolution and increasing the probability of emergence of the outstanding individuals (i.e. models), various initial individuals are constructed. With regard to the model structure, the initial population contains FcNet and FcResNet with regular and random shortcut connections, and the number of layers as well as the number of neurons per hidden layer of each initial individual are randomly set within predetermined ranges. As for activation function, the commonly used ones such as tanh, sin, sigmoid, parametric swish, parametric tanh, parametric sin, etc. and random activation functions are randomly assigned to each initial individual. The random activation functions adopt two forms \cite{bingham2022discovering} as
\begin{equation}
\begin{aligned} 
&f(x)=unary_1(unary_2(x)) 
\end{aligned} 
\end{equation}
and 
\begin{equation}
\begin{aligned} 
&f(x)=binary(unary_1(x), unary_2(x)).
\end{aligned} 
\end{equation}
And the random activation function is parameterized by randomly assigning the learnable parameters to the edges of the corresponding binary tree. The details about initialization are supplemented in Appendix~\ref{sec:Appendix_initial}.

\subsubsection{Selection}
In evolution, linear ranking selection \cite{blickle1996comparison} is adopted based on the fitness that is defined by the negtive of minimum loss function value in the training process \cite{wang2022auto}. Linear ranking selection is appropriate for dealing with various problems with different levels of degree about fitness discrepancy. Compared to fitness proportional selection, linear ranking selection amplifies the differentiation between the individuals when their fitnesses are close, and meanwhile, it increases the chances of low-fitness individuals being selected when there is a large disparity in fitness, benefiting the preservation of genetic diversity during evolution. In linear ranking selection, the rank $S$ is assigned to the best individual and the rank 1 is assigned to the worst one. The selection probability of the individual with rank $i$ is
\begin{equation}
\begin{aligned}
\label{eq:linear}
p_i=\frac{1}{S}\left(\eta^{-}+\left(\eta^{+}-\eta^{-}\right) \frac{i-1}{S-1}\right), \quad i \in\{1, \ldots, S\},
\end{aligned}
\end{equation}
where $S$ is the number of individuals in a generation, and $\eta^{-}/S$ and $\eta^{+}/S$ are the selection probabilities of the worst and best individuals, and $\eta^{-}$ and $\eta^{+}$ are set as $2/(1+S)$ and $2S/(1+S)$ respectively.

\subsubsection{Crossover and Mutation}
Based on the selected parents, their children are generated through crossover or mutation, and the ratio of crossover to the sum of mutation and crossover is ${R}_{cm}$. In crossover operation, the structure genes and activation function genes from both parents are exchanged with crossover rate ${R}_{c}$, following a single-point crossover mechanism illustrated in Fig.~\ref{fig:genome}. In mutation operation, three sub-genes representing the number of layers, number of neurons per hidden layer, and shortcut connections in the structure gene, along with the activation function gene, perform independent mutations with mutation rates ${R}_{l}$, ${R}_{n}$, ${R}_{s}$ and ${R}_{a}$ respectively.

\begin{figure*}[!htb]
\centering
\includegraphics[width=6.8in]{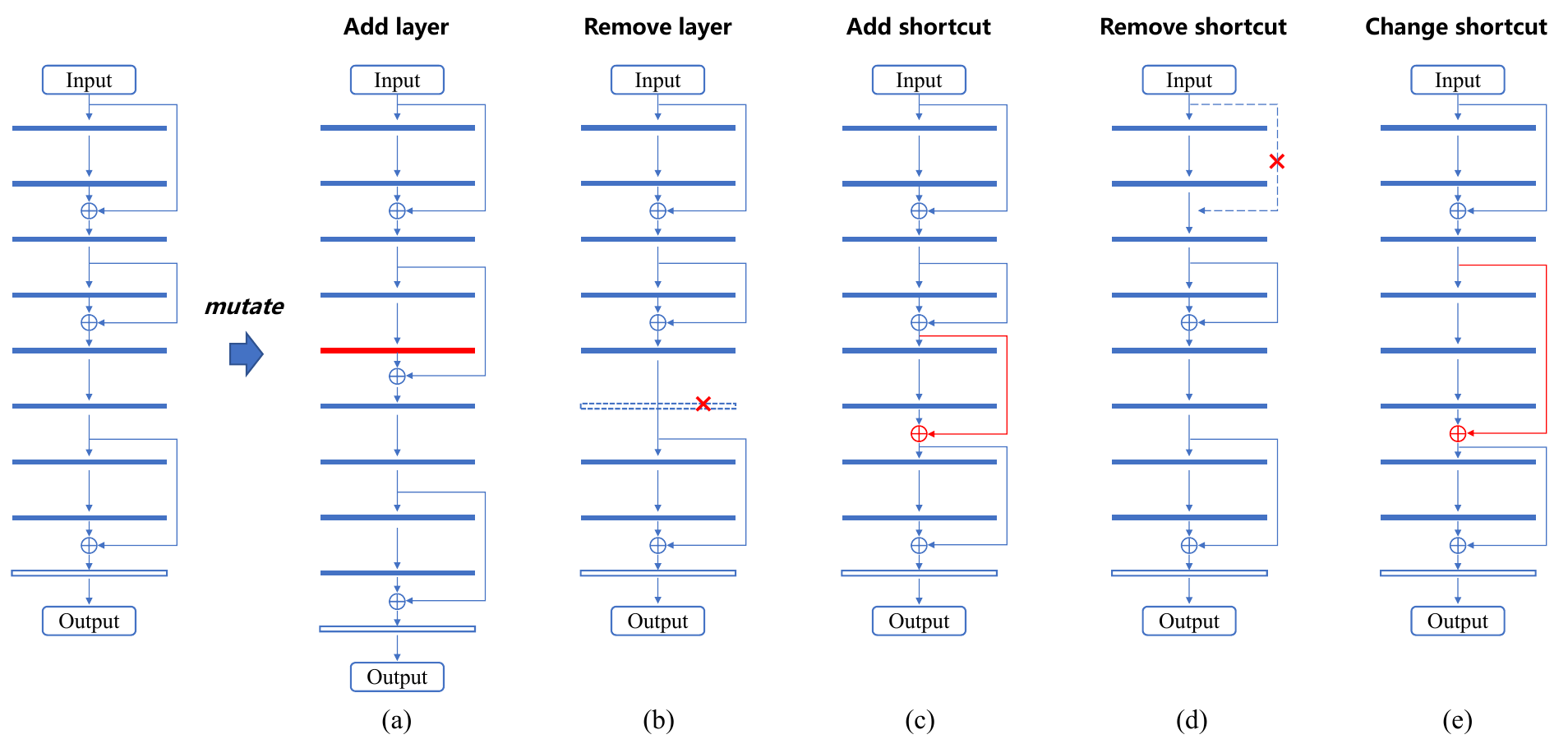}
\caption{Mutation about the structure of neural network model.}
\label{fig:struct_mutate}
\end{figure*}

The mutation operations of structure are illustrated in Fig.~\ref{fig:struct_mutate}. In mutating the number of layers, a new layer will be inserted into the structure (Fig.~\ref{fig:struct_mutate}(a)) or an existing layer will be randomly removed (Fig.~\ref{fig:struct_mutate}(b)). In the mutation of shortcut connections, there are three mutation types including `add shortcut', `remove shortcut' and `change shortcut'. In an `add shortcut' mutation (Fig.~\ref{fig:struct_mutate}(c)), a new shortcut will be added into the structure, ensuring it does not intersect with any other existing shortcut. In a `remove shortcut' mutation (Fig.~\ref{fig:struct_mutate}(d)), an existing shortcut will be removed at random. In a `change shortcut' mutation (Fig.~\ref{fig:struct_mutate}(e)), the position of start layer or end layer of an existing shortcut will be changed. When performing a mutation about the shortcut connection, each layer is guaranteed to be included in at most one shortcut connection. Besides, in mutating the number of neurons per hidden layer, the adjacent alternative neuron number is chosen as the new one.

\begin{figure*}[!htb]
\centering
\includegraphics[width=6.8in]{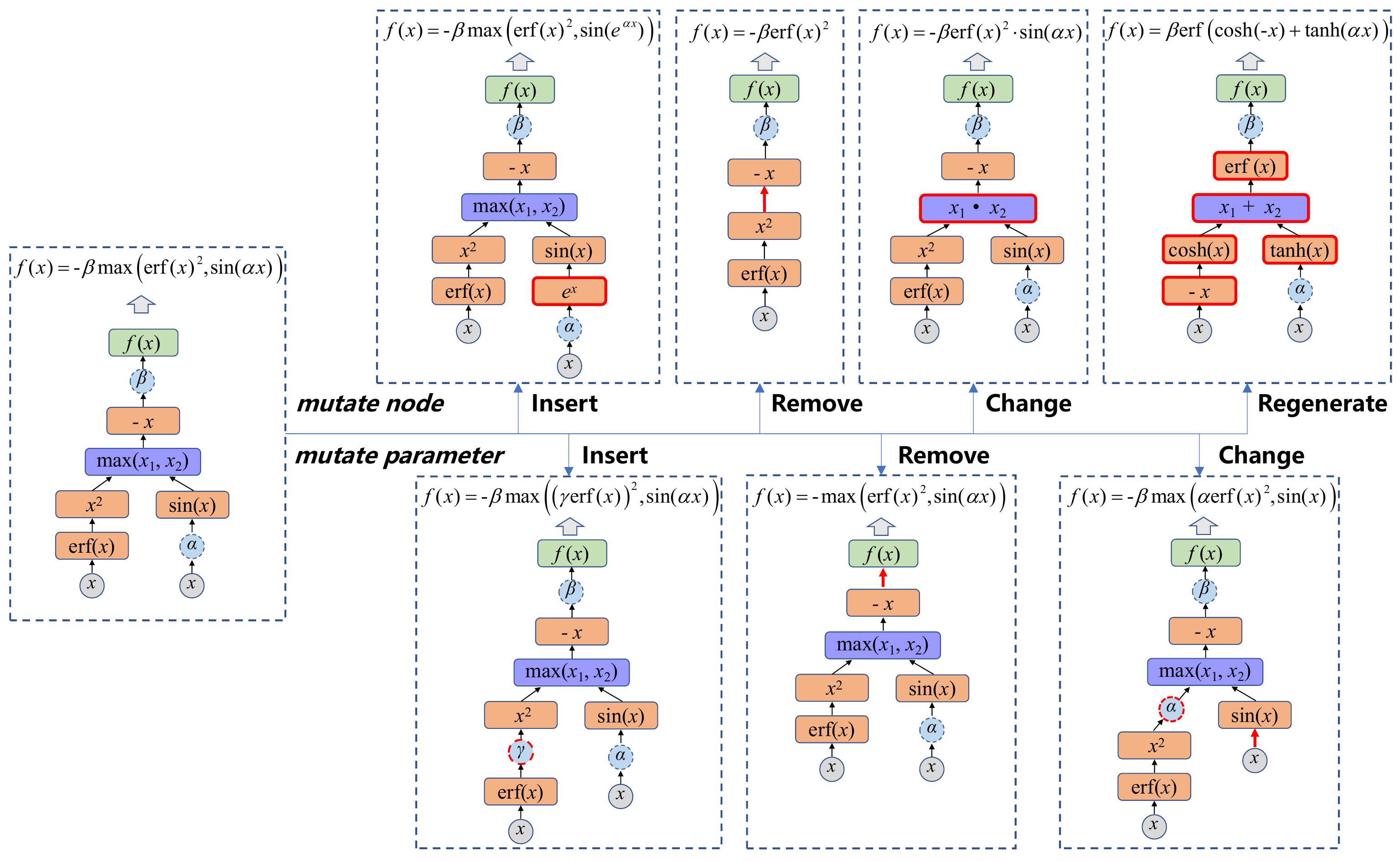}
\caption{Mutation about activation function of neural network model.}
\label{fig:act_mutate}
\end{figure*}

The mutation of activation function mainly draws on \cite{bingham2022discovering}. The main improvement in this work is that the learnable parameters in an activation function can be inherited from its parent in mutation, while in \cite{bingham2022discovering}, the learnable parameters are randomly reassigned to the edges after each mutation of activation function, potentially leading to the loss of beneficial features from its parent. In our implementation, two types of mutation about operator node and learnable parameter are adopted, as shown in Fig.~\ref{fig:act_mutate}.

The mutations about operator node contain `insert node', `remove node', `change node', and `regenerate nodes'. In an `insert node' mutation, an operator is randomly selected from the operator library and inserted on a random edge of the binary tree. If the inserting operator is binary, one of its inputs is set to the subtree below the inserted edge, and another input is set to a random unary operator that takes $x$ as input. In \cite{bingham2022discovering}, the insertion of binary operator node only changes the computation graph but not the computing result, while, in our implementation, the computation graph and computing result are both changed, which is beneficial to improve the evolutionary efficiency. In a `remove node' mutation, a random operator node is removed. If the removed operator node is binary, one of its inputs is kept and connected to its output node and another input is deleted. In a `change node' mutation, the operator of a node is randomly replaced with another operator of the same type. In a `regenerate nodes' mutation, the operator of each node in the binary tree is changed.

Regarding the mutations about learnable parameter, there are three types of `insert parameter', `remove parameter', and `change parameter' as shown in Fig.~\ref{fig:act_mutate}. In an `insert parameter' mutation, a new learnable parameter is added on a random edge where no parameter exists. In a `remove parameter' mutation, an existing learnable parameter is removed at random. In a `change parameter' mutation, the position of a learnable parameter is changed to another random edge where no parameter exists.

\subsection{Key strategies}
\subsubsection{Dynamic population size and training epochs}
For discovering the outstanding individuals, it is beneficial to explore more ones. In most neural network model evolution processes, the population size and training epochs of one individual remain fixed across generations \cite{xie2017genetic, he2021efficient, real2019regularized}. In practice, the computational resource is usually limited, which means that the more individuals are explored, the fewer training epochs can be performed. Nevertheless, inadequate training may not sufficiently reflect the performance of the individuals. So the number of explored individuals and training epochs present a trade-off. To reconcile exploration and exploitation \cite{wang2022evolutionary}, a strategy about dynamic population size and training epochs (DPSTE) is employed. Specifically, the population size decreases as the training epochs of each individual increases during evolution. This approach enables extensive exploration in the early stage of evolution while allowing promising individuals to undergo more profound evaluation with increasing training epochs in subsequent evolution. DPSTE strategy is also instrumental in identifying the fast-convergence individuals, as those with low convergence rate are eliminated early in evolution.

\subsubsection{Elitists preservation}
\begin{figure}[!htb]
\centering
\includegraphics[width=3.3in]{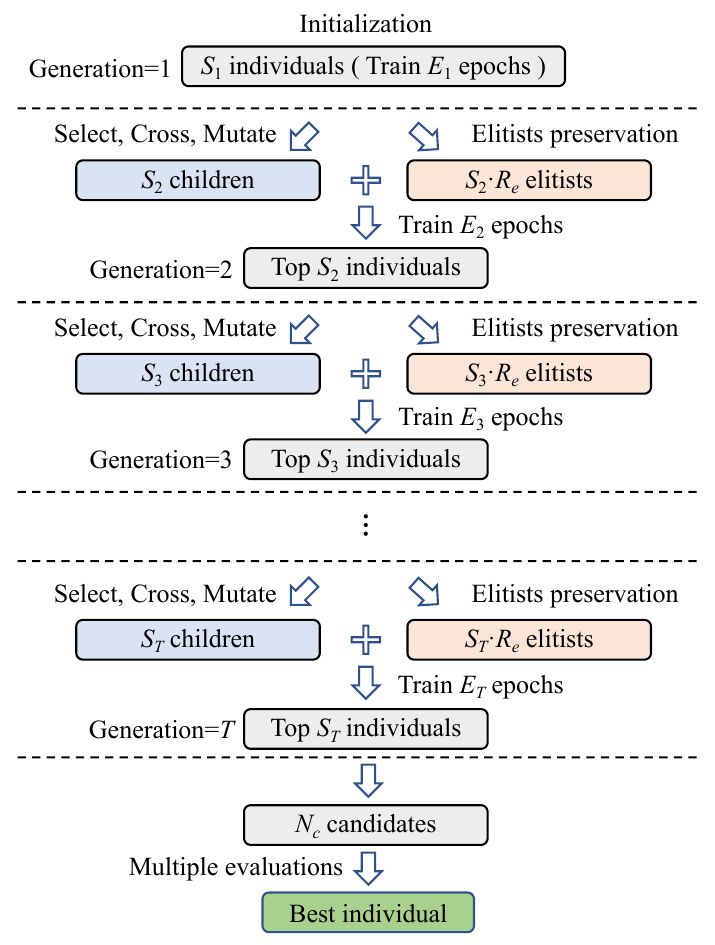}
\caption{Schematic diagram about evolutionary process. $S_g$ and $E_g$ denote the population size and training epochs about generation $g$, where $S_T$ $\leqslant$ $S_{T-1}$ $\leqslant$ ... $\leqslant$ $S_1$ and $E_T$ $\geqslant$ $E_{T-1}$ $\geqslant$ ... $\geqslant$ $E_1$. $R_e$ denotes the proportion of elitists.}
\label{fig:evolutionary_process}
\end{figure}

Elitists preservation strategy is adopted during evolution, which is widely used in evolutionary computation \cite{huang2015proposed}. Under this strategy, the elitists in previous generation are directly copied and combined with the children that are generated by selection, crossover and mutation from the previous generation to constitute a merged population. And the top-performing individuals in the merged population are chosen to constitute the next population. It is worth noting that the training epochs are different across generations under DPSTE strategy, so in choosing the top individuals in the merged population, the elitists in previous generation should not be directly compared with the children based on their fitness. For fair comparison, the elitists are retrained for the same training epochs with the children. The schematic diagram of evolutionary process is displayed in Fig.~\ref{fig:evolutionary_process}.

\subsubsection{Early stopping}
During evolution, numeric overflow often occurs when computing the loss functions with some generated activation functions, which is mainly caused by the computation of exponentiation with large exponent, division with small divisor and so on. In addition, if the searched model has too many parameters or the generated activation function is too complex, the corresponding training will be particularly time consuming. Individuals exhibiting these issues are undesirable. Drawing inspiration from early stopping strategy \cite{zhong2017practical}, the training for a specific individual is immediately stopped if numeric overflow occurs or the training time exceeds a preset threshold, which can save much time.

\subsubsection{Multiple evaluations}
Stability in accuracy across multiple trainings from scratch is a desirable trait of a model. It holds significance for the model's generalization across various conditions of the problem. For obtaining the stable model, multiple evaluations (i.e. multiple trainings) are performed on the candidates that are the top individuals in the final generation, so as to select the best one that has highest mean fitness.

\subsubsection{Parallelization}
The evolution process is naturally parallelizable \cite{hofmann2013performance}, because the training of each individual is independent. In our implement, the individuals in each generation are trained in parallel on multiple GPUs. For pursuing load balancing, the number of individuals assigned to each GPU is kept as equal as possible. The pseudocode of the proposed evolutionary computation method is presented in Algorithm ~\ref{alg:evo}. The time complexity of this algorithm is dominated by the training of individuals. Therefore, the time complexity can be approximated as $O(\sum_{g=1}^{T} S_{g}E_{g})$.

\renewcommand{\algorithmicrequire}{\textbf{Input:}} 
\renewcommand{\algorithmicensure}{\textbf{Output:}} 
\makeatletter
\newcommand\multiline[1]{\parbox[t]{\dimexpr\linewidth-\ALG@thistlm}{#1}}
\makeatother

\begin{algorithm*}[!htb]
\begin{algorithmic}[1]
\caption{}
\label{alg:evo}
\Require{Number of generations $T$; population size ${S}_{g}$ and training epochs ${E}_{g}$ about generation $g$; ratio of crossover to the sum of mutation and crossover ${R}_{cm}$; crossover rate ${R}_{c}$; mutation rates about number of layers ${R}_{l}$, number of neurons per hidden layer ${R}_{n}$, shortcut connections ${R}_{s}$, and activation function ${R}_{a}$; proportion of elitists ${R}_{e}$; number of candidates $N_c$; number of times about evaluating each candidate $N_e$.} 
\State\multiline{$ \boldsymbol{Newpop} \leftarrow $ Initialize the population with $S_1$ individuals} 
  \For{$g = 1$ to $T$}
    \State\multiline {Train each individual in $\boldsymbol{Newpop}$ for ${E}_{g}$ epochs based on PINNs} 
    \If{g $>$ $\rm{1}$}
        \State\multiline{ 
        $\boldsymbol{Elitists} \leftarrow$ Select top ${S_g}\cdot{R}_{e}$ individuals in $\boldsymbol{Pop}$ }
        \State\multiline{Retrain each individual in $\boldsymbol{Elitists}$ for ${E}_{g}$ epochs based on PINNs} 
        \State\multiline{$\boldsymbol{Newpop} \leftarrow$ Select top ${S}_{g}$ individuals in $\boldsymbol{Newpop} \cup \boldsymbol{Elitists}$} 
    \EndIf 
    \State\multiline{$\boldsymbol{Pop} \leftarrow \boldsymbol{Newpop}$}
    \If{g $<$ $T$}
        \State\multiline{Selection probabilities $\boldsymbol{Pro} \leftarrow$ Linear ranking about $\boldsymbol{Pop}$ based on Eq. \ref{eq:linear} } 
        \For{$m = 1$ to ${S_{g+1}}/2$}
        \State\multiline{Select two individuals $parent_1$ and $parent_2$ from $\boldsymbol{Pop}$ based on $\boldsymbol{Pro}$} 
            \If{rand($0,1$) $<$ ${R}_{cm}$}
                \State\multiline{$child_{m,1}$ and $child_{m,2}$ $\leftarrow$ Cross the structure gene and activation function gene of $parent_1$ and $parent_2$ with rate ${R}_{c}$ }
            \Else 
                \State\multiline{$child_{m,1}$ and $child_{m,2}$ $\leftarrow$ Mutate the genes of number of layers, number of neurons per hidden layer, shortcut connections, and activation function in $parent_1$ and $parent_2$ with rates ${R}_{l}$, ${R}_{n}$, ${R}_{s}$ and ${R}_{a}$ respectively} 
            \EndIf
        \EndFor 
      \State\multiline{$\boldsymbol{Newpop} \leftarrow \{{child_{m,1}}, {child_{m,2}}\}_{m=1, 2,...,{S_{g+1}}/2}$}
    \EndIf
  \EndFor
    \State\multiline{$\boldsymbol{Candidates} \leftarrow$ Top $N_c$ individuals in $\boldsymbol{Pop}$ } 
    \State\multiline{Train each individual in $\boldsymbol{Candidates}$ from scratch $N_e$ times, with ${E}_{T}$ epochs per training} 
\Ensure{The individual with the highest mean fitness in $\boldsymbol{Candidates}$ }  
\end{algorithmic}
\end{algorithm*}

\section{Experiments}
\label{sec:Experiments}

\begin{table}[!htb]
\renewcommand\arraystretch{1.0}
\setlength\tabcolsep{2pt}
\caption{Hyperparameters in evolution.}
    \centering
    \begin{footnotesize}
    \begin{tabular}{cc}
    \toprule   Hyperparameters  & Value   \\
    \midrule %\hline 
                          ratio of crossover to the sum of mutation and crossover ${R}_{cm}$  &  0.5    \\
                          crossover rate ${R}_{c}$     &  1.0    \\
                          mutation rate of number of layers ${R}_{l}$  &  0.3   \\    
                          mutation rate of number of neurons per hidden layer ${R}_{n}$  &  0.3  \\
                          mutation rate of shortcut connections ${R}_{s}$ &  0.3    \\
                          mutation rate of activation function ${R}_{a}$ & 0.7    \\
                          proportion of elitists ${R}_{e}$  &  0.25    \\
                          number of candidates $N_c$   &  3    \\
                          number of times about evaluating each candidate $N_e$  &  4    \\
                          maximum number of nodes in activation function $M_N$ &  7    \\
                          maximum number of parameters in activation function $M_E$ &  3    \\
    \bottomrule
    \end{tabular}
    \end{footnotesize} %
    \label{tab:hyperparameters}
\end{table}

For verifying the approximation accuracy and convergence rate of the models discovered by the proposed evolution with DPSTE strategy (evo-w/-DPSTE), three classical PDEs including Klein-Gordon equation, Burgers equation, and Lamé equations are introduced for experiment. The hyperparameters in evolution are listed in Table \ref{tab:hyperparameters}. We compare evo-w/-DPSTE with several other model searching methods. Firstly, the popular Bayesian optimization method \cite{guo2022stochastic, escapil2023hyper, bischof2021multi} is used. In Bayesian optimization, both FcNet and FcResNet with regular shortcut connections (each shortcut spans two fully connected layes) are chosen as the basic network structures, and the alternative activation functions are given in Appendix~\ref{sec:Appendix_initial}. For demonstrating the effect of evolutionary mechanism, the random search method is also employed for comparing. In random search, the linear ranking selection in evolution is replaced by random selection, and the candidates are chosen from individuals with the highest fitness in each generation. At last, the evolution without DPSTE strategy (evo-w/o-DPSTE) is also carried out for comparing. From the perspective of search space, random search, evo-w/o-DPSTE and evo-w/-DPSTE share the same search space, which covers the search space of Bayesian optimization based on FcNet and FcResNet. In all methods, the number of layers is searched from 3 to 11, and the number of neurons per hidden layer is searched from 20 to 50 with an interval of 2.

In random search and evolution, the individuals in each generation are trained in parallel on 4 Tesla V100 (32G memory) GPUs. Bayesian optimization is performed on one GPU. For fair comparison, the total training epochs of evo-w/-DPSTE is kept no more than the other methods. Each searching method runs three times. The consuming time of evo-w/-DPSTE ranges from a few hours to more than 10 hours, contingent on the equations being solved. Throughout all experiments, Kaiming initialization \cite{he2015delving} with uniform distribution is employed to initialize the network parameters, and Limited-Memory Broyden-Fletcher-Goldfarb-Shanno (L-BFGS) algorithm with the strong Wolfe line search strategy is adopted for optimizing the network parameters. In solving PDEs by PINNs, uniform sampling is adopted. All trainings are performed based on double precision (64-bit float).

\subsection{Klein-Gordon equation}
\label{section:Klein-exp}

\begin{figure*}[!H]
\centering
\subfloat[Loss histories (case I)]{\includegraphics[width=2.2in]{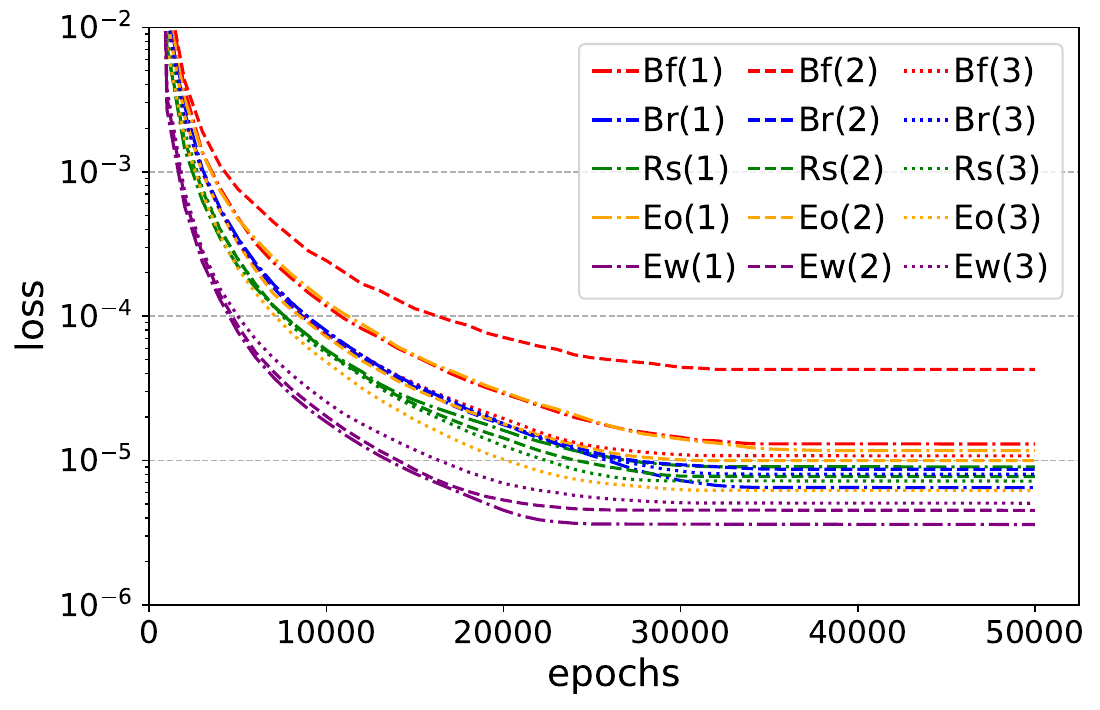}%
\label{Klein-Gordon_loss_ori}}
\hspace{2mm}
\subfloat[Loss histories (case II)]{\includegraphics[width=2.2in]{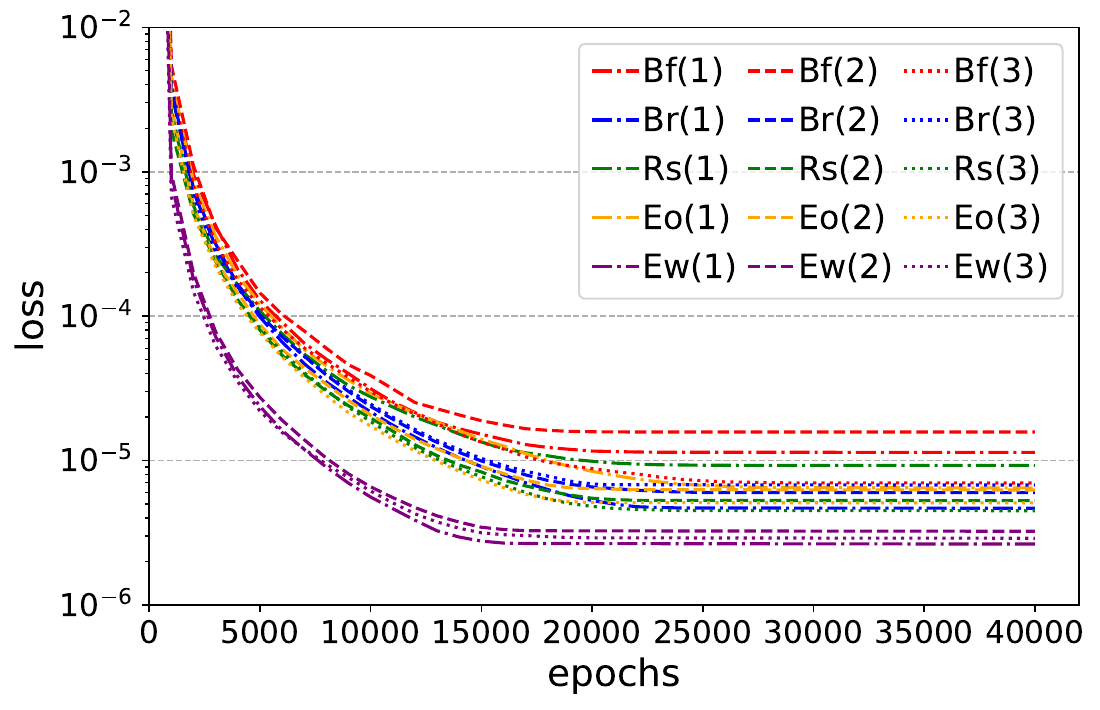}%
\label{Klein-Gordon_loss_gen1}}
\hspace{2mm}
\subfloat[Loss histories (case III)]{\includegraphics[width=2.2in]{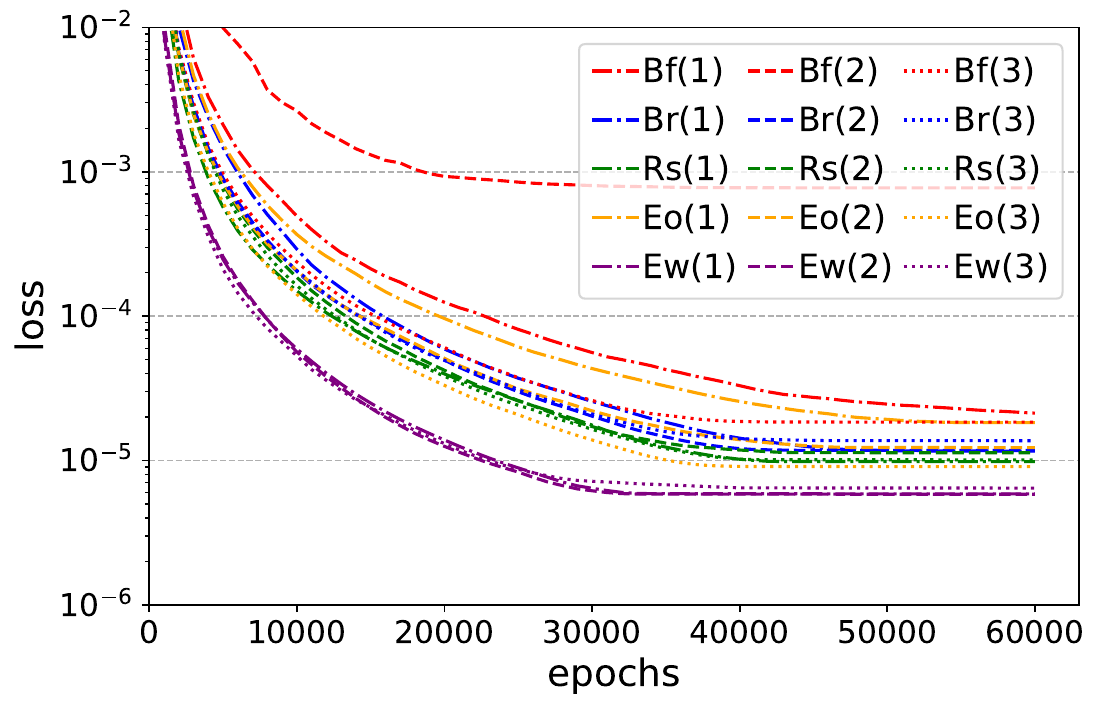}%
\label{Klein-Gordon_loss_gen2}}
\vspace{-1mm}
\subfloat[Error histories (case I)]{\includegraphics[width=2.2in]{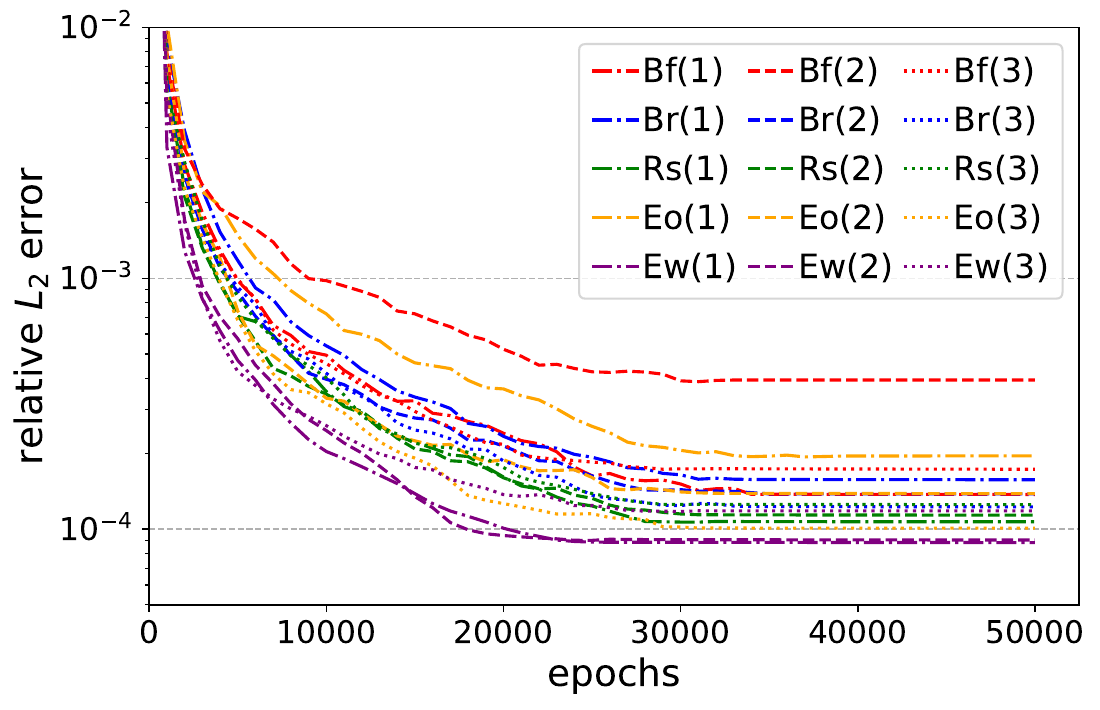}%
\label{Klein-Gordon_error_ori}}
\hspace{2mm}
\subfloat[Error histories (case II)]{\includegraphics[width=2.2in]{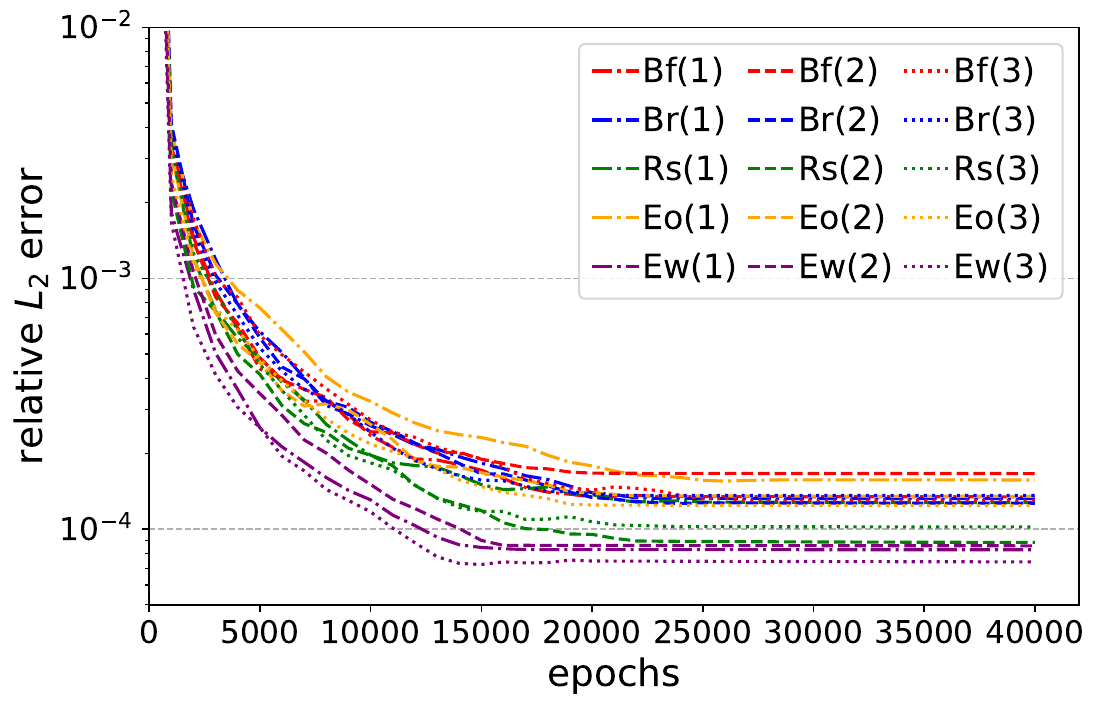}%
\label{Klein-Gordon_error_gen1}}
\hspace{2mm}
\subfloat[Error histories (case III)]{\includegraphics[width=2.2in]{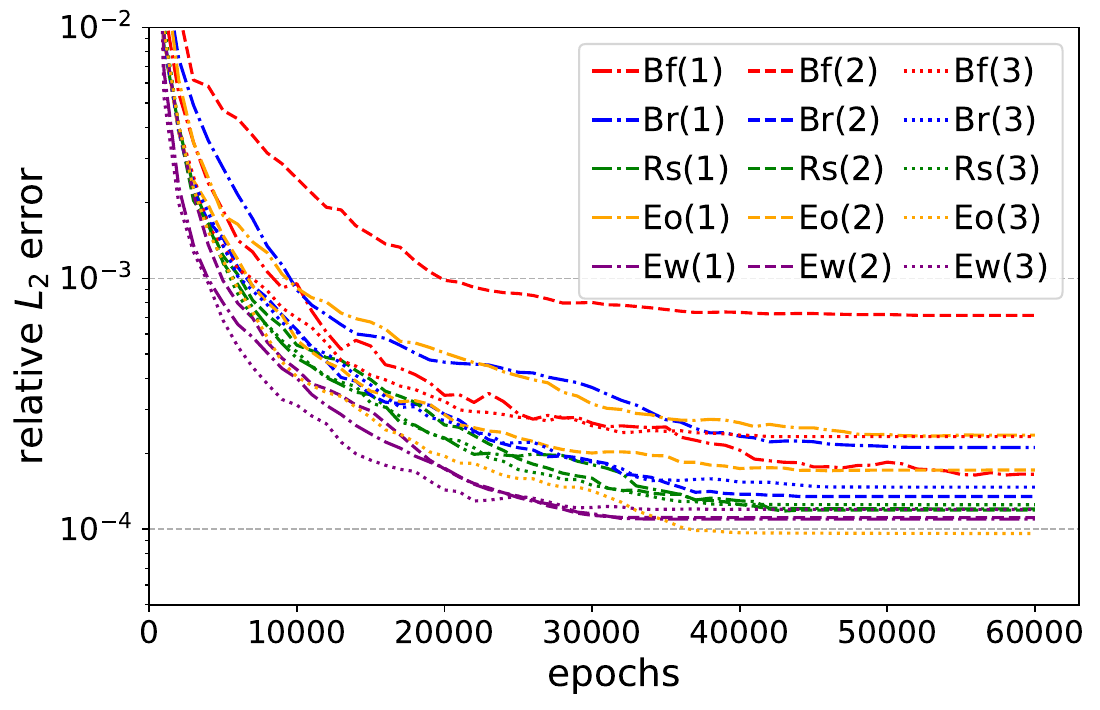}%
\label{Klein-Gordon_error_gen2}}
\caption{Convergence histories of loss and relative $L_2$ error in training the models discovered by each searching method for solving Klein-Gordon equation. (a), (b), and (c) show the loss histories about cases I, II, and III respectively. (d), (e), and (f) show the relative $L_2$ error histories about cases I, II, and III respectively. Each curve of convergence history is the average result of ten independent trials. In legend, Bf and Br represent the Bayesian optimization based on FcNet and FcResNet, Rs represent random search, and Eo and Ew represent the evolution without and with DPSTE strategy. Besides, (1), (2), and (3) represent three independent searches.}
\label{fig:Klein-Gordon_loss_error}
\end{figure*}

The time-dependent Klein-Gordon equation is considered in this example, which is a basic nonlinear equation in quantum field theory. An initial-boundary value problem with one spatial dimension is expressed as 
\begin{equation}
\begin{aligned}
& u_{t t}+\alpha u_{x x}+\beta u+\gamma u^k=f(x, t), && (x, t) \in \Omega \times(0, T], \\
& u(x, 0)=g_1(x), && x  \in \overline{\Omega} , \\
& u_t(x, 0)=g_2(x), && x  \in \overline{\Omega} , \\
& u(x, t)=h(x, t), && (x, t) \in \partial \Omega \times[0, T],
\label{eq:Klein-Gordon}
\end{aligned}
\end{equation}
where the constants are set as $\alpha=-1$, $\beta=0$, $\gamma=1$, $k=3$, the spatial computational domain $\overline\Omega = [0,1]$, and $T=1$. The solution is fabricated as $u(x, t)=x \cos (5 \pi t)+(x t)^3$ (case I), and $f(x, t)$, $g_1(x)$, $g_2(x)$, and $h(x,t)$ are derived by this solution. These settings refer to \cite{wang2021understanding, son2023enhanced, chen2021improved}. And for assessing the generalization capability of the models discovered by each searching method under case I, $f(x, t)$, $g_1(x)$, $g_2(x)$, and $h(x,t)$ are moderately changed, which are realized through transforming the fabricated solution to $u(x, t)=x \cos (4 \pi t)+(x t)^3$ (case II) and $u(x, t)=x \cos (6 \pi t)+1.2(x t)^3$ (case III). In cases II and III, the models discovered in case I are retrained from scratch. In this example, 81 initial points, 162 boundary points, and 3600 collocation points are used in PINNs and 10201 testing points are used to evaluate the relative $L_2$ error of the models. Both penalty coefficients for initial and boundary loss terms are set to 100. In Bayesian optimization, the training epochs and number of explored individuals are set to 5000 and 200. The population size and training epochs about each generation of random search, evo-w/-DPSTE and evo-w/o-DPSTE are listed in Table \ref{tab:DPSTE_epochs_klein_lame}. Due to adopting the DPSTE strategy, evo-w/-DPSTE explores several times more individuals compared to Bayesian optimization and evo-w/o-DPSTE, without increasing the total training epochs.

\begin{figure*}[!htb]
\centering
\subfloat[Case I (original): $u(x, t)=x \cos (5 \pi t)+(x t)^3$]{\includegraphics[width=6.8in]{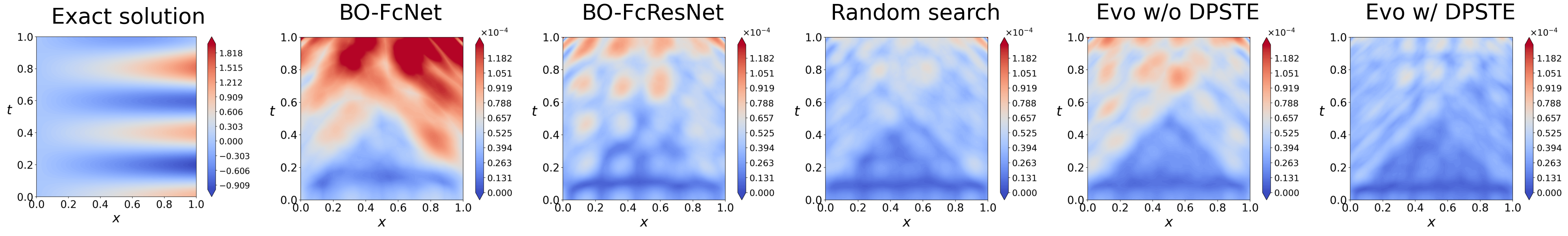}}
\vspace{-1mm}
\subfloat[Case II (generalized): $u(x, t)=x \cos (4 \pi t)+(x t)^3$]{\includegraphics[width=6.8in]{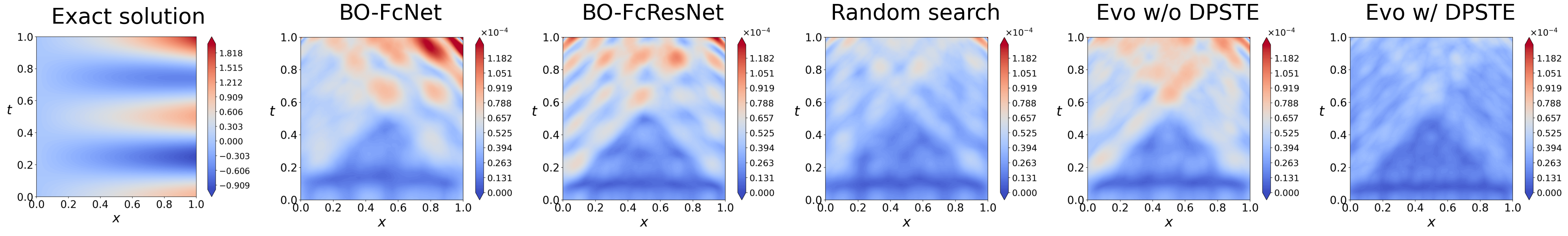}}
\vspace{-1mm}
\subfloat[Case III (generalized): $u(x, t)=x \cos (6 \pi t)+1.2(x t)^3$]{\includegraphics[width=6.8in]{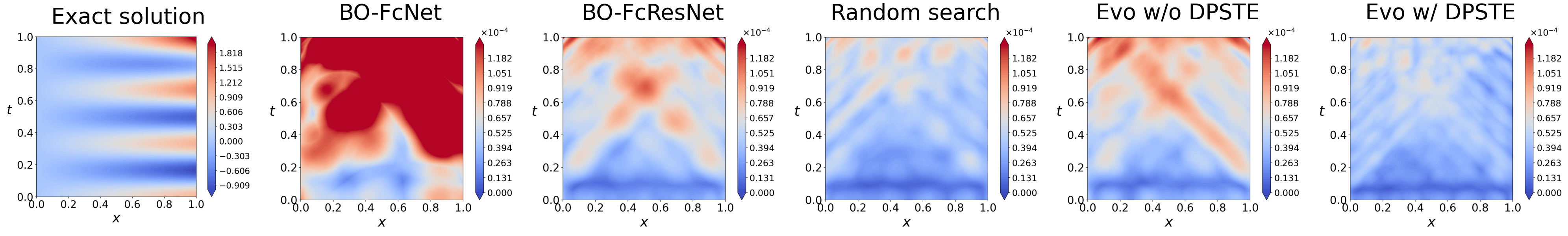}}
\hfil
\caption{The exact solution (first column) of Klein-Gordon equation and absolute error (last five columns) of the models discovered by each searching method. (a) Case I (original), (b) Case II (generalized), and (c) Case III (generalized). `BO' represents Bayesian optimization. The absolute error is the average result of the three models discovered by each searching method.} 
\label{fig:Klein-Gordon_contourf}
\vspace{-4mm}
\end{figure*}

\begin{table*}[!htp]
\renewcommand\arraystretch{1.0}
\setlength\tabcolsep{5pt}
\caption{The population size and training epochs about each generation of random search (2-3 rows), evo-w/-DPSTE (2-3 rows) and evo-w/o-DPSTE (4-5 rows) in discovering the models for solving Klein-Gordon equation.}
    \centering
    \begin{footnotesize}
    \begin{tabular}{cccccccccccccccc}
    \toprule   Generation  & 1 & 2 & 3 & 4 & 5 & 6 & 7 & 8 & 9 & 10 & 11 & 12 & 13 & 14 & 15  \\
    \midrule %\hline 
               Population size      & 1000 & 250 & 125 & 85 & 65 & 50 & 40 & 30 & 25 & 20 & 15 & 15 & 15 & 10 & 10    \\   
               Training epochs   & 100 & 200 & 400 & 600 & 800 & 1000 & 1200 & 1600 & 2000 & 2500 & 3000 & 3500 & 4000 & 4500 & 5000    \\
    \midrule %\hline 
               Population size      & 30 & 30 & 30 & 30 & 30 & 30    \\ 
               Training epochs   & 5000 & 5000 & 5000 & 5000 & 5000 & 5000    \\
    \bottomrule
    \end{tabular}
    \end{footnotesize} %
    \label{tab:DPSTE_epochs_klein_lame}
\end{table*}

The convergence histories of loss and relative $L_2$ error in training the models discovered by each searching method are shown in Fig.~\ref{fig:Klein-Gordon_loss_error}. In Fig.~\ref{Klein-Gordon_loss_ori}, ~\ref{Klein-Gordon_loss_gen1} and ~\ref{Klein-Gordon_loss_gen2}, the three models discovered by evo-w/-DPSTE exhibit the lowest minimum loss value and the fastest convergence rate compared to the other models in all three cases. In Fig.~\ref{Klein-Gordon_error_ori}, ~\ref{Klein-Gordon_error_gen1} and ~\ref{Klein-Gordon_error_gen2}, the comparisons about relative $L_2$ error histories among models are close to their loss histories. From the generalized problems (cases II and III), evo-w/-DPSTE models exhibit commendable generalization performance under different conditions. In order to show the results more intuitively, the exact solution and absolute error of the models discovered by each searching method are illustrated in Fig.~\ref{fig:Klein-Gordon_contourf}. Obviously, the BO-FcNet models without shortcut connections perform the worst, indicating that shortcut connections play a crucial role in solving this problem. On average, the model discovered by evo-w/-DPSTE has the highest approximation accuracy in three cases. The superiority of evo-w/-DPSTE mainly stems from its large search space and the exploration of numerous individuals. The detailed value of relative $L_2$ error, structure and activation function of each discovered model are exhibited in Appendix~\ref{sec:Appendix_discovered_model}. And the model structures and activation functions discovered by evo-w/-DPSTE are visualized in Appendix~\ref{sec:Appendix_discovered_structure} and Appendix~\ref{sec:Appendix_discovered_activation}, respectively.

\subsection{Burgers equation}

\begin{figure*}[!htp]
\centering
\subfloat[Loss histories (case I)]{\includegraphics[width=2.2in]{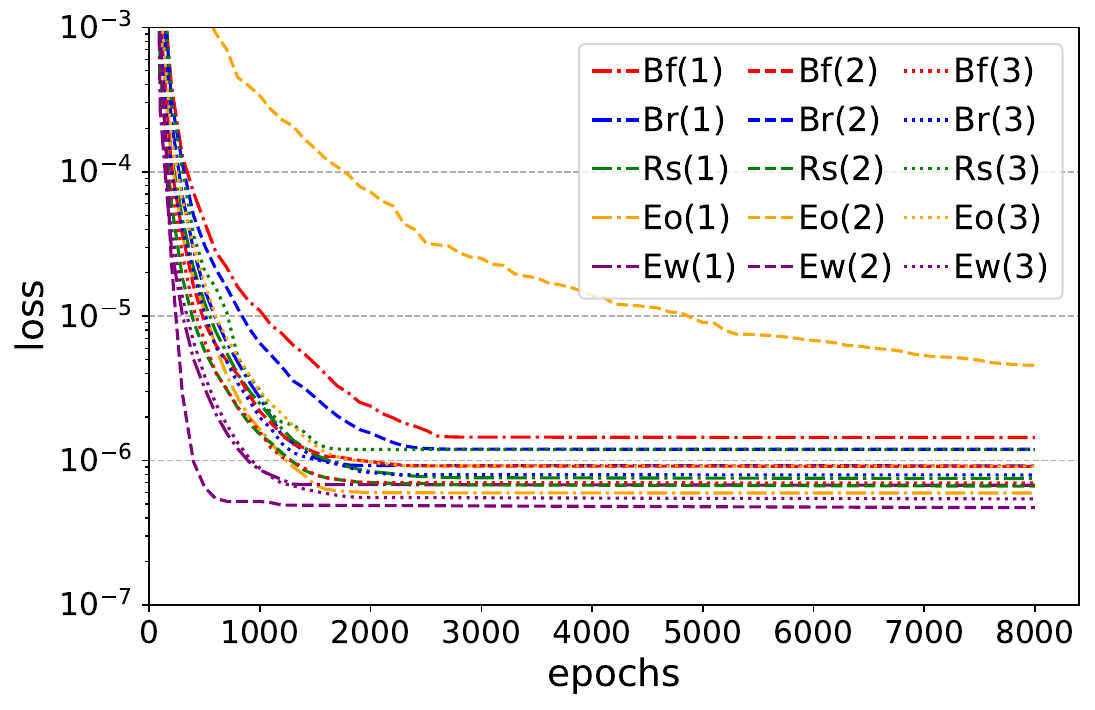}%
\label{fig:Burgers_loss_ori}}
\hspace{2mm}
\subfloat[Loss histories (case II)]{\includegraphics[width=2.2in]{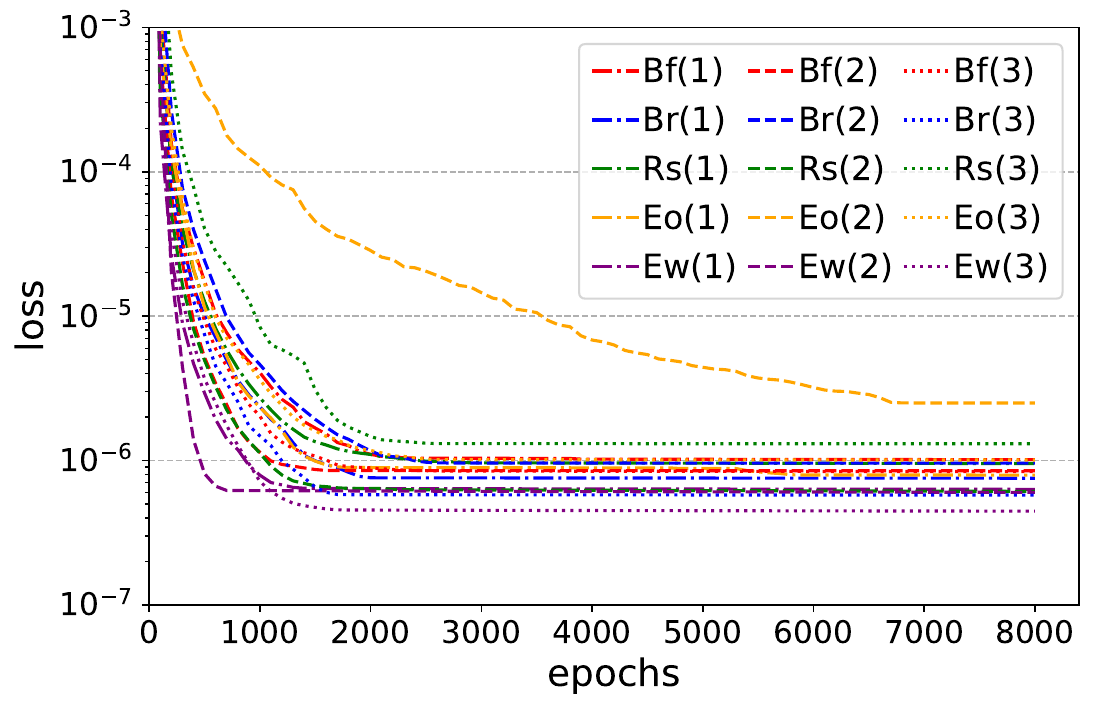}%
\label{fig:Burgers_loss_gen1}}
\hspace{2mm}
\subfloat[Loss histories (case III)]{\includegraphics[width=2.2in]{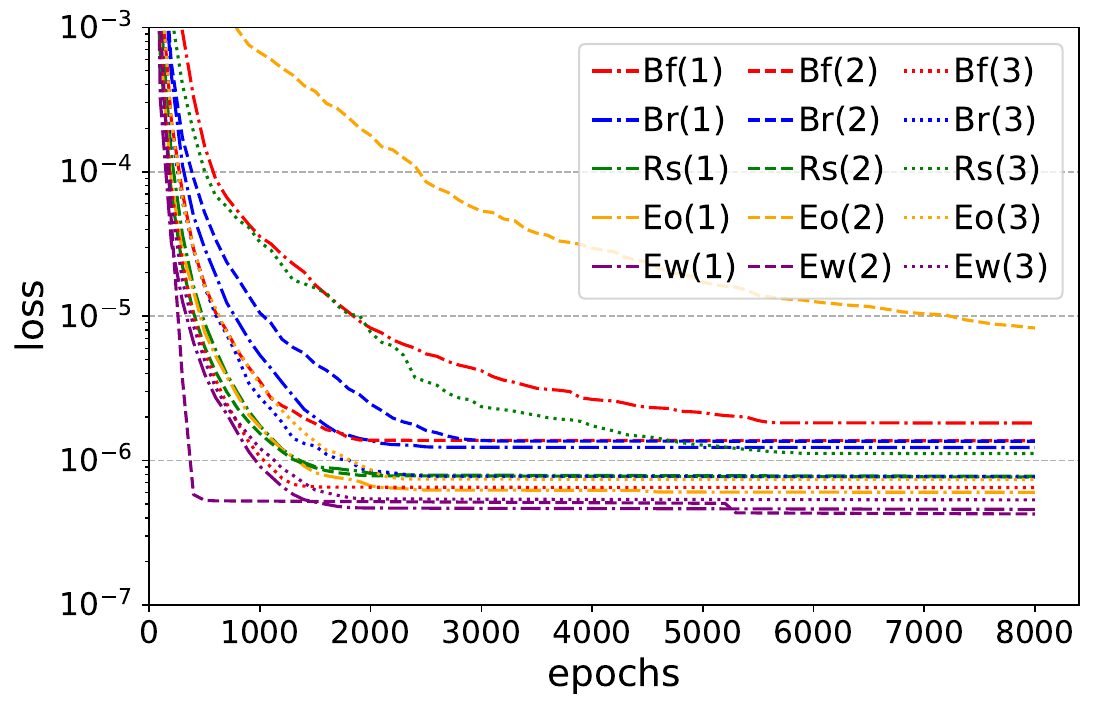}%
\label{fig:Burgers_loss_gen2}}
\vspace{-1mm}
\subfloat[Error histories (case I)]{\includegraphics[width=2.2in]{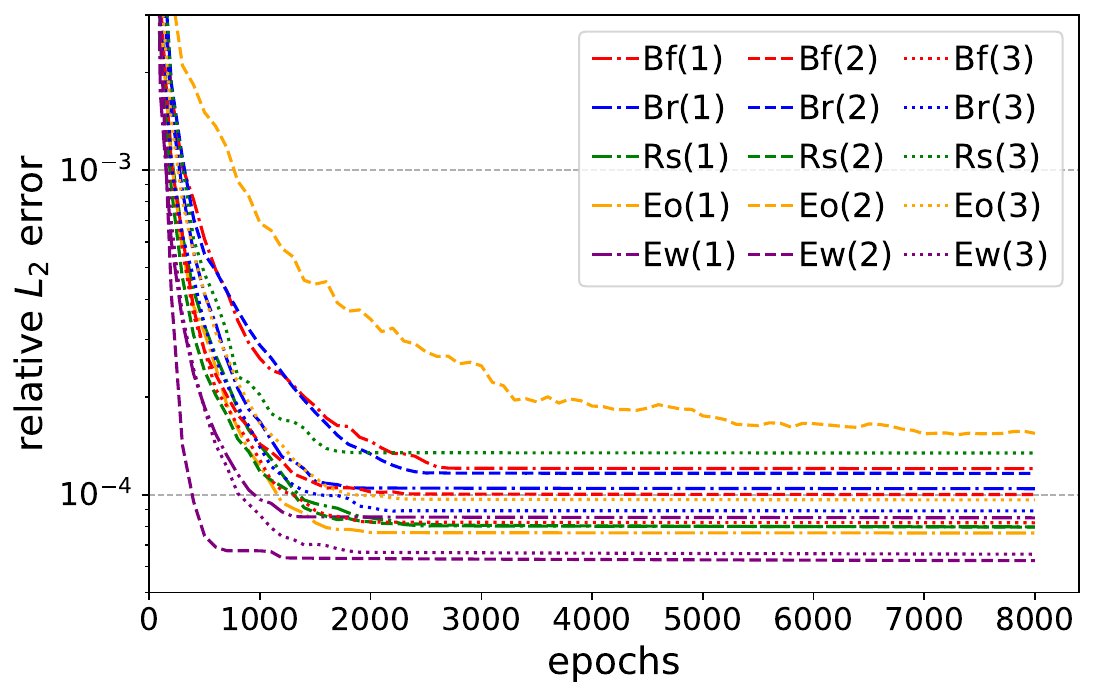}%
\label{fig:Burgers_error_ori}}
\hspace{2mm}
\subfloat[Error histories (case II)]{\includegraphics[width=2.2in]{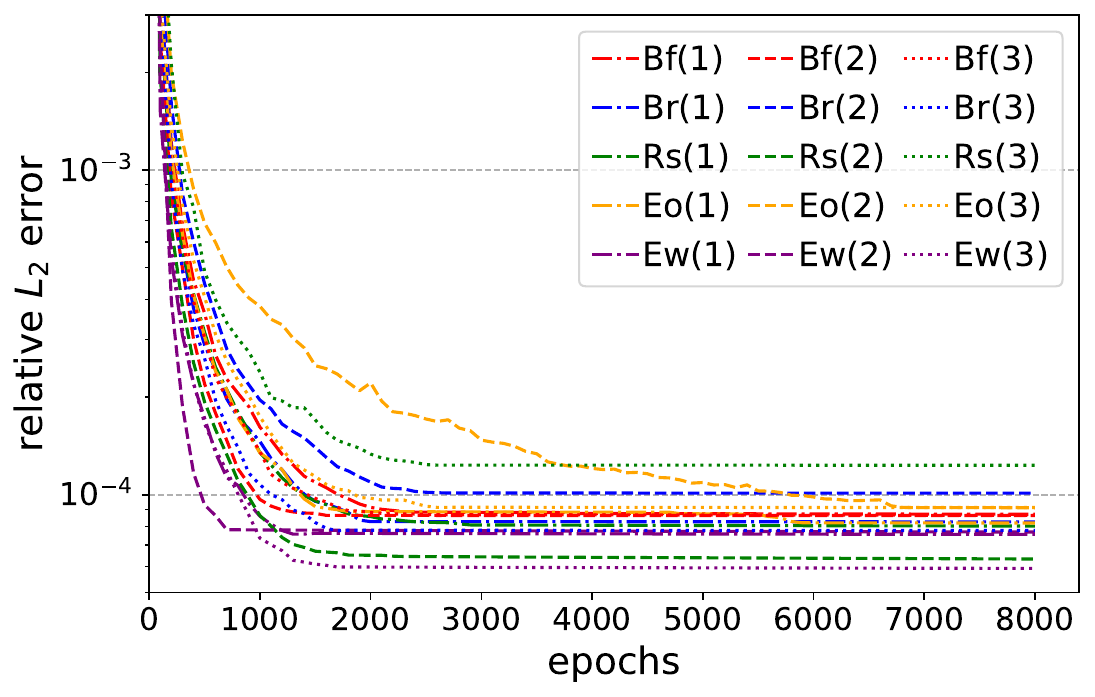}%
\label{fig:Burgers_error_gen1}}
\hspace{2mm}
\subfloat[Error histories (case III)]{\includegraphics[width=2.2in]{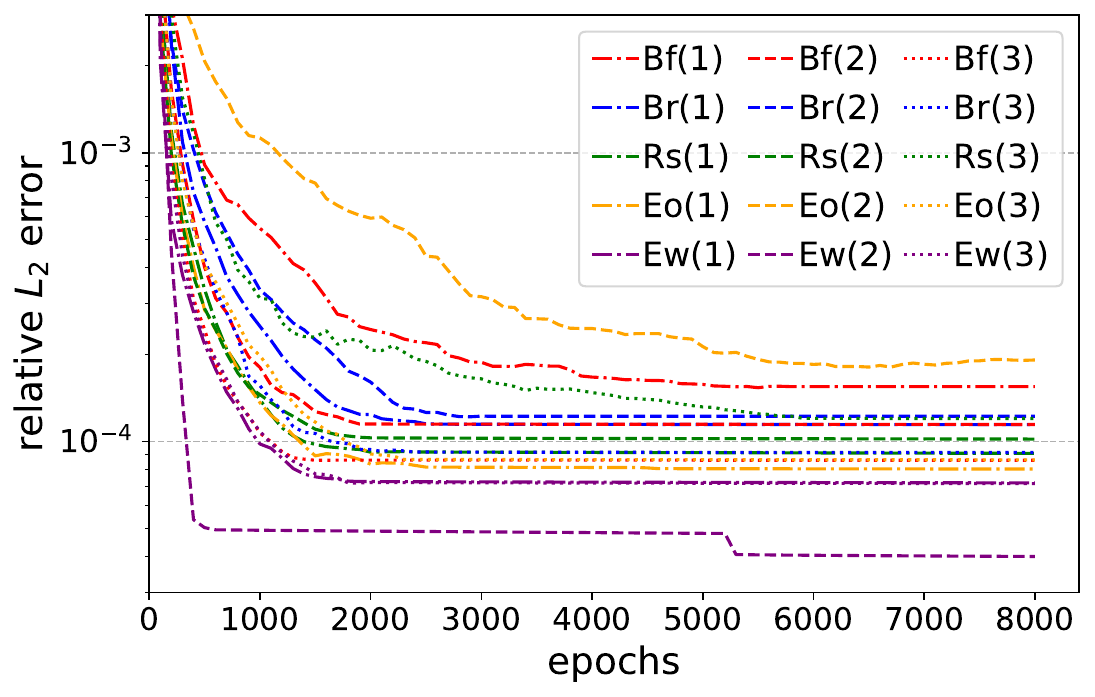}%
\label{fig:Burgers_error_gen2}}
\caption{Convergence histories of loss and relative $L_2$ error in training the models discovered by each searching method for solving Burgers equation. (a), (b), and (c) show the loss histories about cases I, II, and III respectively. (d), (e), and (f) show the relative $L_2$ error histories about cases I, II, and III respectively. Each curve of convergence history is the average result of ten independent trials.}
\vspace{-4mm}
\label{fig:Burgers_loss_error}
\end{figure*}

\begin{figure*}[!htp]
\centering
\subfloat[Case I (original): $\alpha$ = 0.1]{\includegraphics[width=6.7in]{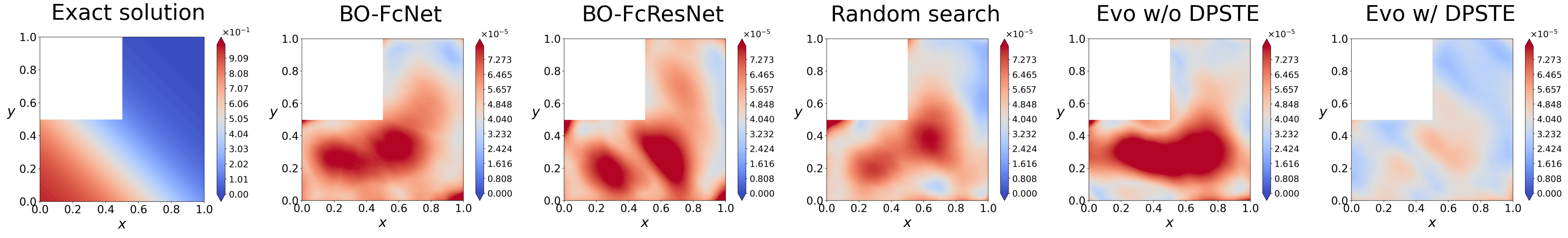}}
\vspace{-1mm}
\subfloat[Case II (generalized): $\alpha$ = 0.15]{\includegraphics[width=6.7in]{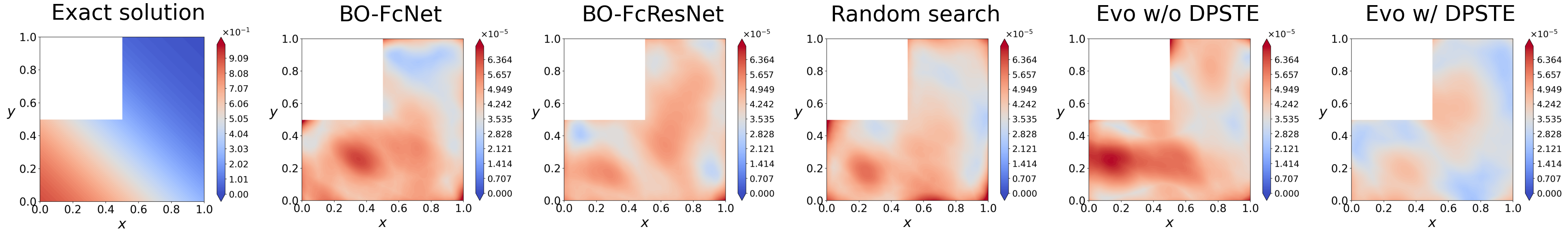}}
\vspace{-1mm}
\subfloat[Case III (generalized): $\alpha$ = 0.05]{\includegraphics[width=6.7in]{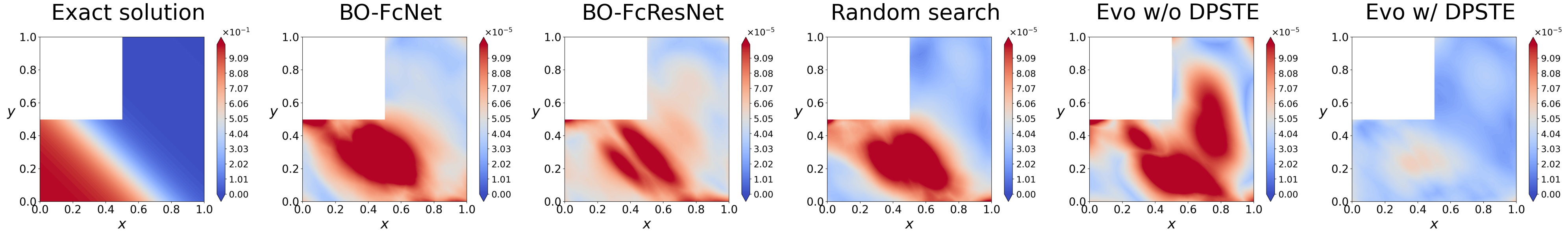}}
\caption{The snapshot ($t=0.7$) of the exact solution (first column) of Burgers equation and absolute error (last five columns) of the models discovered by each searching method. (a) Case I (original), (b) Case II (generalized), and (c) Case III (generalized). The absolute error is the average result of the three models discovered by each searching method.}
\vspace{-4mm}
\label{fig:Burgers_contourf_0.7}
\end{figure*}

The nonlinear Burgers equation has been applied in numerous fields, such as fluid mechanics, nonlinear acoustics, gas dynamics and so on \cite{yang2021class}. The two-dimensional Burgers equation with scalar field $u$ can be expressed as
\begin{equation}
\begin{aligned}
& u_t + u\left(u_x+u_y\right) - \alpha\left(u_{xx}+u_{yy}\right) = 0 , && (x, y) \in \Omega, t \in (0, T], \\
& u(x, y, 0)=g(x, y), && (x, y) \in \overline{\Omega}, \\
& u(x, y, t)=h(x, y, t), && (x, y) \in \partial \Omega, t \in [0, T],
\label{eq:Burgers}
\end{aligned}
\end{equation}
where $\alpha$ is the reciprocal of Reynolds number. The exact solution is set as $u(x, y, t)=1 /\left(1+e^{(0.5/\alpha)(x+y-t)}\right)$, and $g(x, y)$ and $h(x,y,t)$ are derived by this solution. The spatial computational domain $\overline\Omega = [0,1]^2 \backslash([0,0.5)\times(0.5,1])$, and $T=2$. In original problem (case I), $\alpha$ is set to 0.1 according to \cite{li2021deep}. In this example, the generalization capability of the discovered models with respect to the equation coefficient is assessed. In generalized problems, $\alpha$ is set to 0.15 (case II) and 0.05 (case III) respectively. In training the models based on PINNs, 721 initial points, 3720 boundary points, and 6000 collocation points are employed. And 100776 testing points are used in evaluating the relative $L_2$ error of models. Both penalty coefficients for initial and boundary loss terms are set to 100. In Bayesian optimization, the training epochs and number of explored individuals are set to 3000 and 300. The population size and training epochs about each generation of random search, evo-w/-DPSTE and evo-w/o-DPSTE are listed in Table \ref{tab:DPSTE_epochs_burgers}.

\begin{figure*}[!htp]
\centering
\subfloat[Case I (original): $\alpha$ = 0.1]{\includegraphics[width=6.7in]{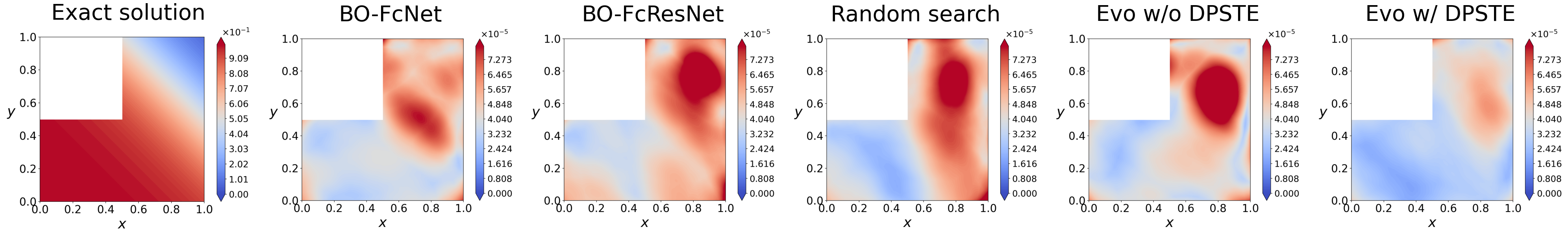}}
\vspace{-2mm}
\subfloat[Case II (generalized): $\alpha$ = 0.15]{\includegraphics[width=6.7in]{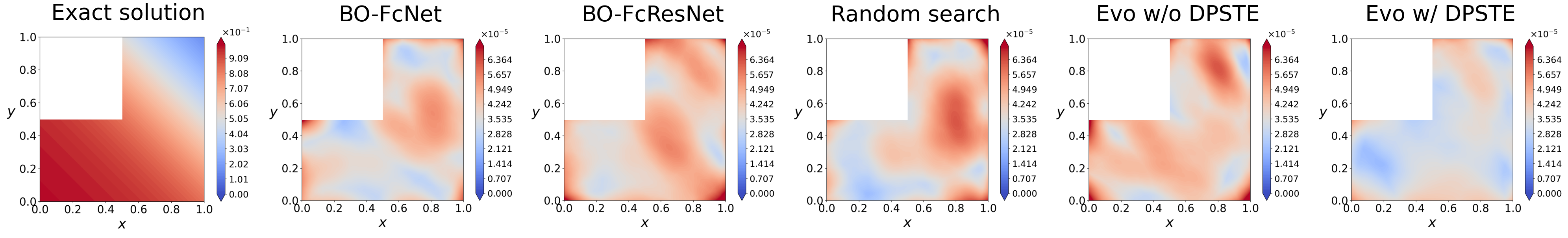}}
\vspace{-2mm}
\subfloat[Case III (generalized): $\alpha$ = 0.05]{\includegraphics[width=6.7in]{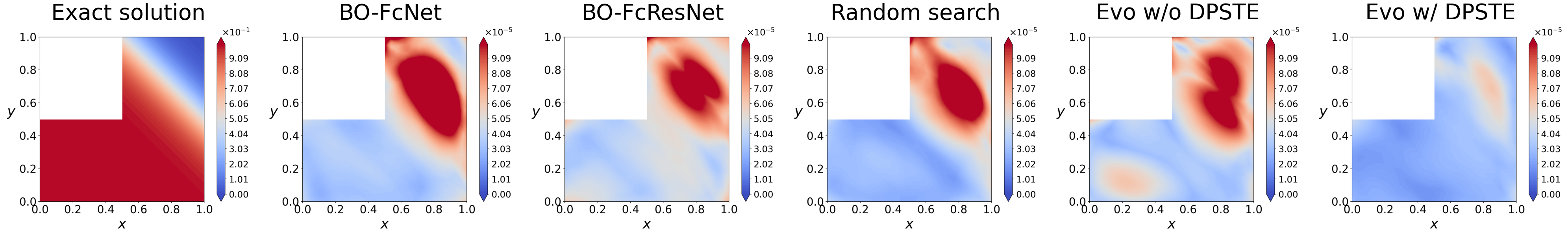}}
\caption{The snapshot ($t=1.5$) of the exact solution (first column) of Burgers equation and absolute error (last five columns) of the models discovered by each searching method. (a) Case I (original), (b) Case II (generalized), and (c) Case III (generalized). The absolute error is the average result of the three models discovered by each searching method.}
\vspace{-3mm}
\label{fig:Burgers_contourf_1.5}
\end{figure*}

\begin{table*}[!htp]
\renewcommand\arraystretch{1.0}
\setlength\tabcolsep{5pt}
\caption{The population size and training epochs about each generation of random search (2-3 rows), evo-w/-DPSTE (2-3 rows) and evo-w/o-DPSTE (4-5 rows) in discovering the models for solving Burgers equation.}
    \centering
    \begin{footnotesize}
    \begin{tabular}{cccccccccccccccc}
    \toprule   Generation  & 1 & 2 & 3 & 4 & 5 & 6 & 7 & 8 & 9 & 10 & 11 & 12 & 13 & 14 & 15  \\
    \midrule %\hline 
               Population size      & 1000 & 200 & 100 & 65 & 50 & 40 & 35 & 30 & 25 & 20 & 20 & 20 & 15 & 15 & 15    \\  
               Training epochs   & 100 & 200 & 400 & 600 & 800 & 1000 & 1200 & 1400 & 1600 & 1800 & 2000 & 2200 & 2400 & 2700 & 3000    \\     
    \midrule %\hline 
                Population size      & 40 & 40 & 40 & 40 & 40 & 40    \\ 
               Training epochs    & 3000 & 3000 & 3000 & 3000 & 3000 & 3000    \\  
    \bottomrule
    \end{tabular}
    \end{footnotesize} %
    \label{tab:DPSTE_epochs_burgers}
\vspace{-1mm}
\end{table*}

After performing each searching method, the convergence histories of loss and relative $L_2$ error in training the discovered models are shown in Fig.~\ref{fig:Burgers_loss_error}. As can be seen from this figure, the convergence rate and approximation accuracy of different models may vary greatly. In the original problem (case I), the model with highest approximation accuracy and fastest convergence rate is discovered by evo-w/-DPSTE. In the generalized problems (cases II and III), evo-w/-DPSTE models also show their superiority overall, although a few other models, such as Rs(2) in case II, present the competitiveness. Two snapshots of the exact solution and absolute error of the models discovered by each searching method are illustrated in Fig.~\ref{fig:Burgers_contourf_0.7} ($t=0.7$) and Fig.~\ref{fig:Burgers_contourf_1.5} ($t=1.5$), which intuitively demonstrate the superior approximation accuracy of evo-w/-DPSTE models. Different from the experiment \ref{section:Klein-exp}, the discovered superior models have sparse shortcut connections as shown in Table \ref{tab:Burgers_appendix} and Fig.~\ref{fig:Burgers_str}, suggesting that the shortcut connections may not be essential in solving this problem. Appendix~\ref{sec:Appendix_discovered_model} exhibits the relative $L_2$ error, structure and activation function of each discovered model in detail. And Appendix~\ref{sec:Appendix_discovered_structure} and Appendix~\ref{sec:Appendix_discovered_activation} visualize the structure and activation function of the models discovered by evo-w/-DPSTE, respectively.

\subsection{Lamé equations}

\begin{figure}[!htp]
\centering
\includegraphics[width=2.5in]{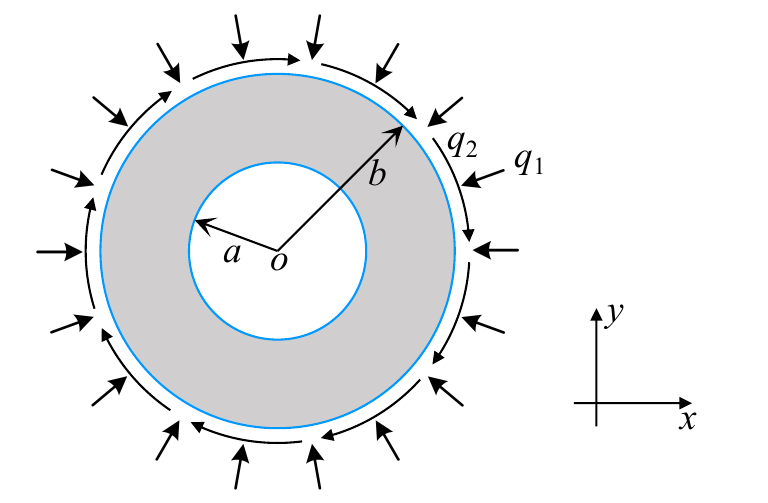}
\caption{The cross section of a thick-walled cylinder.} % under the uniform pressure $q_1$ and shearing force $q_2$
\label{fig:annulus}
\vspace{-3mm}
\end{figure}

\begin{figure*}[H]
\centering
\subfloat[Loss histories (case I)]{\includegraphics[width=2.0in]{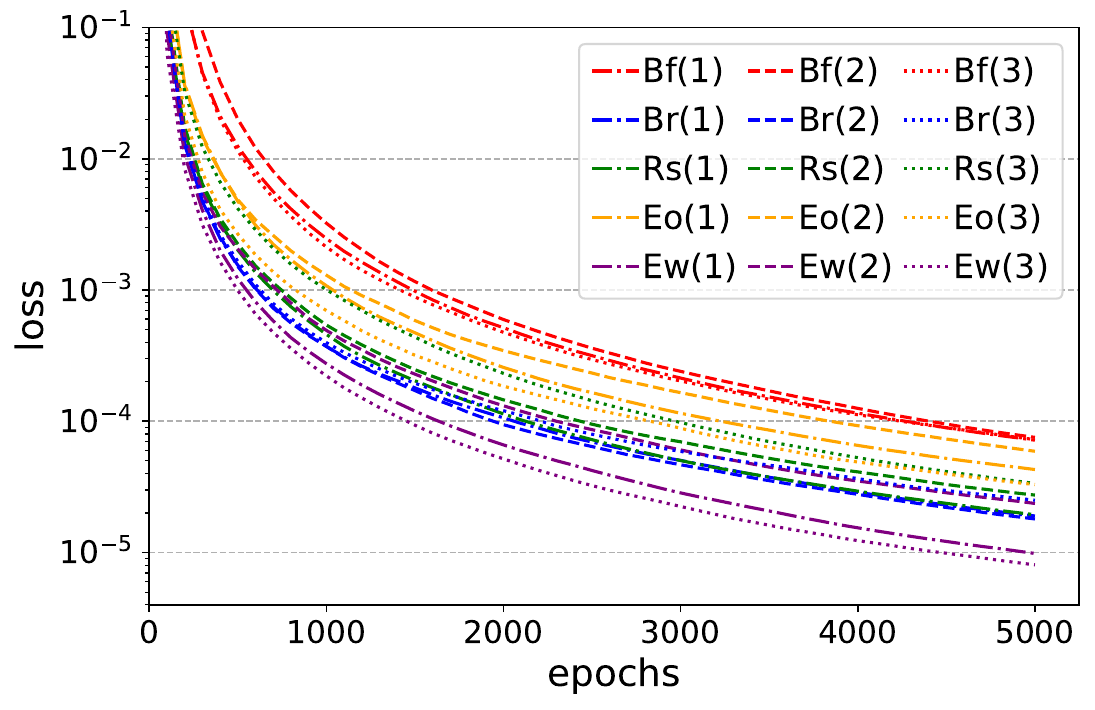}%
\label{fig:Lame-loss-history-ori}}
\hspace{3mm}
\subfloat[Error histories about $u$ (case I)]{\includegraphics[width=2.0in]{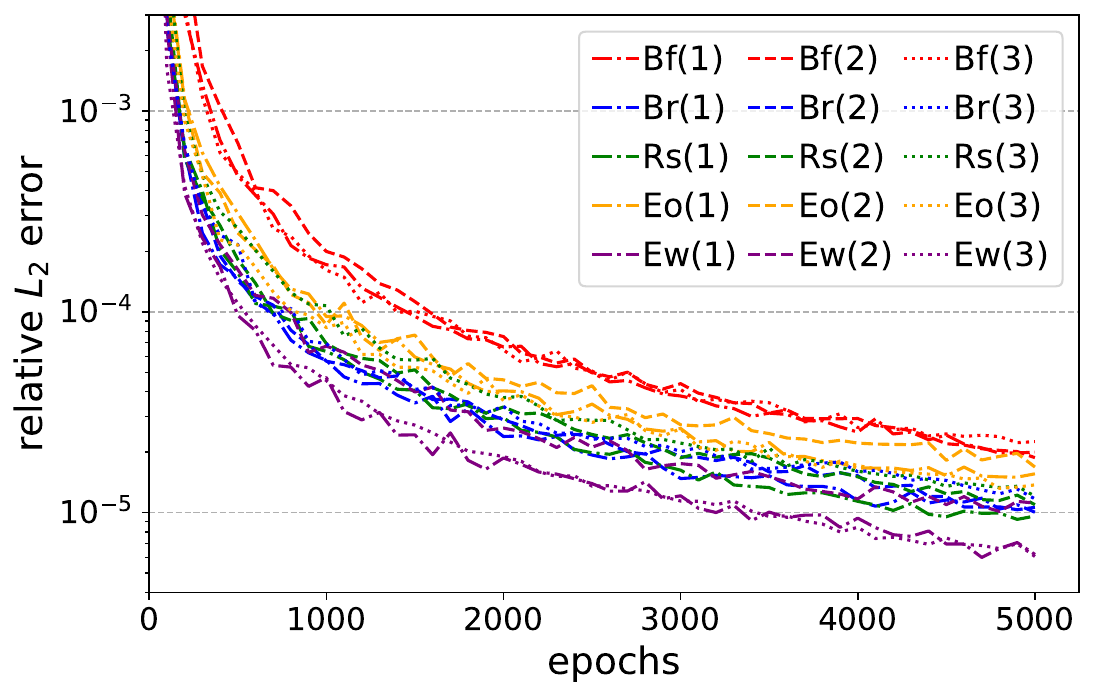}%
\label{fig:Lame-error0-history-ori}}
\hspace{3mm}
\subfloat[Error histories about $v$ (case I)]{\includegraphics[width=2.0in]{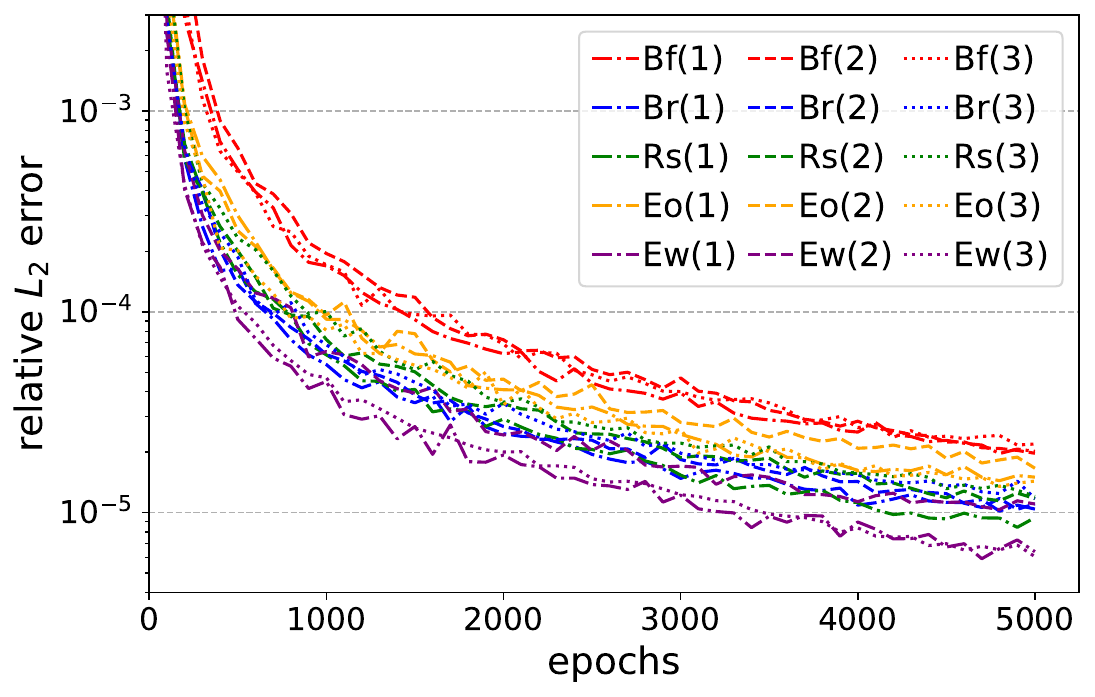}%
\label{fig:Lame-error1-history-ori}}
\vspace{-1mm}
\subfloat[Loss histories (case II)]{\includegraphics[width=2.0in]{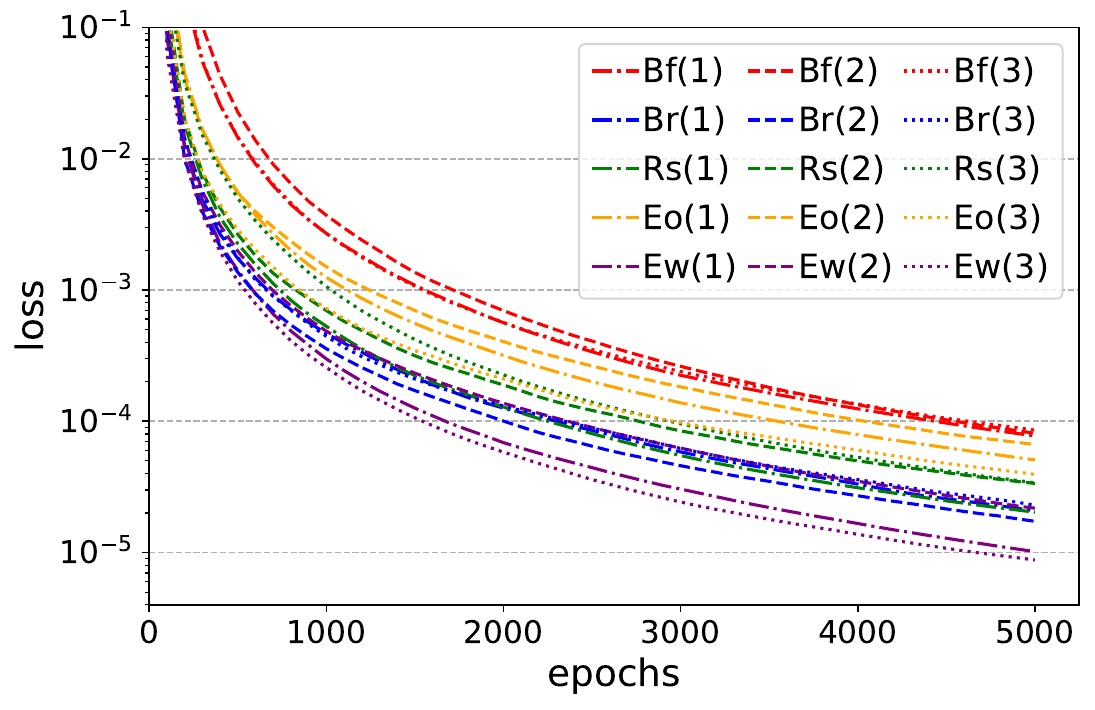}%
\label{fig:Lame-loss-history-genload}}
\hspace{3mm}
\subfloat[Error histories about $u$ (case II)]{\includegraphics[width=2.0in]{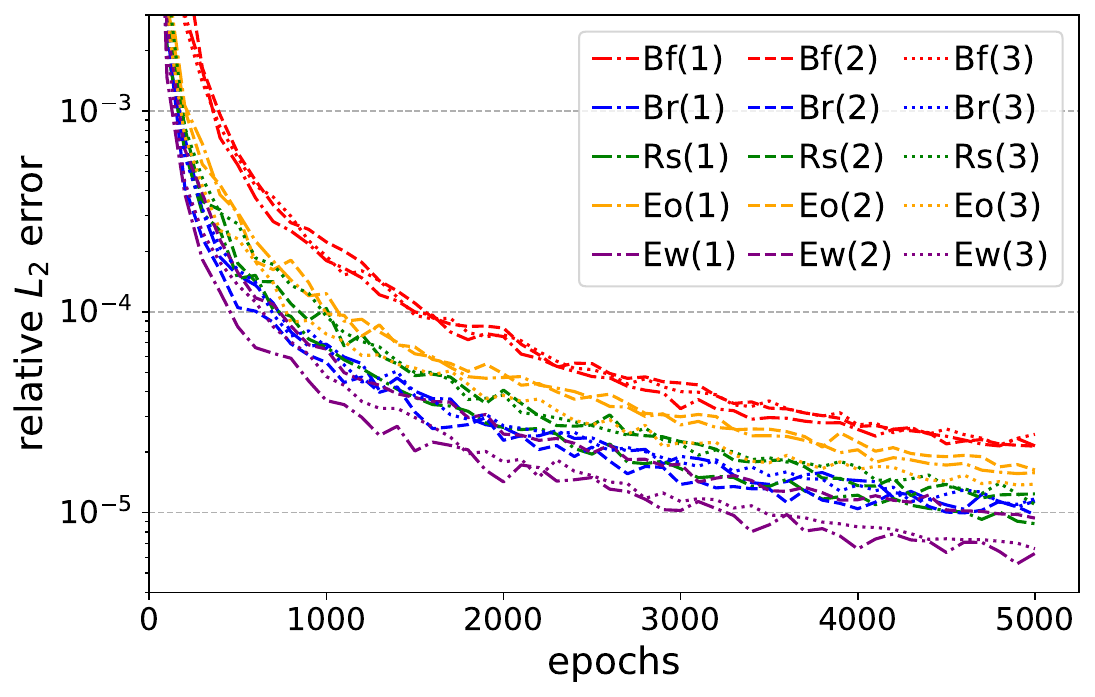}%
\label{fig:Lame-error0-history-genload}}
\hspace{3mm}
\subfloat[Error histories about $v$ (case II)]{\includegraphics[width=2.0in]{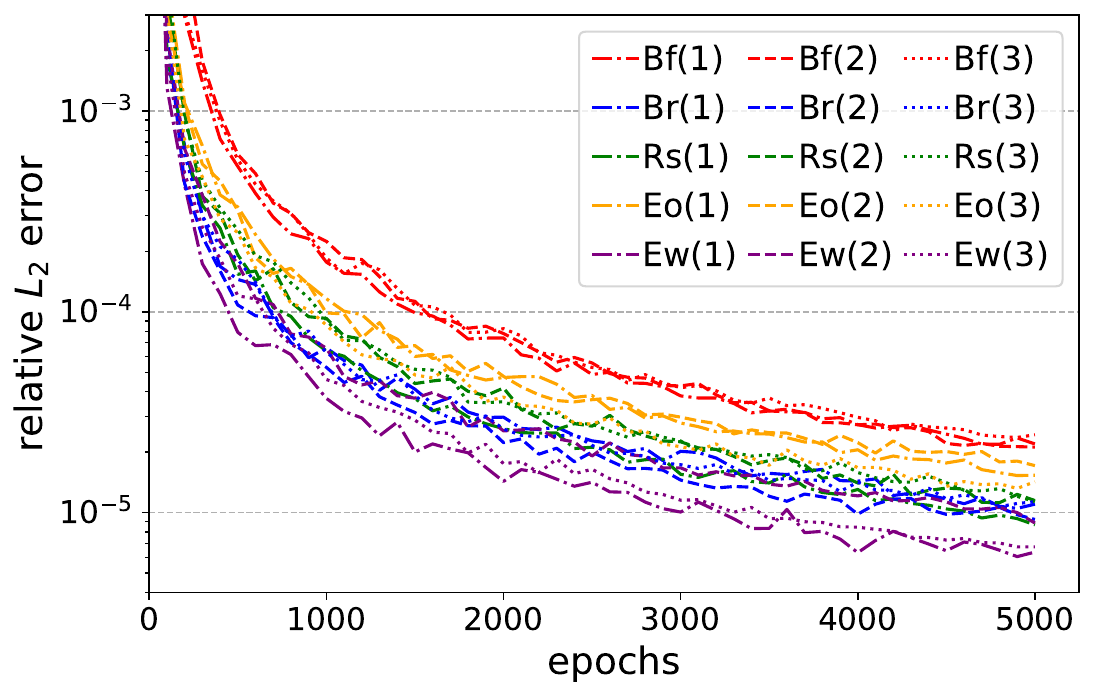}%
\label{fig:Lame-error1-history-genload}}
\vspace{-1mm}
\subfloat[Loss histories (case III)]{\includegraphics[width=2.0in]{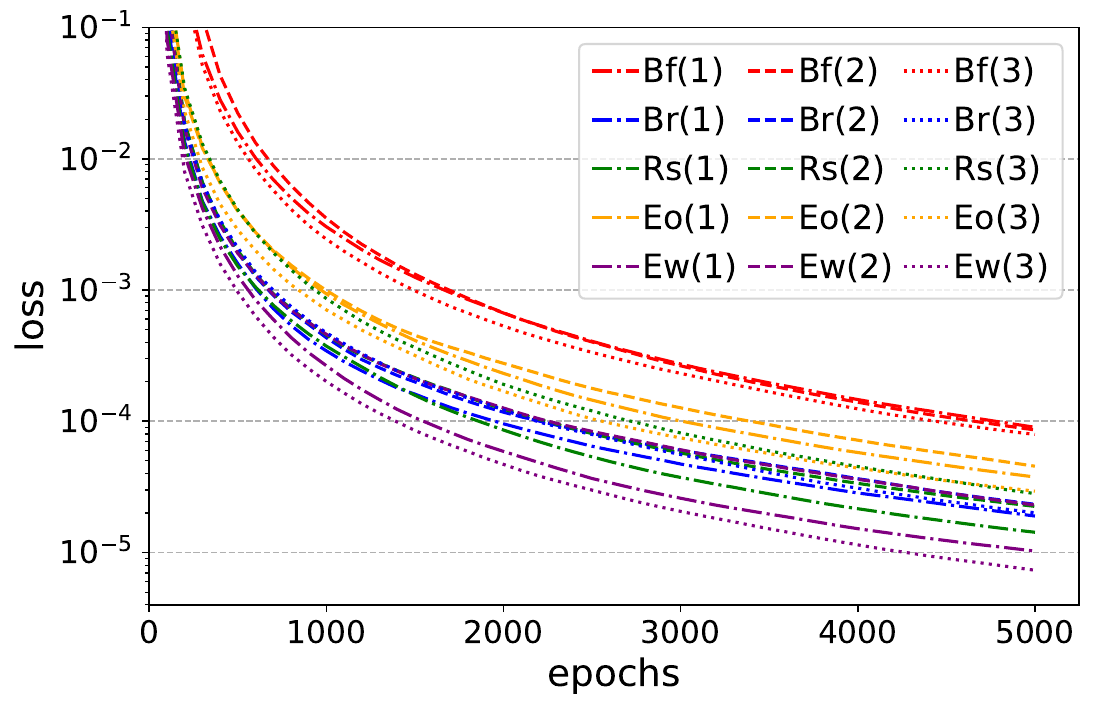}%
\label{fig:Lame-loss-history-genpara}}
\hspace{3mm}
\subfloat[Error histories about $u$ (case III)]{\includegraphics[width=2.0in]{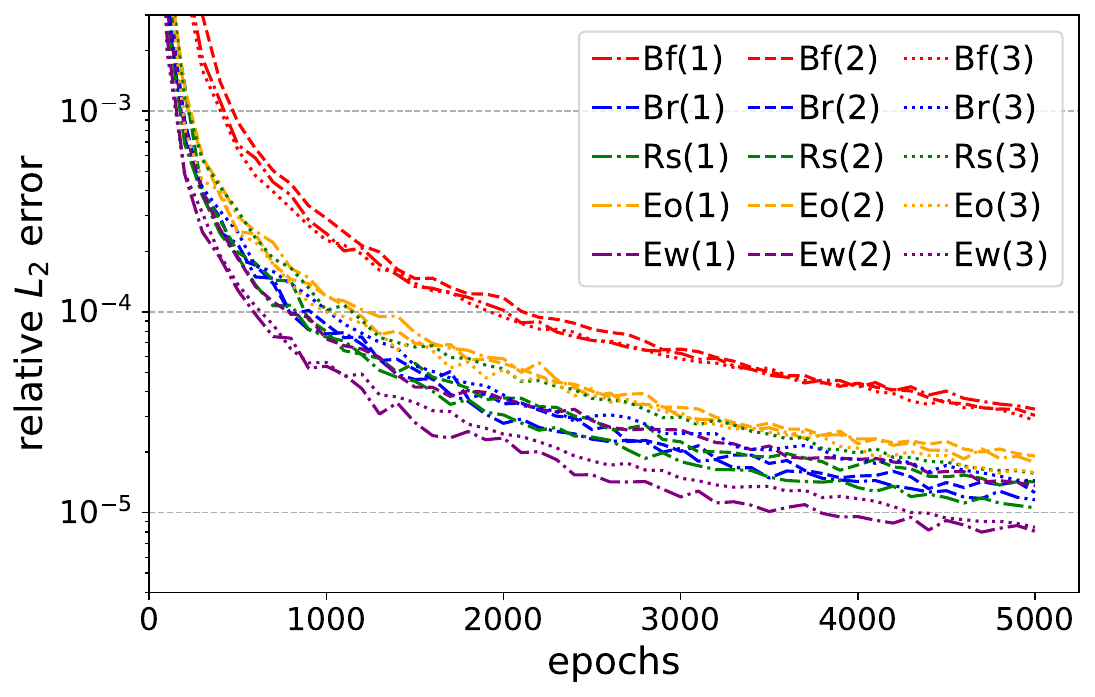}%
\label{fig:Lame-error0-history-genpara}}
\hspace{3mm}
\subfloat[Error histories about $v$ (case III)]{\includegraphics[width=2.0in]{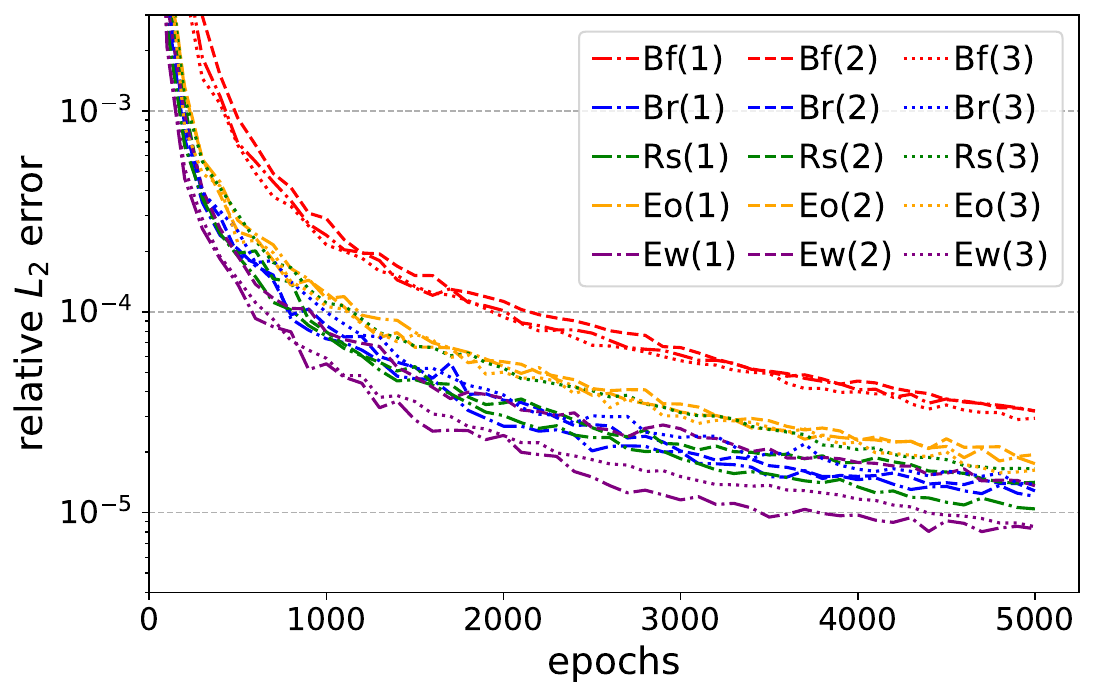}%
\label{fig:Lame-error1-history-genpara}}
\vspace{-1mm}
\subfloat[Loss histories (case IV)]{\includegraphics[width=2.0in]{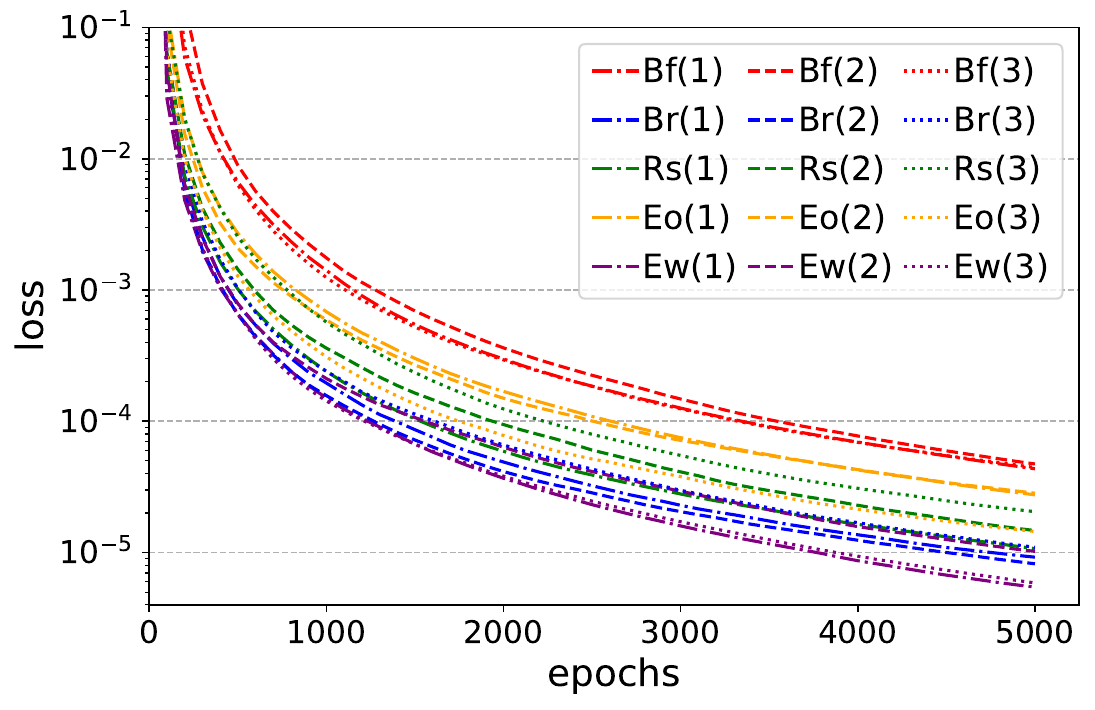}%
\label{fig:Lame-loss-history-genrad1}}
\hspace{3mm}
\subfloat[Error histories about $u$ (case IV)]{\includegraphics[width=2.0in]{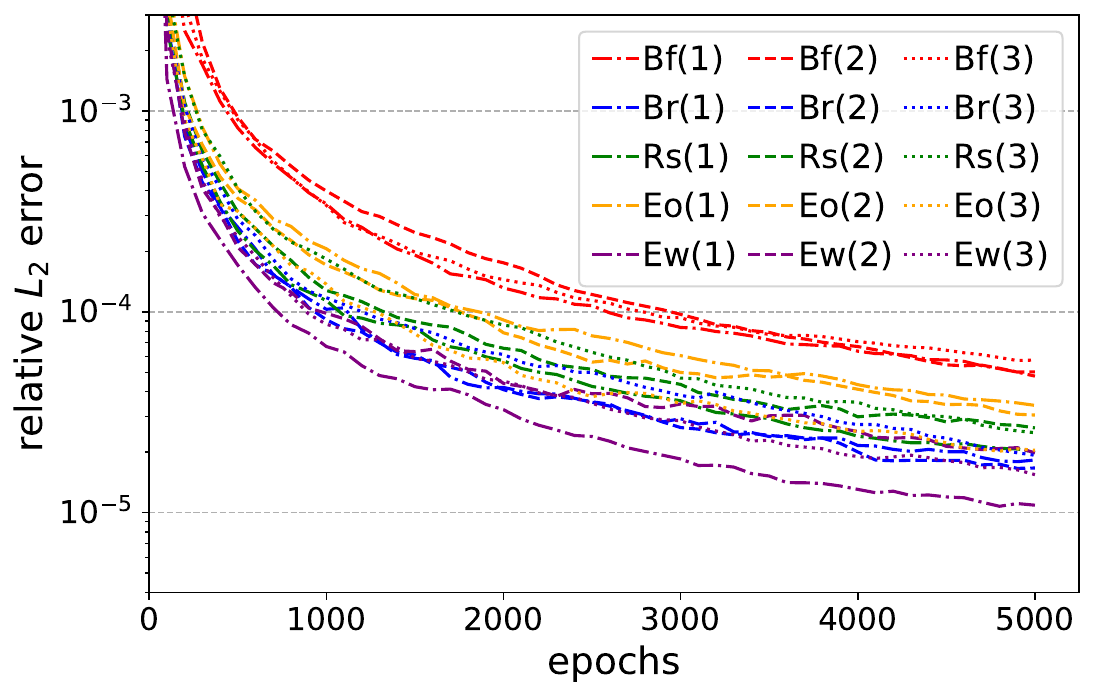}%
\label{fig:Lame-error0-history-genrad1}}
\hspace{3mm}
\subfloat[Error histories about $v$ (case IV)]{\includegraphics[width=2.0in]{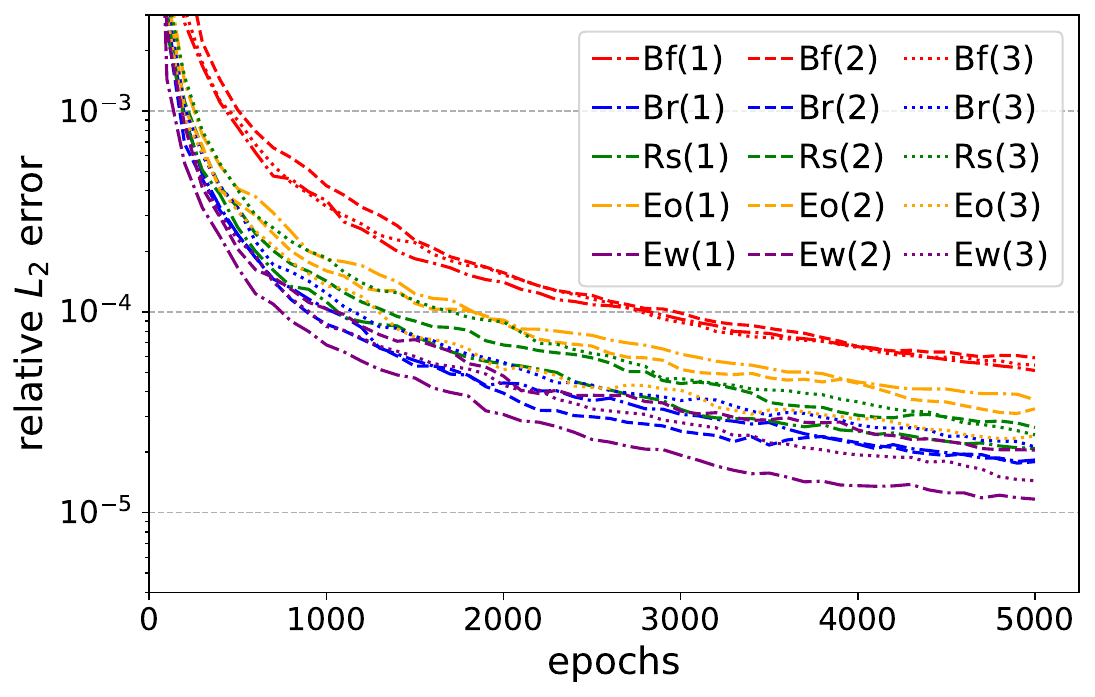}%
\label{fig:Lame-error1-history-genrad1}}
\vspace{-1mm}
\subfloat[Loss histories (case V)]{\includegraphics[width=2.0in]{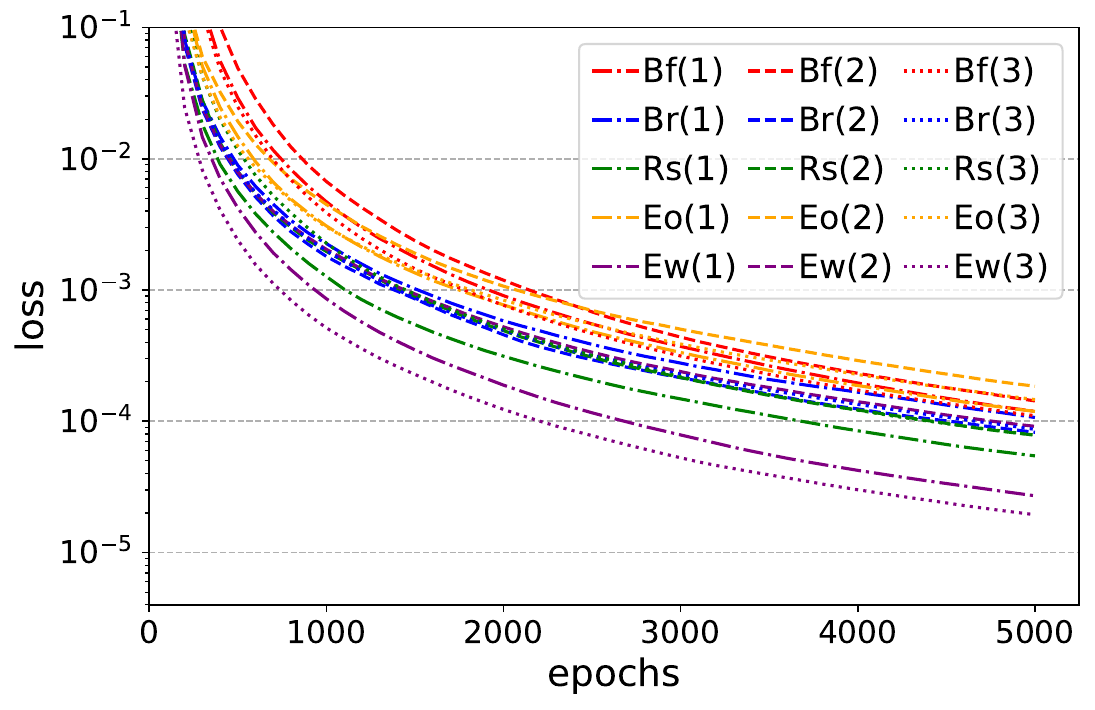}%
\label{fig:Lame-loss-history-genrad2}}
\hspace{3mm}
\subfloat[Error histories about $u$ (case V)]{\includegraphics[width=2.0in]{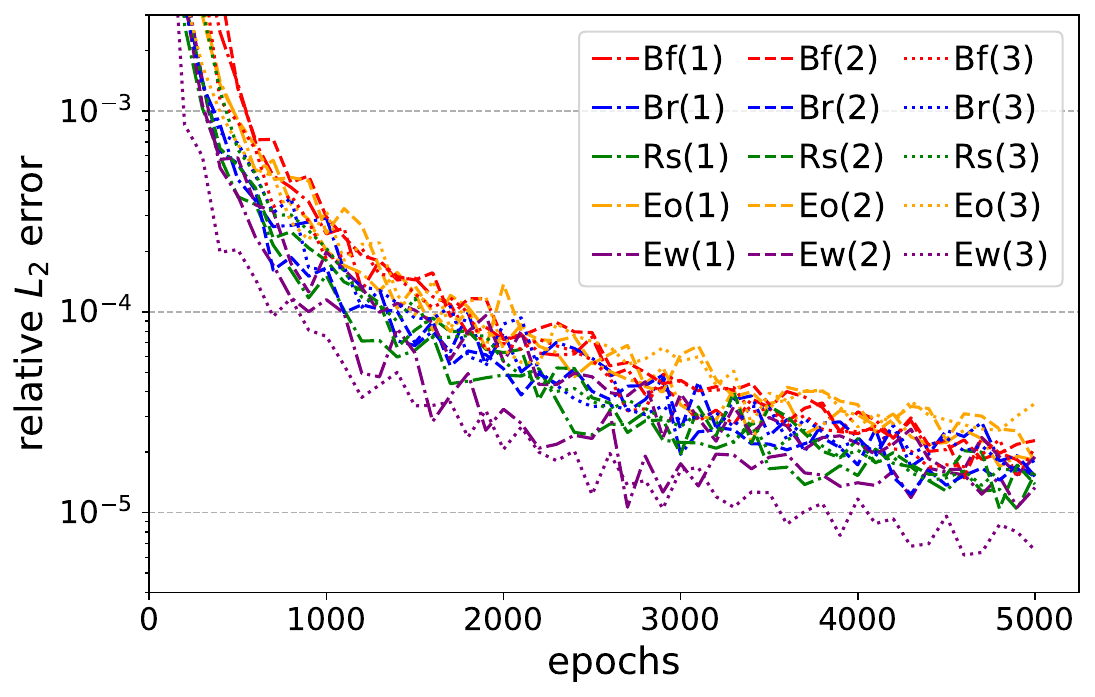}%
\label{fig:Lame-error0-history-genrad2}}
\hspace{3mm}
\subfloat[Error histories about $v$ (case V)]{\includegraphics[width=2.0in]{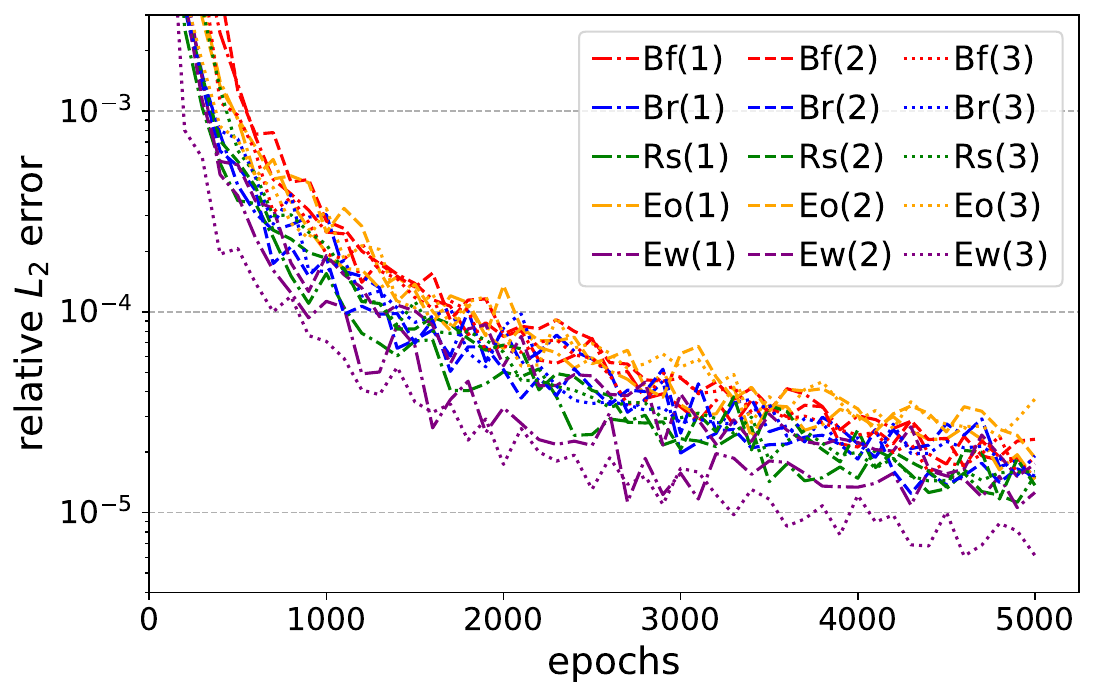}%
\label{fig:Lame-error1-history-genrad2}}
\caption{Convergence histories of loss and relative $L_2$ error in training the models discovered by each searching method for solving Lamé equations. (a), (b) and (c) show the loss histories, error histories about $u$ and $v$ in case I respectively. Similarly, (d), (e) and (f); (g), (h) and (i); (j), (k) and (l); (m), (n) and (o) are corresponding to cases II, III, IV and V respectively. Each curve of convergence history is the average result of ten independent trials.}
\label{fig:Lame_loss_error}
\vspace{-5mm}
\end{figure*}

In the last example, a plane strain problem in elastic mechanics is considered. The problem is discribed as a thick-walled cylinder subjected to the uniform pressure $q_1$ and shearing force $q_2$ loading on the outer boundary $\partial \Omega_o$, while the inner boundary $\partial \Omega_i$ remains fixed. This configuration is illustrated in Fig.~\ref{fig:annulus}, where $a$ and $b$ are the inner and outer radiuses of the cylinder. The equilibrium equations in this problem can be expressed in terms of displacement as Eq. \eqref{eq:Lame_govern}, which are referred to as Lamé equations.    

\vspace{-3mm}

\begin{figure*}[!htp]
\centering
\subfloat[Case I about $u$ (original): $E=2.1$, $\mu=0.25$, $q_1=23$, $q_2=3$, $a=1$, $b=2$]{\includegraphics[width=6.7in]{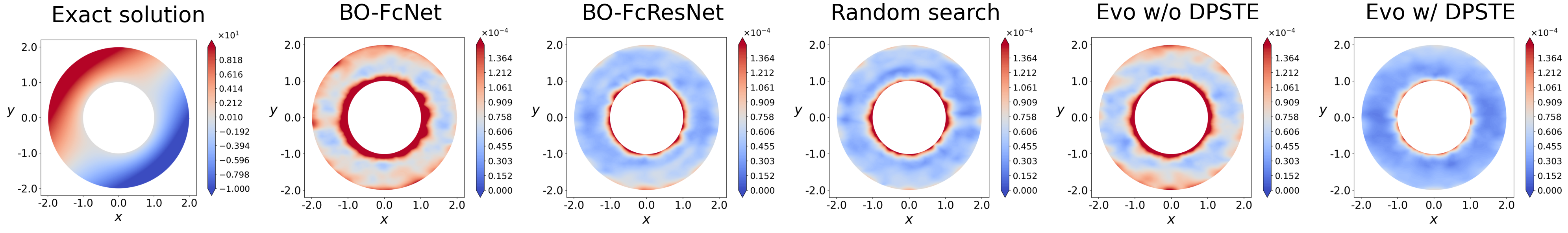}}
\vspace{-2mm}
\subfloat[Case II about $u$ (generalized): $E=2.1$, $\mu=0.25$, $q_1=30$, $q_2=2$, $a=1$, $b=2$]{\includegraphics[width=6.7in]{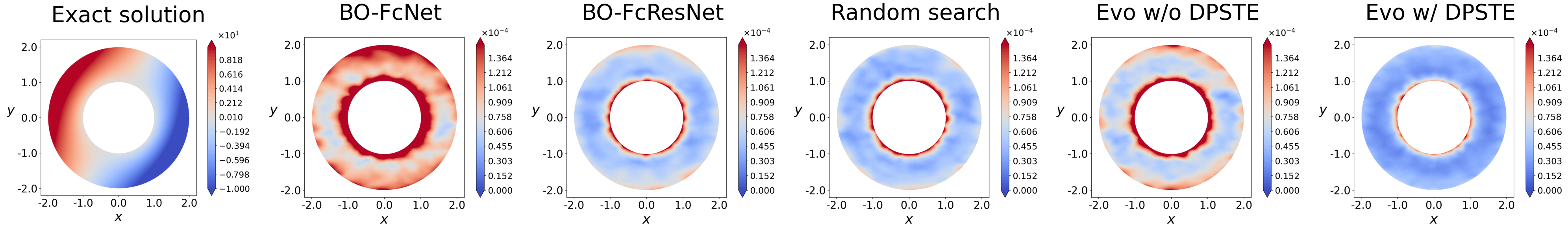}}
\vspace{-2mm}
\subfloat[Case III about $u$ (generalized): $E=3$, $\mu=0.3$, $q_1=23$, $q_2=3$, $a=1$, $b=2$]{\includegraphics[width=6.7in]{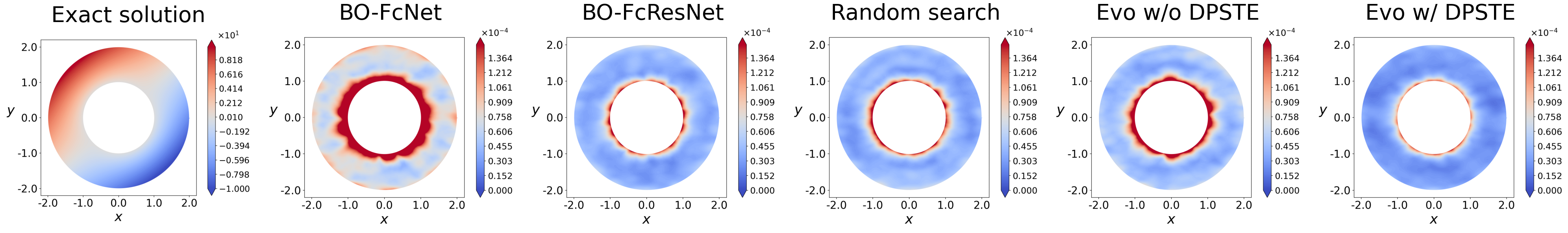}}
\vspace{-2mm}
\subfloat[Case IV about $u$ (generalized): $E=2.1$, $\mu=0.25$, $q_1=23$, $q_2=3$, $a=1.2$, $b=1.8$]{\includegraphics[width=6.7in]{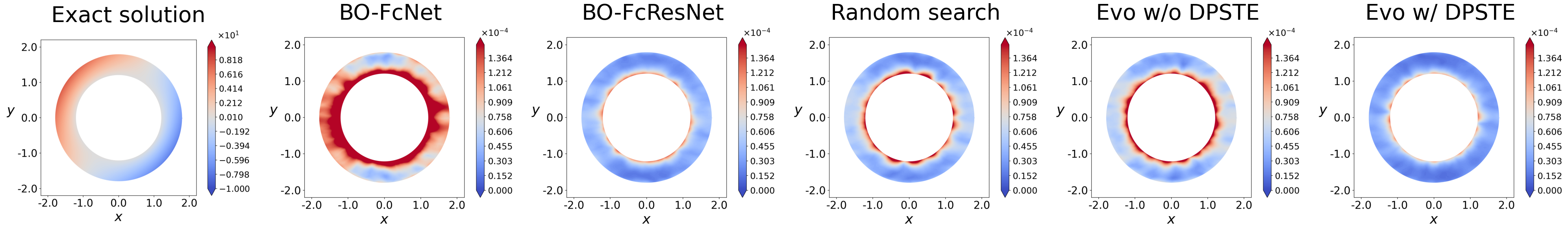}}
\vspace{-2mm}
\subfloat[Case V about $u$ (generalized): $E=2.1$, $\mu=0.25$, $q_1=23$, $q_2=3$, $a=0.8$, $b=2.2$]{\includegraphics[width=6.7in]{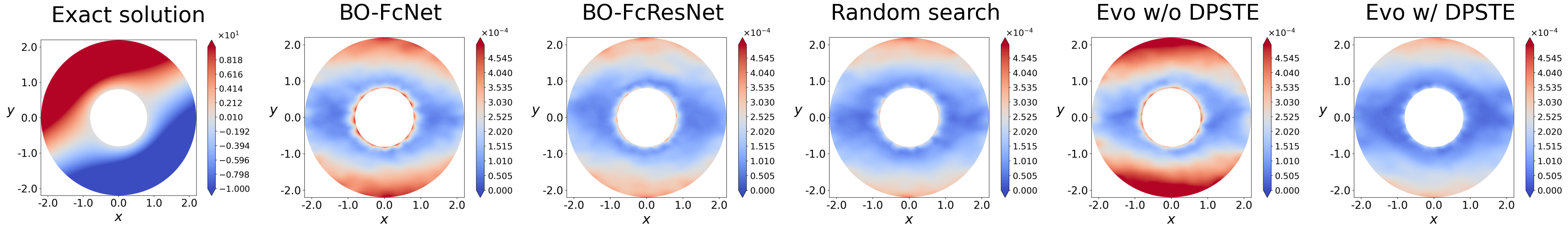}}
\caption{The exact solution (first column) of Lamé equations and absolute error (last five columns) about $u$ of the models discovered by each searching method. (a) Case I (original), (b) Case II (generalized), (c) Case III (generalized), (d) Case IV (generalized), and (e) Case V (generalized). The absolute error is the average result of the three models discovered by each searching method.}
\vspace{-3mm}
\label{fig:Lame_contourf_u}
\end{figure*}

\begin{equation}
\label{eq:Lame_govern}
\begin{aligned}
& \frac{E}{1-\mu^2}\left(u_{x x}+\frac{1-\mu}{2} u_{y y}+\frac{1+\mu}{2} v_{x y}\right)=0, \quad(x, y) \in \Omega,  \\
& \frac{E}{1-\mu^2}\left(v_{y y}+\frac{1-\mu}{2} v_{x x}+\frac{1+\mu}{2} u_{x y}\right)=0, \quad(x, y) \in \Omega, 
\end{aligned}
\end{equation}
where $u$ and $v$ represent the displacements in $x$-direction and $y$-direction, $E$ is the Young’s modulus, and $\mu$ is the Poisson’s ratio. The body force in this problem is ignored. The boundary conditions are given in Eq. \eqref{eq:Lame_boundary}.

\vspace{-4mm}

\begin{equation}
\label{eq:Lame_boundary}
\begin{aligned}
\frac{E}{1-\mu^2}\left[n_{1}\left(u_x+\mu v_y\right)+n_{2} \frac{1-\mu}{2}\left(u_y+v_x\right)\right]=h_{1}(x, y),&  \\
\quad(x, y)  \in \partial \Omega_o,& \\
\frac{E}{1-\mu^2}\left[n_{2}\left(v_y+\mu u_x\right)+n_{1} \frac{1-\mu}{2}\left(v_x+u_y\right)\right]=h_{2}(x, y),&  \\
\quad(x, y)  \in \partial \Omega_o,& \\
u(x, y)=0, \quad(x, y) \in \partial \Omega_i,& \\
v(x, y)=0, \quad(x, y) \in \partial \Omega_i,&
\end{aligned}
\end{equation}
where $(n_{1}, n_{2})\textsuperscript{T}$ is the unit outward normal vector on $\partial \Omega_o$.
According to the loads on $\partial \Omega_o$, $h_{1}(x, y)=-q_1 n_{1} + q_2 n_{2}$, and $h_{2}(x, y)=-q_1 n_{2} - q_2 n_{1}$. Due to the symmetry of the problem, the analytical solution can be derived as Eq. \eqref{eq:Lame_solution} through Airy stress function approach \cite{sadd2009elasticity}. 

\vspace{-3mm}

\begin{figure*}[!htp]
\centering
\subfloat[Case I about $v$ (original): $E=2.1$, $\mu=0.25$, $q_1=23$, $q_2=3$, $a=1$, $b=2$]{\includegraphics[width=6.7in]{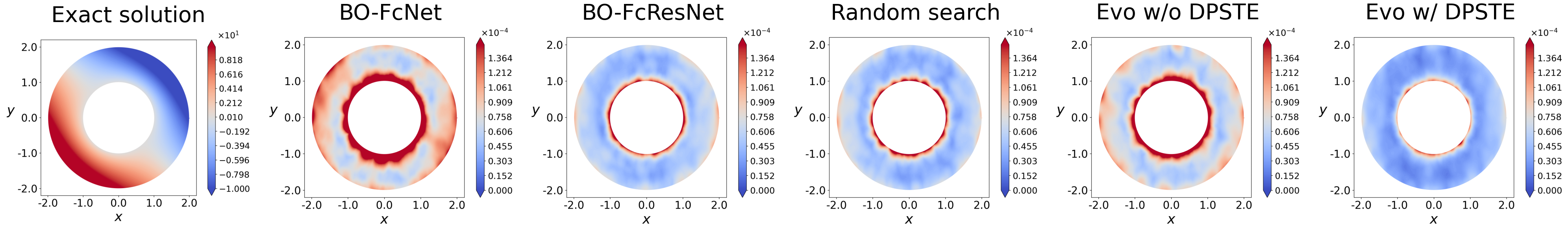}}
\vspace{-2mm}
\subfloat[Case II about $v$ (generalized): $E=2.1$, $\mu=0.25$, $q_1=30$, $q_2=2$, $a=1$, $b=2$]{\includegraphics[width=6.7in]{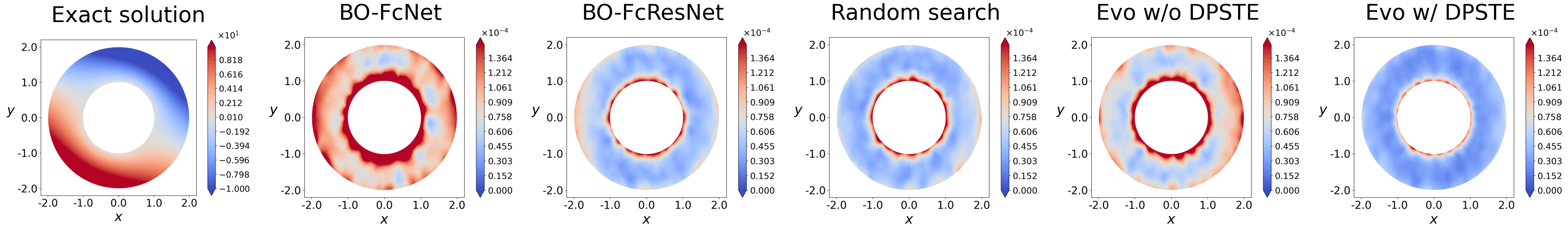}}
\vspace{-2mm}
\subfloat[Case III about $v$ (generalized): $E=3$, $\mu=0.3$, $q_1=23$, $q_2=3$, $a=1$, $b=2$]{\includegraphics[width=6.7in]{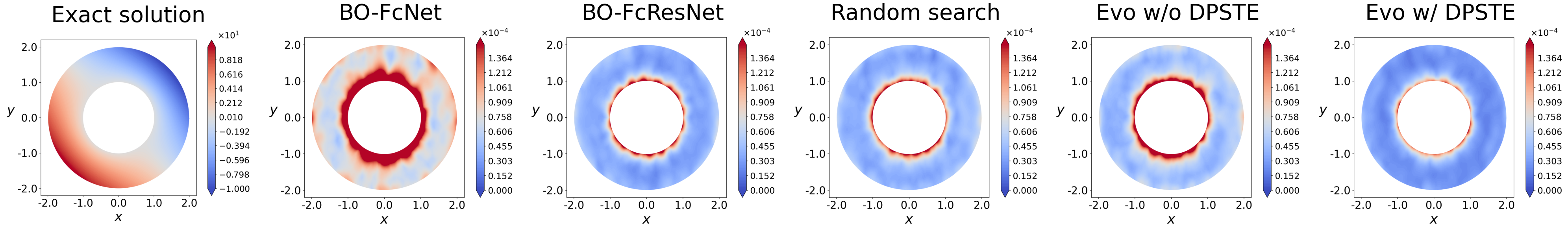}}
\vspace{-2mm}
\subfloat[Case IV about $v$ (generalized): $E=2.1$, $\mu=0.25$, $q_1=23$, $q_2=3$, $a=1.2$, $b=1.8$]{\includegraphics[width=6.7in]{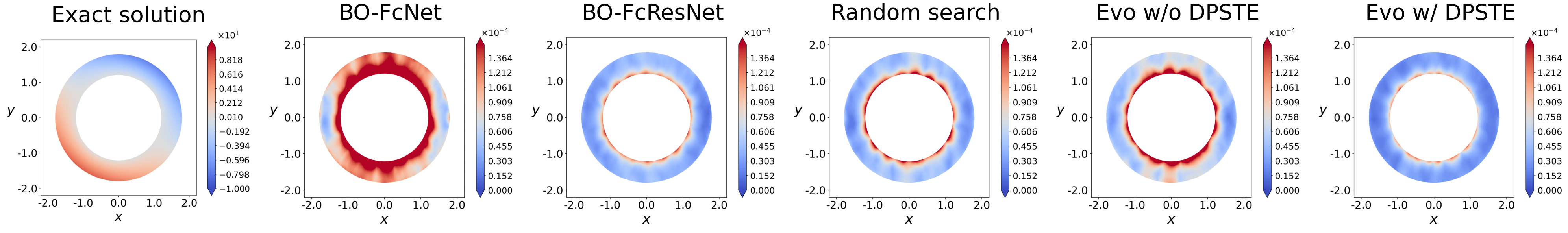}}
\vspace{-2mm}
\subfloat[Case V about $v$ (generalized): $E=2.1$, $\mu=0.25$, $q_1=23$, $q_2=3$, $a=0.8$, $b=2.2$]{\includegraphics[width=6.7in]{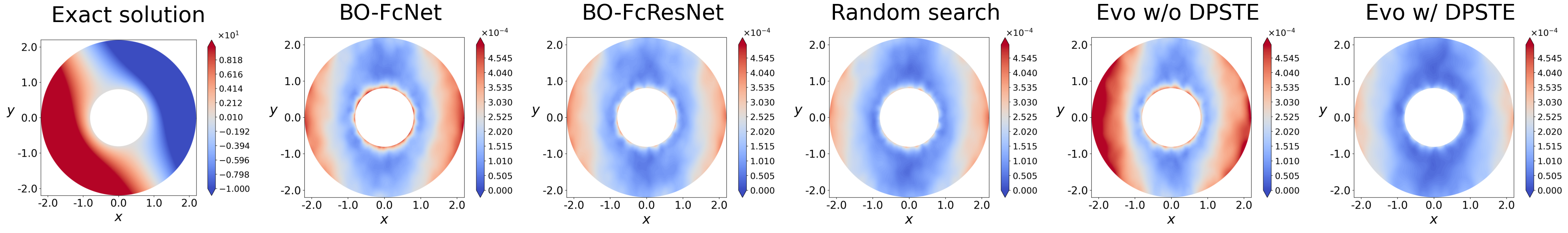}}
\caption{The exact solution (first column) of Lamé equations and absolute error (last five columns) about $v$ of the models discovered by each searching method. (a) Case I (original), (b) Case II (generalized), (c) Case III (generalized), (d) Case IV (generalized), and (e) Case V (generalized). The absolute error is the average result of the three models discovered by each searching method.}
\label{fig:Lame_contourf_v}
\vspace{-5mm}
\end{figure*}

\begin{equation}
\label{eq:Lame_solution}
\begin{aligned}
& u=A\left(\frac{a^2}{x^2+y^2}-1\right) x+B\left(1-\frac{a^2}{x^2+y^2}\right) y, \\
& v=A\left(\frac{a^2}{x^2+y^2}-1\right) y-B\left(1-\frac{a^2}{x^2+y^2}\right) x, \\
& A=\frac{1}{E} \frac{q_1\left(1-\mu^2\right) b^2}{b^2(1+\mu)+a^2(1-\mu)}, \\
& B=\frac{1}{E} \frac{q_2(1+\mu) b^2}{a^2}.
\end{aligned}
\vspace{-3mm}
\end{equation}

Each searching method is employed to discover the models for solving the problem with $E=2.1$, $\mu=0.25$, $q_1=23$, $q_2=3$, $a=1$ and $b=2$ (case I). And for evaluating the generalization capability of the models discovered in case I, different conditions are changed based on case I. At first, the loads on $\partial \Omega_o$ are changed to $q_1=30$, $q_2=2$ (case II). In case III, the material property coefficients are changed to $E=3$, $\mu=0.3$. For assessing the generalization capability of the models in terms of computational domain, the inner and outer radiuses are changed to $a=1.2$, $b=1.8$ (case IV) and $a=0.8$, $b=2.2$ (case V). The discovered models in case I are retrained from scratch for solving the problems about cases II, III, IV, V. In training the models based on PINNs, the numbers of inner and outer boundary points, collocation points, and testing points are listed in Table \ref{tab:Lame_points}. The penalty coefficient for boundary loss terms is set to 10. The number of training individuals and training epochs in Bayesian optimization, random search, evo-w/-DPSTE and evo-w/o-DPSTE are same with the experiment about Klein-Gordon equation. Sometimes, it is unnecessary to fully train the models, especially when the accuracy requirement is met in the practical application. In this experiment, only 5000 epochs are performed in training the discovered models.

\begin{table}[!htb]
\renewcommand\arraystretch{1.0}
\setlength\tabcolsep{4pt}
\caption{Number of sampling points.}
    \centering
    \begin{footnotesize}
    \begin{tabular}{p{2.2cm}<{\centering}p{1.1cm}<{\centering}p{1.1cm}<{\centering}p{1.1cm}<{\centering}p{1.1cm}<{\centering}}
    \toprule              Sampling domain  & $\partial \Omega_i$(train)  & $ \partial \Omega_o$(train)  & $\Omega$(train)  & $\overline\Omega$(test)  \\
    \midrule %\hline 
                          Case I/II/III  &  200  &  400  &  5740  &  23276   \\
                          Case IV        &  200  &  300  &  4232  &  17268   \\
                          Case V         &  200  &  550  &  6648  &  26944   \\    
    \bottomrule
    \end{tabular}
    \end{footnotesize} %
    \label{tab:Lame_points}
\vspace{-5mm}
\end{table}

Fig.~\ref{fig:Lame_loss_error} displays the convergence histories of loss and relative $L_2$ error during the training of models discovered by each searching method. In this figure, the convergence tendency of loss is similar with the relative $L_2$ error in each case. Despite the models discovered by Bayesian optimization based on FcResNet and random search show competitiveness, on the whole, the evo-w/-DPSTE models exhibit superior convergence rate. It also implies that the evo-w/-DPSTE models require fewer training epochs to achieve the same level of accuracy as other models, even for low accuracy. The exact solution and absolute error about $u$ and $v$ of the models discovered by each searching method are illustrated in Fig.~\ref{fig:Lame_contourf_u} and Fig.~\ref{fig:Lame_contourf_v}. In the original problem (case I) and generalized problems (cases II, III, IV, and V), evo-w/-DPSTE models intuitively demonstrate their advantage of approximation accuracy under incomplete training. Among the novel activation functions discovered by evo-w/-DPSTE, there are both complex composite function like atan($\alpha \cdot$ $\beta \cdot x \cdot$ sigmoid($\beta \cdot x$)) and simple function like $x$/($e^{x}+e^{-x}$), all of which exhibit commendable performance, as shown in Table~\ref{tab:Lame_appendix}. Due to the good generalization performance of evo-w/-DPSTE models, in practical application, the discovered model in one case can be employed in some other slightly changed cases without re-searching the model, especially when the computational resource is limited. The relative $L_2$ error, structure and activation function of each discovered model are exhibited detailedly in Appendix~\ref{sec:Appendix_discovered_model}. And the model structures and activation functions discovered by evo-w/-DPSTE are respectively visualized in Appendix~\ref{sec:Appendix_discovered_structure} and Appendix~\ref{sec:Appendix_discovered_activation}.

\vspace{-2mm}

\section{Conclusion and outlook}
An evolutionary computation method is proposed to discover superior model of PINNs in solving PDEs. This method simultaneously explores the model's structure and parametric activation function, offering a larger search space compared to the most existing methods, forming the foundation for discovering the superior models. The DPSTE strategy enhances model exploration and facilitates the discovery of fast convergence models. In experiments, the models discovered by the proposed evolutionary computation method (evo-w/-DPSTE) generally exhibit superior approximation accuracy and convergence rate compared to the ones obtained by Bayesian optimization, random search and evolution without DPSTE. Moreover, the evo-w/-DPSTE models generalize well across different conditions, implying that the discovered model can be employed in other similar cases without requiring a new search.

There are still some issues worth further research. (1) This work employs the physics-informed mean squared loss as the evaluation indicator of the model, while it may not always accurately reflect the model error. Exploring a more effective evaluation indicator is worthwhile. (2) The performance of the activation function is intricately linked to the distribution of initial parameters of neural network. However, the same initialization way of network parameters is used for different activation functions in this work, which may not give full play to the performance of the discovered activation functions. Developing an activation function-specific initialization way that is compatible with model searching is attractive. (3) The consuming time of training and model size can also be integrated into the fitness of evolution for meeting the needs of practical applications. (4) The proposed method may be extended to the inverse problems in the future.

\vspace{-3mm}

\section*{Acknowledgments}
We thank Garrett Bingham from University of Texas at Austin for meaningful discussion about searching parametric activation functions. This study was supported by the National Natural Science Foundation of China (No. 12131002, No. 62306018), the China Postdoctoral Science Foundation (No. 2022M710211) and high-performance computing platform of Peking University.

\printcredits

\appendix
\renewcommand\thetable{\Alph{section}\arabic{table}}

\vspace{-3mm}
\section{Appendix: Initialization details}
\label{sec:Appendix_initial}
In initializing the structure in random search and evolution, FcNet and FcResNet with regular and random shortcut connections are adopted. The regular shortcut connections within a FcResNet span the same number of layers, and the number is randomly set from 1 to 5 across different individuals. The number of random shortcut connections within a FcResNet should be less than the number of whole layers. In initializing the activation functions in random search and evolution, the commonly used 13 activation functions are contained, including tanh, atan, sin, cos, asinh, sigmoid, ReCU: (max(0, $x$))$^3$, fixed swish: $x$ $\cdot $ sigmoid($x$), parametric tanh: tanh($\alpha \cdot x$), parametric sin: sin($\alpha \cdot x$), parametric cos: cos($\alpha \cdot x$), parametric sigmoid: sigmoid($\alpha \cdot x$), and parametric swish: $x$ $\cdot $ sigmoid($\alpha \cdot x$). In the initial generation, the ratio of individuals using these 13 activation functions to those using random activation functions is set to 1:3. In experiments, Bayesian optimization searches the activation functions also from these 13 options.

\begin{table*}[!htp]
\fontsize{6.0}{0.0}\selectfont 
\renewcommand\arraystretch{-5.0}
\caption{The relative $L_2$ error, structure and activation function of each discovered model in solving Klein-Gordon equation.}
    \centering

    \begin{tabular}{p{1.8cm}<{\centering}p{0.4cm}<{\centering}p{3.5cm}<{\centering}|p{3.5cm}<{\centering}|p{3.5cm}<{\centering}} 
    \toprule Models &Run & Case I & Case II & Case III  \\
    \midrule %\hline 
    \multirow{6}{*}[-7.7ex]{BO-FcNet}    & \multirow{2}{*}{(1)}  &  1.38e-04$\pm$4.65e-05  &  1.35e-04$\pm$4.11e-05  &  1.66e-04$\pm$7.07e-05    \\
                           \cmidrule{3-5} %\hline 
                           && \multicolumn{3}{c}{layer num: 5; neuron num: 32; shortcuts: none; activation: sin($x$)}    \\
                           \cmidrule{2-5} %\hline 
                           & \multirow{2}{*}{(2)}  &  3.93e-04$\pm$4.13e-04  &  1.67e-04$\pm$8.24e-05  &  7.12e-04$\pm$1.15e-03    \\
                           \cmidrule{3-5} %\hline 
                           && \multicolumn{3}{c}{layer num: 5; neuron num: 48; shortcuts: none; activation: sin($\alpha \cdot x$)}     \\
                           \cmidrule{2-5} %\hline 
                           & \multirow{2}{*}{(3)}  &  1.74e-04$\pm$5.66e-05  &  1.30e-04$\pm$5.09e-05   &  2.34e-04$\pm$1.17e-04     \\
                           \cmidrule{3-5} %\hline 
                           && \multicolumn{3}{c}{layer num: 10; neuron num: 48; shortcuts: none; activation:  sin($\alpha \cdot x$)}    \\
    \midrule %\hline 
    \multirow{6}{*}[-7.7ex]{BO-FcResNet}    & \multirow{2}{*}{(1)}  &  1.58e-04$\pm$3.66e-05  &  1.33e-04$\pm$3.00e-05  &  2.12e-04$\pm$8.45e-05    \\
                           \cmidrule{3-5} %\hline 
                           && \multicolumn{3}{c}{layer num: 9; neuron num: 50; shortcuts: [0-2,2-4,4-6,6-8], activation: tanh($x$)}     \\
                           \cmidrule{2-5} %\hline 
                           & \multirow{2}{*}{(2)} &  1.38e-04$\pm$4.85e-05  &  1.27e-04$\pm$6.25e-05  &  1.35e-04$\pm$3.66e-05    \\
                           \cmidrule{3-5} %\hline 
                           && \multicolumn{3}{c}{layer num: 8; neuron num: 40; shortcuts: [0-2,2-4,4-6,6-7], activation: sin($x$)}     \\
                           \cmidrule{2-5} %\hline 
                           & \multirow{2}{*}{(3)}  &  1.23e-04$\pm$2.55e-05  &  1.36e-04$\pm$4.54e-05   &  1.47e-04$\pm$5.01e-05     \\
                           \cmidrule{3-5} %\hline 
                           && \multicolumn{3}{c}{layer num: 8; neuron num: 34; shortcuts: [0-2,2-4,4-6,6-7]; activation: sin($x$)}    \\
    \midrule %\hline                     
    \multirow{6}{*}[-7.7ex]{Random search}    & \multirow{2}{*}{(1)}  &  1.07e-04$\pm$2.47e-05  &  1.28e-04$\pm$3.52e-05  &  1.21e-04$\pm$2.35e-05    \\
                           \cmidrule{3-5} %\hline 
                           && \multicolumn{3}{c}{layer num: 6; neuron num: 32; shortcuts: none; activation: $\alpha \cdot$ cos($x$) $\cdot$ erf($\beta \cdot x$)}    \\
                           \cmidrule{2-5} %\hline 
                           & \multirow{2}{*}{(2)}  &  1.14e-04$\pm$3.34e-05  &  8.87e-05$\pm$2.12e-05  &  1.19e-04$\pm$2.38e-05    \\
                           \cmidrule{3-5} %\hline 
                           && \multicolumn{3}{c}{layer num: 8; neuron num: 40; shortcuts: [0-1,1-2,2-3,3-4,4-5,5-6]; activation: $\alpha \cdot$ sin($\beta \cdot x$) }     \\
                           \cmidrule{2-5} %\hline 
                           & \multirow{2}{*}{(3)}  &  1.25e-04$\pm$3.95e-05   &  1.02e-04$\pm$2.17e-05   &  1.25e-04$\pm$3.35e-05     \\
                           \cmidrule{3-5} %\hline 
                           && \multicolumn{3}{c}{layer num: 7; neuron num: 48; shortcuts: [0-1,1-2,2-3,4-5,5-6]; activation: sin($x$)}    \\
    \midrule %\hline 
    \multirow{6}{*}[-7.7ex]{Evo-w/o-DPSTE}    & \multirow{2}{*}{(1)}  &  1.96e-04$\pm$6.84e-05  &  1.57e-04$\pm$4.16e-05  &  2.37e-04$\pm$8.61e-05    \\
                           \cmidrule{3-5} %\hline
                           && \multicolumn{3}{c}{layer num: 6; neuron num: 48; shortcuts: [0-1]; activation: $\alpha \cdot$ tanh($x$)}    \\
                           \cmidrule{2-5} %\hline 
                           & \multirow{2}{*}{(2)}  &  1.39e-04$\pm$4.56e-05  &  1.35e-04$\pm$4.95e-05  &  1.72e-04$\pm$3.90e-05    \\
                           \cmidrule{3-5} %\hline 
                           && \multicolumn{3}{c}{layer num: 9; neuron num: 32; shortcuts: [0-1,1-3,3-4,4-5,5-7,7-8]; activation: $\alpha \cdot$ sin($x$)}     \\
                           \cmidrule{2-5} %\hline 
                           & \multirow{2}{*}{(3)}  &  1.01e-04$\pm$2.91e-05   &  1.24e-04$\pm$4.64e-05   &  \textbf{9.62e-05$\pm$2.77e-05}     \\
                           \cmidrule{3-5} %\hline 
                           && \multicolumn{3}{c}{layer num: 9; neuron num: 50; shortcuts: [0-1,1-2,2-3,3-4,4-5,5-6,6-7,7-8]; activation: $\alpha \cdot$ sin($\beta \cdot x$)}    \\
    \midrule %\hline                      
    \multirow{6}{*}[-7.7ex]{Evo-w/-DPSTE}    & \multirow{2}{*}{(1)}   &  \textbf{8.85e-05$\pm$3.40e-05}  &  8.29e-05$\pm$3.89e-05  &  1.10e-04$\pm$4.11e-05    \\
                           \cmidrule{3-5} %\hline
                           && \multicolumn{3}{c}{layer num: 6; neuron num: 48; shortcuts: [0-1,1-2,2-3,3-4,4-5]; activation: asinh($x$) $\cdot$ cos($x$) }    \\
                           \cmidrule{2-5} %\hline 
                           & \multirow{2}{*}{(2)}  &  9.07e-05$\pm$1.65e-05  &  8.59e-05$\pm$2.39e-05  &  1.11e-04$\pm$2.10e-05    \\
                           \cmidrule{3-5} %\hline
                           && \multicolumn{3}{c}{layer num: 9; neuron num: 44; shortcuts: [0-1,1-2,2-3,3-5,7-8]; activation: $\alpha \cdot$ tanh($\beta \cdot x$) $\cdot$ cos($x$)}     \\
                           \cmidrule{2-5} %\hline 
                           & \multirow{2}{*}{(3)}  &   1.18e-04$\pm$3.48e-05   &  \textbf{7.42e-05$\pm$1.67e-05}   &  1.20e-04$\pm$2.67e-05     \\
                           \cmidrule{3-5} %\hline 
                           && \multicolumn{3}{c}{layer num: 7; neuron num: 50; shortcuts: [0-1,1-2,2-4,4-5,5-6]; activation: $\alpha \cdot$ cos($x$) $\cdot \beta \cdot$ atan($x$) $\cdot$ sigmoid($\gamma \cdot x$)}    \\
    \bottomrule
    \end{tabular}

    % \end{scriptsize} %
    \label{tab:Klein-Gordon_appendix}

\vspace{-3mm}
\end{table*}

\begin{table*}[!htb]
\fontsize{6.0}{0.0}\selectfont 
\renewcommand\arraystretch{1.0}
\caption{The relative $L_2$ error, structure and activation function of each discovered model in solving Burgers equation.}
    \centering
    % \begin{footnotesize}
    \begin{tabular}{p{1.8cm}<{\centering}p{0.4cm}<{\centering}p{3.5cm}<{\centering}|p{3.5cm}<{\centering}|p{3.5cm}<{\centering}}
    \toprule Models &Run & Case I & Case II & Case III \\
    \midrule %\hline 
    \multirow{6}{*}[-7.7ex]{BO-FcNet} &\multirow{2}{*}{(1)}   &  1.20e-04$\pm$7.21e-05   &  8.71e-05$\pm$3.59e-05  &  1.55e-04$\pm$6.83e-05   \\
                                \cmidrule{3-5} %\hline 
                                && \multicolumn{3}{c}{layer num: 11; neuron num: 20; shortcuts: none; activation: $x \cdot $ sigmoid($x$)}    \\
                                \cmidrule{2-5} %\hline 
                               & \multirow{2}{*}{(2)}  &  1.00e-04$\pm$2.97e-05  &  8.63e-05$\pm$2.63e-05  &  1.15e-04$\pm$3.27e-05  \\
                               \cmidrule{3-5} %\hline 
                                && \multicolumn{3}{c}{layer num: 3; neuron num: 24; shortcuts: none; activation: tanh($\alpha \cdot x$)}     \\
                                \cmidrule{2-5} %\hline 
                               & \multirow{2}{*}{(3)}  &  8.19e-05$\pm$2.25e-05  &  8.65e-05$\pm$2.42e-05  &  8.61e-05$\pm$2.73e-05  \\
                               \cmidrule{3-5} %\hline 
                                && \multicolumn{3}{c}{layer num: 11; neuron num: 32; shortcuts: none; activation: sin($\alpha \cdot x$)}     \\
    \midrule %\hline 
    \multirow{6}{*}[-7.7ex]{BO-FcResNet} &\multirow{2}{*}{(1)}  &  1.04e-04$\pm$2.94e-05  &  8.25e-05$\pm$2.46e-05  &  1.14e-04$\pm$3.76e-05   \\
                                \cmidrule{3-5} %\hline
                                && \multicolumn{3}{c}{layer num: 3; neuron num: 34; shortcuts: [0-2]; activation: tanh($\alpha \cdot x$)}    \\
                                \cmidrule{2-5} %\hline 
                                & \multirow{2}{*}{(2)}  &  1.16e-04$\pm$2.64e-05  &  1.01e-04$\pm$2.04e-05  &  1.22e-04$\pm$3.24e-05   \\
                                 \cmidrule{3-5} %\hline 
                                && \multicolumn{3}{c}{layer num: 6; neuron num: 22; shortcuts: [0-2,2-4,4-5]; activation: tanh($\alpha \cdot x$)}     \\
                               \cmidrule{2-5} %\hline 
                               & \multirow{2}{*}{(3)}  &  8.92e-05$\pm$3.01e-05  &  7.77e-05$\pm$1.79e-05  &  9.16e-05$\pm$2.79e-05  \\
                               \cmidrule{3-5} %\hline 
                                && \multicolumn{3}{c}{layer num: 3; neuron num: 20; shortcuts: [0-2]; activation: tanh($\alpha \cdot x$)}     \\
    \midrule %\hline                     
    \multirow{6}{*}[-7.7ex]{Random search} &\multirow{2}{*}{(1)}  &  7.96e-05$\pm$1.73e-05  &  8.01e-05$\pm$1.49e-05  &  9.10e-05$\pm$1.64e-05   \\
                            \cmidrule{3-5} %\hline
                           && \multicolumn{3}{c}{layer num: 8; neuron num: 28; shortcuts: [0-4,4-6]; activation: atan($\alpha \cdot$ sin($x$))}    \\
                          \cmidrule{2-5} %\hline 
                          & \multirow{2}{*}{(2)}  &  7.95e-05$\pm$1.06e-05  &  6.34e-05$\pm$7.74e-06  &  1.02e-04$\pm$2.50e-05   \\
                          \cmidrule{3-5} %\hline 
                          && \multicolumn{3}{c}{layer num: 6; neuron num: 34; shortcuts: none; activation: sin($x \cdot $ sigmoid($x$))}     \\
                              \cmidrule{2-5} %\hline 
                               & \multirow{2}{*}{(3)}  &  1.34e-04$\pm$5.22e-05  &  1.23e-04$\pm$3.56e-05  &  1.20e-04$\pm$3.43e-05  \\
                               \cmidrule{3-5} %\hline 
                                && \multicolumn{3}{c}{layer num: 8; neuron num: 26; shortcuts: [0-1,1-2,2-3,3-4,4-5,6-7]; activation: $\alpha \cdot$ sigmoid($\beta \cdot x$)}     \\
    \midrule %\hline 
    \multirow{6}{*}[-7.7ex]{Evo-w/o-DPSTE} &\multirow{2}{*}{(1)}   &  7.63e-05$\pm$2.09e-05   &  8.18e-05$\pm$2.23e-05  &  8.03e-05$\pm$1.12e-05   \\
                            \cmidrule{3-5} %\hline
                             && \multicolumn{3}{c}{layer num: 8; neuron num: 22; shortcuts: [0-2]; activation: $ \alpha \cdot$ (-tanh($\beta \cdot x$))}    \\
                             \cmidrule{2-5} %\hline 
                             & \multirow{2}{*}{(2)}   &  1.55e-04$\pm$2.57e-04   &  9.12e-05$\pm$9.83e-05  &  1.92e-04$\pm$2.93e-04   \\
                             \cmidrule{3-5} %\hline 
                             && \multicolumn{3}{c}{layer num: 10; neuron num: 42; shortcuts: none; activation: $\alpha \cdot x$ $\cdot $ sigmoid($\alpha \cdot x$)}     \\
                                \cmidrule{2-5} %\hline 
                               & \multirow{2}{*}{(3)}  &  9.64e-05$\pm$1.30e-05  &  9.13e-05$\pm$2.31e-05  &  8.64e-05$\pm$2.05e-05  \\
                               \cmidrule{3-5} %\hline 
                                && \multicolumn{3}{c}{layer num: 6; neuron num: 32; shortcuts: [0-5]; activation: tanh($\alpha \cdot x$)}     \\
    \midrule %\hline                      
    \multirow{6}{*}[-7.7ex]{Evo-w/-DPSTE} &\multirow{2}{*}{(1)}  &  8.49e-05$\pm$1.59e-05  &   7.55e-05$\pm$1.92e-05  &  7.19e-05$\pm$1.42e-05  \\
                        \cmidrule{3-5} %\hline
                        && \multicolumn{3}{c}{layer num: 8; neuron num: 40; shortcuts: [3-5]; activation: sin($\alpha \cdot x$) $\cdot$ sigmoid($x$) }    \\
                        \cmidrule{2-5} %\hline 
                         & \multirow{2}{*}{(2)}  &  \textbf{6.27e-05$\pm$3.20e-05}  &  7.68e-05$\pm$4.81e-05  &  \textbf{4.00e-05$\pm$4.55e-05}  \\
                         \cmidrule{3-5} %\hline 
                          && \multicolumn{3}{c}{layer num: 5; neuron num: 20; shortcuts: [0-3]; activation: $\alpha \cdot$ sigmoid($\beta \cdot x$) }     \\
                              \cmidrule{2-5} %\hline 
                               & \multirow{2}{*}{(3)}  &  6.56e-05$\pm$1.45e-05  &  \textbf{5.93e-05$\pm$8.37e-06}  &  7.15e-05$\pm$1.32e-05  \\
                               \cmidrule{3-5} %\hline 
                                && \multicolumn{3}{c}{layer num: 8; neuron num: 42; shortcuts: [0-3]; activation: $\alpha \cdot$ asinh($\beta \cdot x $ $\cdot$ sigmoid($\beta \cdot x$)) }     \\
    \bottomrule
    \end{tabular}

    % \end{footnotesize} %
    \label{tab:Burgers_appendix}
 \vspace{-3mm}
\end{table*}

\begin{table*}[!htp]
\fontsize{6.0}{0.0}\selectfont 
\renewcommand\arraystretch{1.0}
\caption{The relative $L_2$ error, structure and activation function of each discovered model in solving Lamé equations.}  
    \centering
    % \begin{footnotesize}
    \begin{tabular}{p{1.8cm}<{\centering}p{0.4cm}<{\centering}p{0.4cm}<{\centering}p{2.0cm}<{\centering}|p{2.0cm}<{\centering}|p{2.0cm}<{\centering}|p{2.0cm}<{\centering}|p{2.0cm}<{\centering}}
    \toprule Models &Run & $u/v$ & Case I & Case II & Case III & Case IV & Case V \\
    \midrule %\hline 
    \multirow{9}{*}[-13.1ex]{BO-FcNet}   & \multirow{3}{*}[-1.6ex]{(1)}  & $u:$ &  2.00e-05$\pm$3.07e-06   &  2.15e-05$\pm$6.00e-06   &  3.27e-05$\pm$7.42e-06  &  4.78e-05$\pm$6.80e-06   &  1.53e-05$\pm$6.27e-06 \\
                          \cmidrule{3-8} %\hline  
                                                &      & $v:$         &  2.02e-05$\pm$4.50e-06   &  2.20e-05$\pm$5.93e-06   &  3.21e-05$\pm$4.50e-06  &  5.09e-05$\pm$1.01e-05   &  1.49e-05$\pm$4.67e-06 \\
                          \cmidrule{3-8} %\hline  
                          && \multicolumn{6}{c}{layer num: 7; neuron num: 50; shortcuts: none; activation: $x \cdot $ sigmoid($x$)}            \\
                          \cmidrule{2-8} %\hline  
                                       & \multirow{3}{*}[-1.6ex]{(2)}  & $u:$ &   1.88e-05$\pm$4.67e-06   &  2.14e-05$\pm$5.62e-06   &  3.04e-05$\pm$7.60e-06   &  5.02e-05$\pm$1.61e-05   &  2.28e-05$\pm$7.82e-06 \\
                          \cmidrule{3-8} %\hline 
                                                &      & $v:$       &  1.97e-05$\pm$3.36e-06   &  2.12e-05$\pm$4.94e-06   &  3.20e-05$\pm$6.15e-06   &  5.89e-05$\pm$1.78e-05   &  2.32e-05$\pm$7.90e-06 \\ 
                          \cmidrule{3-8} %\hline  
                          && \multicolumn{6}{c}{layer num: 10; neuron num: 50; shortcuts: none; activation: $x \cdot $ sigmoid($\alpha \cdot x$)}            \\
                          \cmidrule{2-8} %\hline  
                                       & \multirow{3}{*}[-1.6ex]{(3)}  & $u:$  &  2.26e-05$\pm$5.14e-06   &  2.45e-05$\pm$7.98e-06   &  2.88e-05$\pm$8.00e-06   &  5.74e-05$\pm$6.50e-06     &  2.11e-05$\pm$1.30e-05  \\
                          \cmidrule{3-8} %\hline 
                                                 &    & $v:$     &  2.20e-05$\pm$4.78e-06   &  2.44e-05$\pm$6.80e-06   &  2.95e-05$\pm$6.48e-06   &  5.42e-05$\pm$1.12e-05     &  1.90e-05$\pm$1.32e-05  \\
                          \cmidrule{3-8} %\hline  
                          && \multicolumn{6}{c}{layer num: 7; neuron num: 46; shortcuts: none; activation: $x \cdot $ sigmoid($\alpha \cdot x$)}            \\
    \midrule %\hline 
    \multirow{9}{*}[-13.1ex]{BO-FcResNet}   & \multirow{3}{*}[-1.6ex]{(1)}  & $u:$  &  1.00e-05$\pm$2.84e-06     &  1.14e-05$\pm$3.24e-06      &  1.16e-05$\pm$2.60e-06    &  1.83e-05$\pm$4.64e-06   &  1.51e-05$\pm$7.77e-06 \\
                          \cmidrule{3-8} %\hline 
                                                 &     & $v:$      &  1.04e-05$\pm$2.72e-06     &  1.10e-05$\pm$3.05e-06      &  1.20e-05$\pm$2.47e-06    &  1.83e-05$\pm$4.52e-06   &  1.42e-05$\pm$6.52e-06 \\ 
                          \cmidrule{3-8} %\hline 
                          && \multicolumn{6}{c}{layer num: 9; neuron num: 50; shortcuts: [0-2,2-4,4-6,6-8]; activation: tanh($\alpha \cdot x$)}   \\   
                          \cmidrule{2-8} %\hline 
                                         & \multirow{3}{*}[-1.6ex]{(2)}  & $u:$  &  1.06e-05$\pm$1.92e-06     &  9.74e-06$\pm$2.71e-06      &  1.26e-05$\pm$3.52e-06    &  1.67e-05$\pm$4.01e-06   &  1.51e-05$\pm$1.11e-05 \\
                          \cmidrule{3-8} %\hline 
                                                 &     & $v:$      &  1.11e-05$\pm$2.67e-06     &  9.24e-06$\pm$2.47e-06      &  1.28e-05$\pm$5.05e-06    &  1.79e-05$\pm$3.98e-06   &  1.52e-05$\pm$1.26e-05 \\
                          \cmidrule{3-8} %\hline  
                          && \multicolumn{6}{c}{layer num: 11; neuron num: 50; shortcuts: [0-2,2-4,4-6,6-8,8-10]; activation: tanh($\alpha \cdot x$)}   \\
                          \cmidrule{2-8} %\hline  
                                         & \multirow{3}{*}[-1.6ex]{(3)}  & $u:$  &  1.17e-05$\pm$2.52e-06    &  1.17e-05$\pm$2.59e-06      &  1.44e-05$\pm$2.75e-06    &  1.93e-05$\pm$3.90e-06   &  1.82e-05$\pm$1.08e-05  \\
                          \cmidrule{3-8} %\hline  
                                                 &     & $v:$      &  1.17e-05$\pm$3.02e-06    &  1.13e-05$\pm$2.27e-06      &  1.39e-05$\pm$2.41e-06    &  2.12e-05$\pm$5.75e-06   &  1.92e-05$\pm$9.78e-06  \\
                          \cmidrule{3-8} %\hline  
                          && \multicolumn{6}{c}{layer num: 11; neuron num: 42; shortcuts: [0-2,2-4,4-6,6-8,8-10]; activation: tanh($\alpha \cdot x$)}            \\
    \midrule %\hline                     
    \multirow{9}{*}[-13.1ex]{Random search}   & \multirow{3}{*}[-1.6ex]{(1)}  & $u:$  &  9.61e-06$\pm$2.74e-06    &  8.81e-06$\pm$2.52e-06   &  1.06e-05$\pm$2.22e-06    &  2.00e-05$\pm$4.09e-06    &  1.57e-05$\pm$1.01e-05 \\
                          \cmidrule{3-8} %\hline 
                                                &     & $v:$      &  9.44e-06$\pm$3.07e-06   &  8.69e-06$\pm$2.44e-06      &  1.04e-05$\pm$2.05e-06     &  2.08e-05$\pm$3.34e-06    &  1.52e-05$\pm$9.39e-06 \\
                          \cmidrule{3-8} %\hline  
                          && \multicolumn{6}{c}{layer num: 9; neuron num: 42; shortcuts: [0-1,1-2,2-3,3-4,4-5,5-6,6-7,7-8]; activation: erf($\alpha \cdot x$)  $\cdot$ sigmoid(erf($\alpha \cdot x$))}            \\
                          \cmidrule{2-8} %\hline  
                                             & \multirow{3}{*}[-1.6ex]{(2)}  & $u:$  &  1.09e-05$\pm$3.39e-06   &  1.24e-05$\pm$2.12e-06    &  1.42e-05$\pm$3.69e-06     &  2.64e-05$\pm$7.43e-06    &  1.38e-05$\pm$5.17e-06 \\
                          \cmidrule{3-8} %\hline   
                                                 &    & $v:$       &  1.17e-05$\pm$3.67e-06    &  1.15e-05$\pm$2.52e-06      &  1.42e-05$\pm$2.55e-06     &  2.65e-05$\pm$6.87e-06     &  1.37e-05$\pm$6.37e-06 \\ 
                          \cmidrule{3-8} %\hline  
                          && \multicolumn{6}{c}{layer num: 9; neuron num: 42; shortcuts: [0-2,2-4,4-6,6-8]; activation: min($\alpha \cdot$ ($e^{\beta \cdot x}+e^{-\beta \cdot x}$), atan($\gamma \cdot x$))}   \\
                          \cmidrule{2-8} %\hline  
                                             & \multirow{3}{*}[-1.6ex]{(3)}   & $u:$  &  1.21e-05$\pm$2.53e-06    &  1.10e-05$\pm$2.23e-06   &  1.57e-05$\pm$2.86e-06     &  2.50e-05$\pm$5.19e-06    &  1.55e-05$\pm$9.29e-06  \\
                          \cmidrule{3-8} %\hline  
                                                &     & $v:$        &  1.20e-05$\pm$2.61e-06    &  1.13e-05$\pm$2.17e-06   &  1.65e-05$\pm$3.45e-06     &  2.44e-05$\pm$7.48e-06    &  1.60e-05$\pm$9.52e-06  \\ 
                          \cmidrule{3-8} %\hline  
                          && \multicolumn{6}{c}{layer num: 9; neuron num: 26; shortcuts: [1-2,2-5,5-8]; activation: atan($x$) $\cdot$ (-sigmoid($x$)) }            \\
    \midrule %\hline 
    \multirow{9}{*}[-13.1ex]{Evo-w/o-DPSTE}   & \multirow{3}{*}[-1.6ex]{(1)}  & $u:$  &  1.56e-05$\pm$4.15e-06     &  1.58e-05$\pm$3.10e-06    &  1.78e-05$\pm$2.51e-06    &  3.42e-05$\pm$6.38e-06    &  1.82e-05$\pm$7.86e-06 \\
                          \cmidrule{3-8} %\hline 
                                                 &     & $v:$      &  1.50e-05$\pm$4.51e-06     &  1.53e-05$\pm$3.07e-06    &  1.75e-05$\pm$2.97e-06    &  3.64e-05$\pm$6.40e-06     &  1.46e-05$\pm$5.22e-06 \\
                          \cmidrule{3-8} %\hline 
                          && \multicolumn{6}{c}{layer num: 11; neuron num: 22; shortcuts: [0-2,2-3,3-4,4-5,5-6,6-7,7-8,8-9,9-10]; activation: $\alpha \cdot$ tanh($x \cdot $sigmoid($x$))}   \\  
                          \cmidrule{2-8} %\hline 
                                             & \multirow{3}{*}[-1.6ex]{(2)}  & $u:$   &  1.69e-05$\pm$3.76e-06     &  1.63e-05$\pm$5.28e-06    &  1.91e-05$\pm$9.30e-06     &  3.06e-05$\pm$7.53e-06    &  1.79e-05$\pm$1.36e-05 \\
                          \cmidrule{3-8} %\hline 
                                                &     & $v:$        &  1.66e-05$\pm$4.19e-06     &  1.71e-05$\pm$4.51e-06    &  1.95e-05$\pm$7.65e-06     &  3.29e-05$\pm$8.58e-06    &  1.88e-05$\pm$1.40e-05 \\
                          \cmidrule{3-8} %\hline 
                          && \multicolumn{6}{c}{layer num: 7; neuron num: 32; shortcuts: [0-1,1-3,3-4,4-5,5-6]; activation: $\alpha \cdot$ tanh($x$)}   \\
                          \cmidrule{2-8} %\hline  
                                            & \multirow{3}{*}[-1.6ex]{(3)}  & $u:$  &  1.37e-05$\pm$3.46e-06    &  1.39e-05$\pm$4.95e-06     &  1.58e-05$\pm$3.74e-06     &  2.04e-05$\pm$5.40e-06    &  3.49e-05$\pm$2.40e-05  \\
                          \cmidrule{3-8} %\hline  
                                               &     & $v:$         &  1.44e-05$\pm$4.20e-06    &  1.42e-05$\pm$5.61e-06      &  1.63e-05$\pm$4.06e-06     &  2.41e-05$\pm$8.32e-06   &  3.68e-05$\pm$2.17e-05  \\ 
                          \cmidrule{3-8}
                          && \multicolumn{6}{c}{layer num: 7; neuron num: 50; shortcuts: [0-2,2-4,4-6]; activation: tanh($\alpha \cdot x$)}            \\
    \midrule %\hline                      
    \multirow{9}{*}[-13.1ex]{Evo-w/-DPSTE}   & \multirow{3}{*}[-1.6ex]{(1)}  & $u:$  &  6.18e-06$\pm$1.74e-06  &  \textbf{6.26e-06$\pm$2.23e-06}  &  \textbf{8.07e-06$\pm$1.37e-06}  &  \textbf{1.09e-05$\pm$1.30e-06}  &  1.33e-05$\pm$6.27e-06 \\
                          \cmidrule{3-8} %\hline 
                                              &     & $v:$          &  6.39e-06$\pm$1.50e-06   &  \textbf{6.36e-06$\pm$2.10e-06}         &  \textbf{8.31e-06$\pm$1.53e-06}            &  \textbf{1.17e-05$\pm$2.18e-06}          &  1.26e-05$\pm$6.29e-06 \\
                          \cmidrule{3-8} %\hline 
                          && \multicolumn{6}{c}{layer num: 10; neuron num: 42; shortcuts: [0-1,1-2,2-3,3-4,4-5,5-6,6-7,7-9]; activation: $x$/($e^{x}+e^{-x}$)}    \\
                          \cmidrule{2-8} %\hline 
                                            & \multirow{3}{*}[-1.6ex]{(2)}  & $u:$  &  1.11e-05$\pm$2.23e-06    &  9.39e-06$\pm$2.07e-06     &  1.36e-05$\pm$3.98e-06      &  1.98e-05$\pm$6.63e-06     &  1.89e-05$\pm$7.06e-06 \\
                          \cmidrule{3-8} %\hline 
                                             &      & $v:$          &  1.10e-05$\pm$2.07e-06    &  8.87e-06$\pm$1.75e-06     &  1.36e-05$\pm$4.36e-06      &  2.05e-05$\pm$4.91e-06     &  1.75e-05$\pm$7.71e-06 \\
                          \cmidrule{3-8} %\hline 
                          && \multicolumn{6}{c}{layer num: 8; neuron num: 48; shortcuts: [0-1,1-3,3-4,4-6,6-7]; activation: sin(tanh($\alpha \cdot x$))}   \\
                          \cmidrule{2-8} %\hline  
                                            & \multirow{3}{*}[-1.6ex]{(3)}   & $u:$  &  \textbf{6.05e-06$\pm$1.09e-06}   &  6.61e-06$\pm$1.55e-06    &  8.44e-06$\pm$1.54e-06    &  1.54e-05$\pm$2.00e-06   &  \textbf{6.49e-06$\pm$2.30e-06}  \\
                          \cmidrule{3-8} %\hline  
                                             &      & $v:$          &  \textbf{6.04e-06$\pm$6.70e-07}    &  6.75e-06$\pm$1.60e-06      &  8.50e-06$\pm$1.42e-06      &  1.44e-05$\pm$1.45e-06     &  \textbf{6.12e-06$\pm$2.27e-06}  \\ 
                          \cmidrule{3-8} %\hline  
                          && \multicolumn{6}{c}{layer num: 11; neuron num: 48; shortcuts: [0-1,1-2,2-3,3-4,4-6,6-7,7-8,8-9,9-10]; activation: atan($\alpha \cdot$ $\beta \cdot x \cdot$ sigmoid($\beta \cdot x$)) }            \\
    \bottomrule  
    \end{tabular}

    % \end{footnotesize} %
    \label{tab:Lame_appendix}
\vspace{0mm}
\end{table*}

\begin{figure*}[!htp]
\centering
\subfloat[Experiment 4.1 (Klein-Gordon equation)]{\includegraphics[width=3.3in]{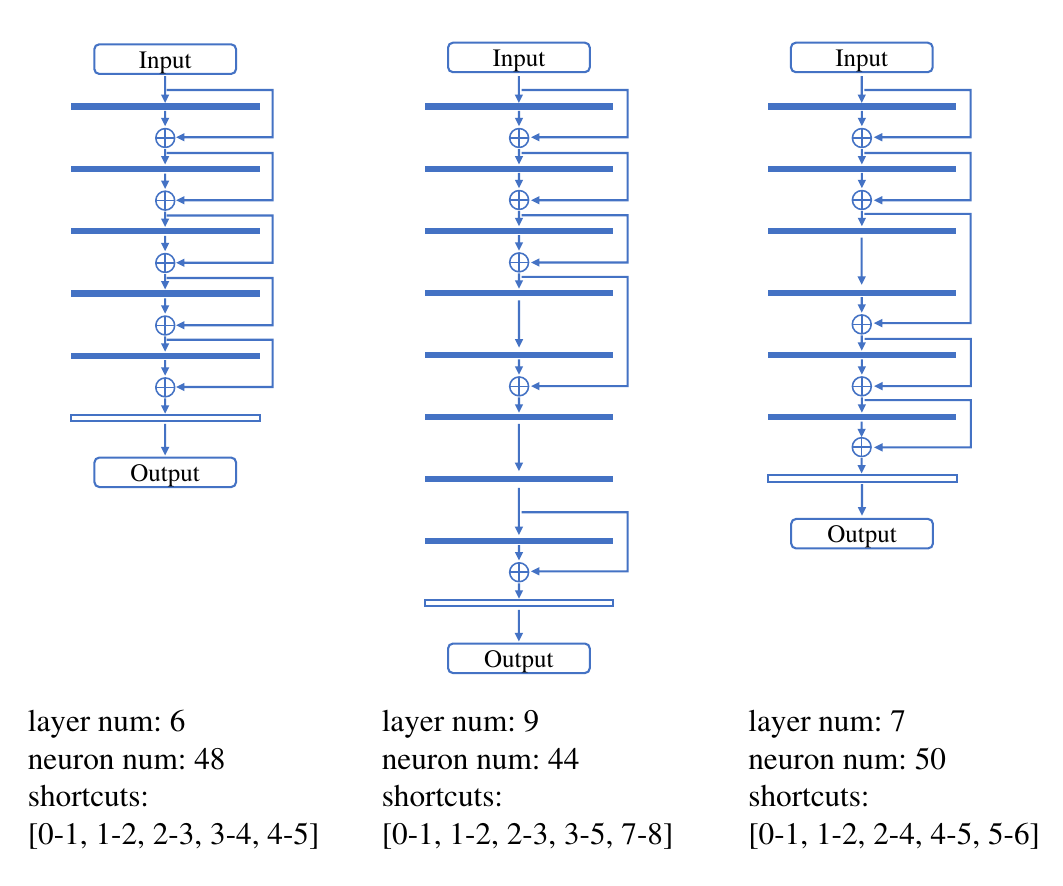}
\label{fig:Klein_str}}
\hspace{4.1mm}
\subfloat[Experiment 4.2 (Burgers equation)]{\includegraphics[width=3.3in]{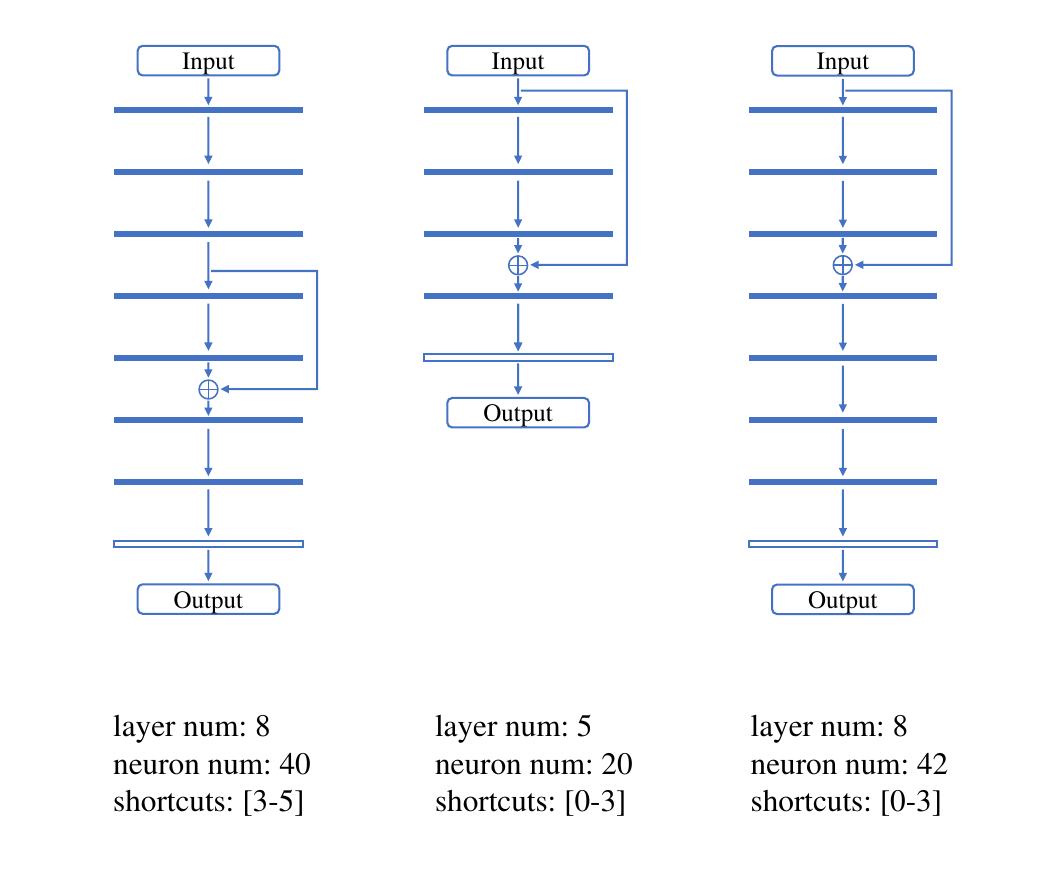}%
\label{fig:Burgers_str}}
\vspace{2mm}
\subfloat[Experiment 4.3 (Lamé equations)]{\includegraphics[width=5.5in]{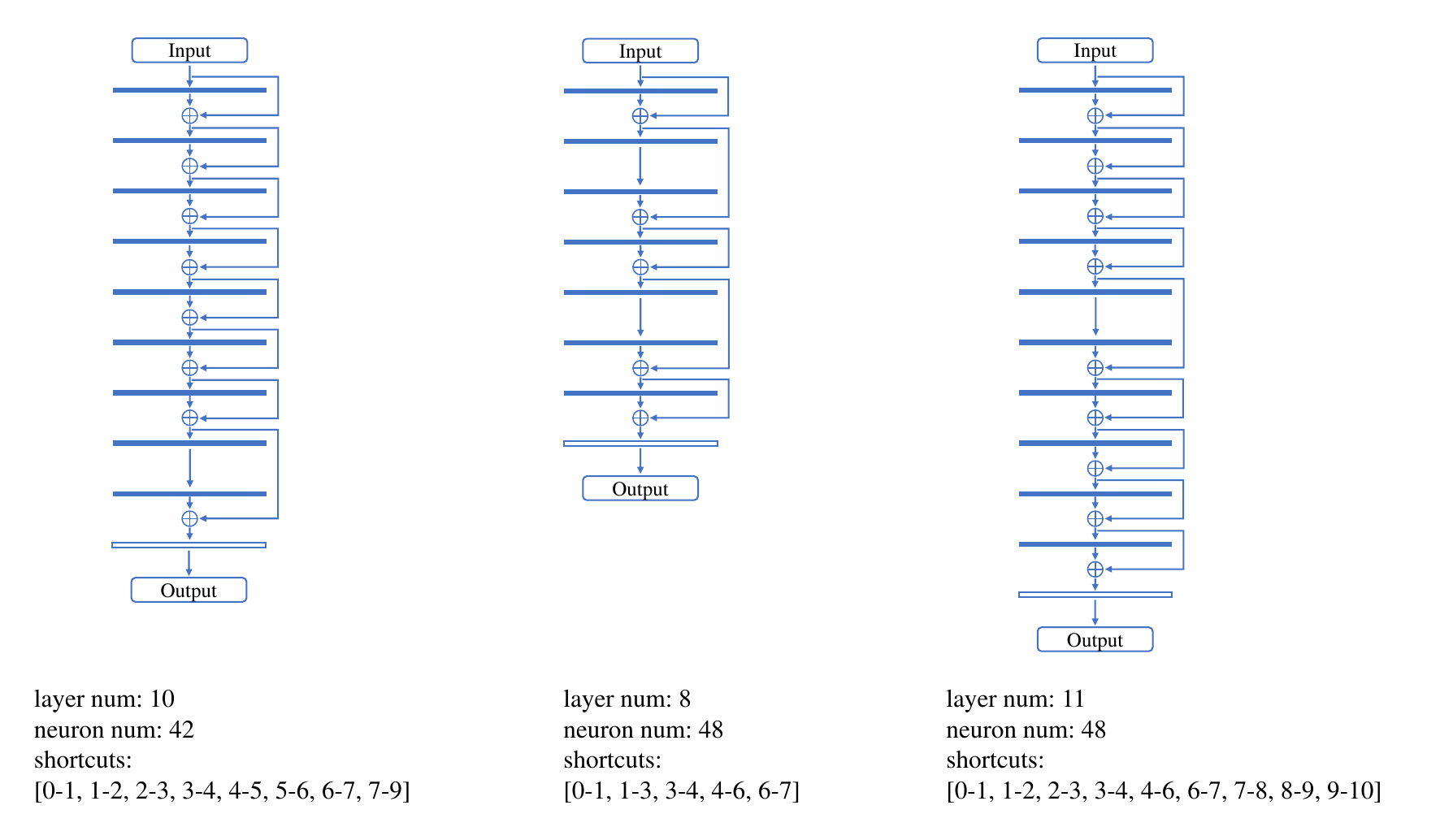}%
\label{fig:Lame_str}}
\caption{The model structures discovered though evo-w/-DPSTE in the three experiments.}
\label{fig:discovered_str}
\vspace{0mm}
\end{figure*}

\begin{figure*}[!htp]
\centering
\subfloat[tanh($\alpha \cdot x$)]{\includegraphics[width=3.4in]{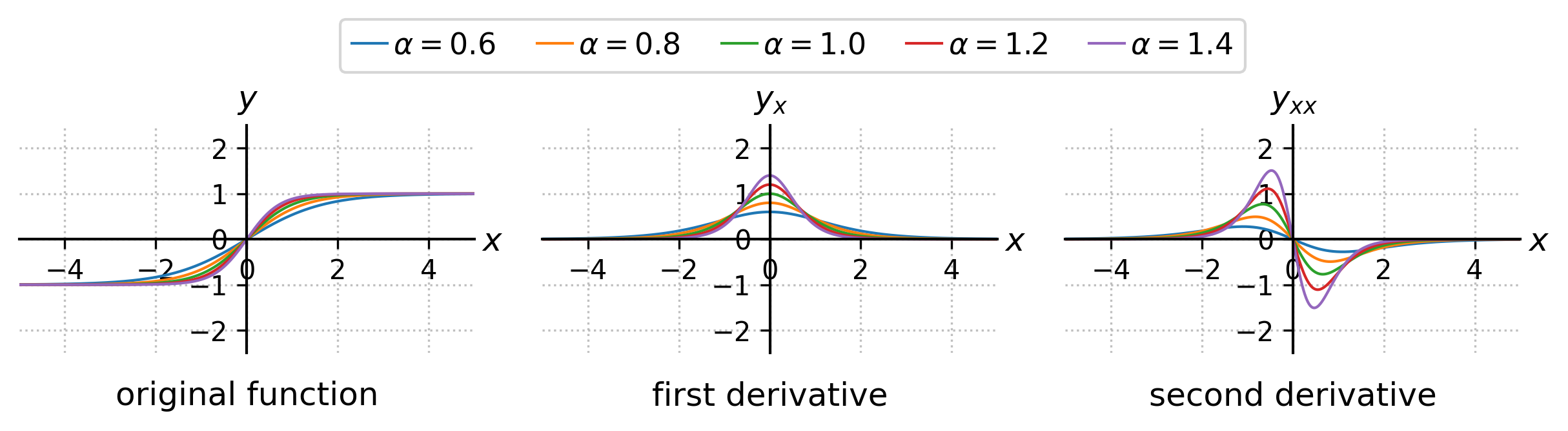}%
\label{fig:tanh_L1}}
\hspace{0.05mm}
\subfloat[sin($\alpha \cdot x$)]{\includegraphics[width=3.4in]{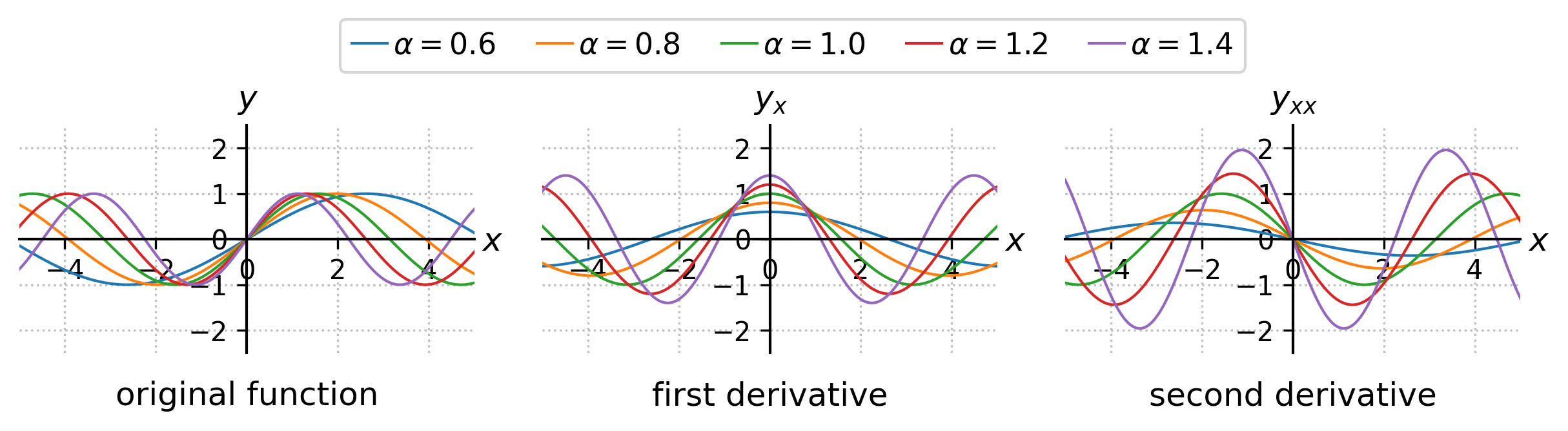}%
\label{fig:sin_L1}}
\vspace{-0.3mm}
\subfloat[sigmoid($\alpha \cdot x$)]{\includegraphics[width=3.4in]{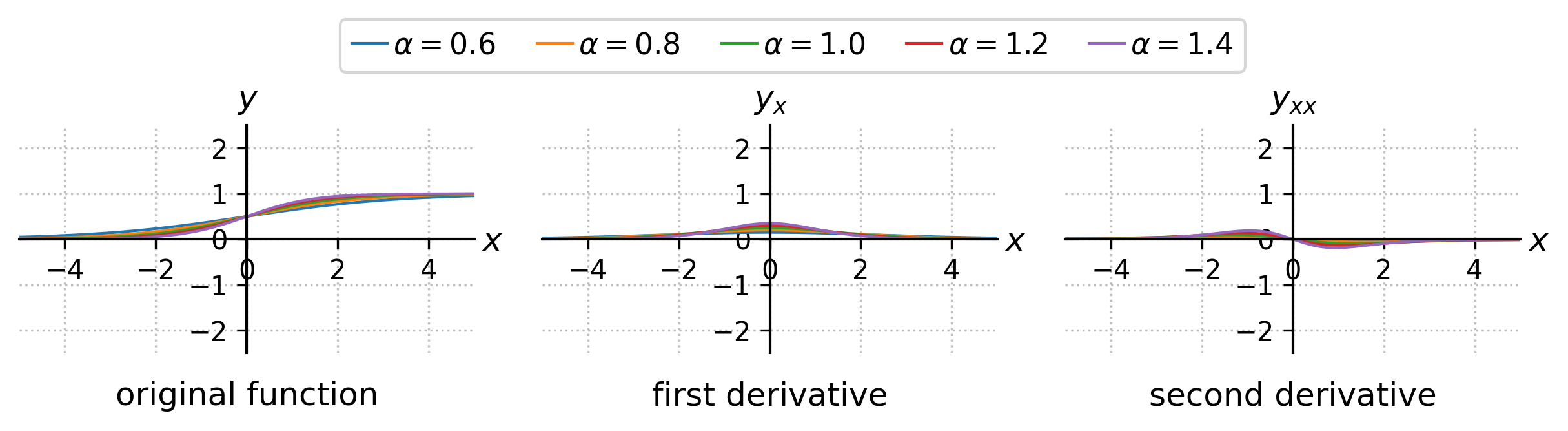}%
\label{fig:sigmoid_L1}}
\hspace{0.05mm}
\subfloat[asinh($x$) $\cdot$ cos($x$)]{\includegraphics[width=3.4in]{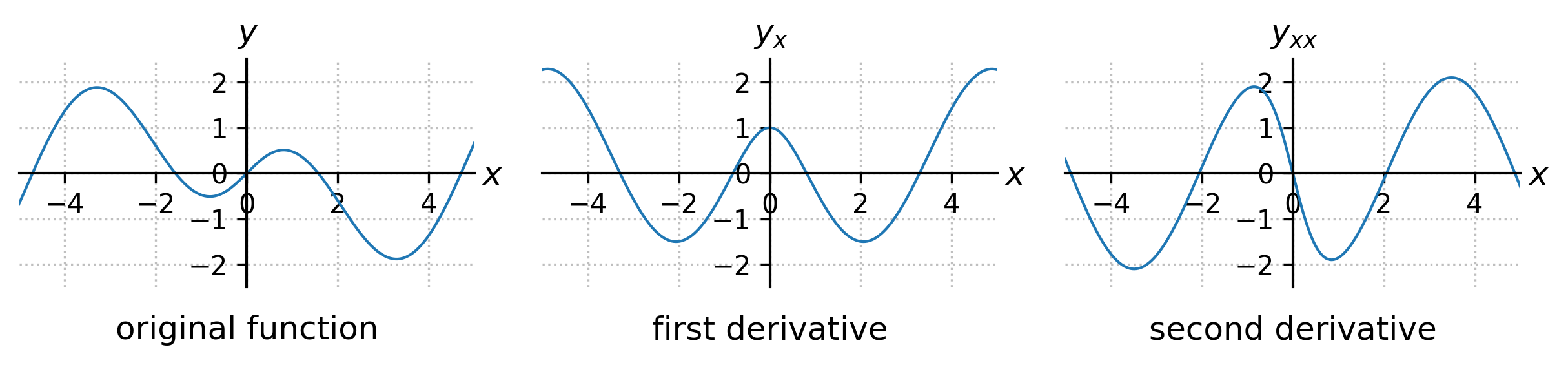}%
\label{fig:asinh_X_cos}}
\vspace{-0.3mm}
\subfloat[$\alpha \cdot$ tanh($\beta \cdot x$) $\cdot$ cos($x$)]{\includegraphics[width=3.4in]{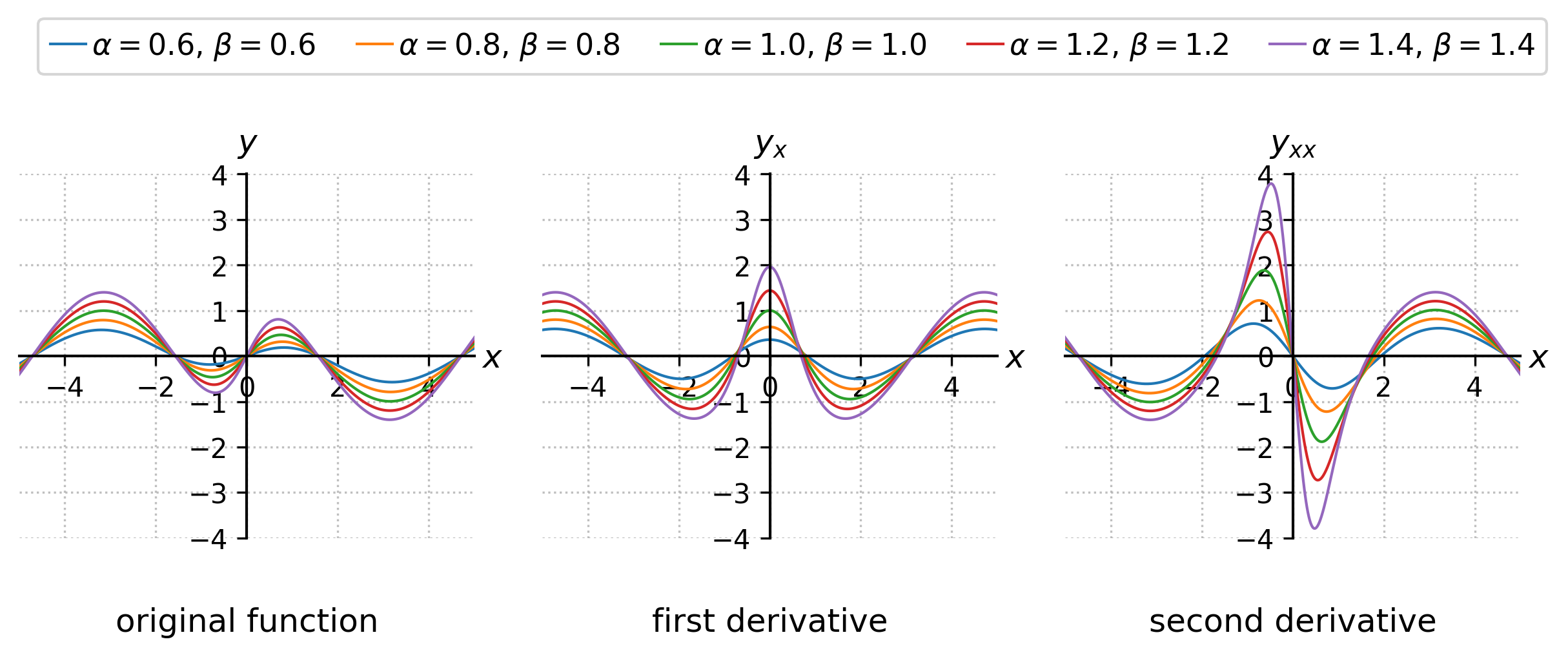}%
\label{fig:tanh_X_cos_L1L2}}
\hspace{0.05mm}
\subfloat[$\alpha \cdot$ cos($x$) $\cdot \beta \cdot$ atan($x$) $\cdot$ sigmoid($\gamma \cdot x$)]{\includegraphics[width=3.4in]{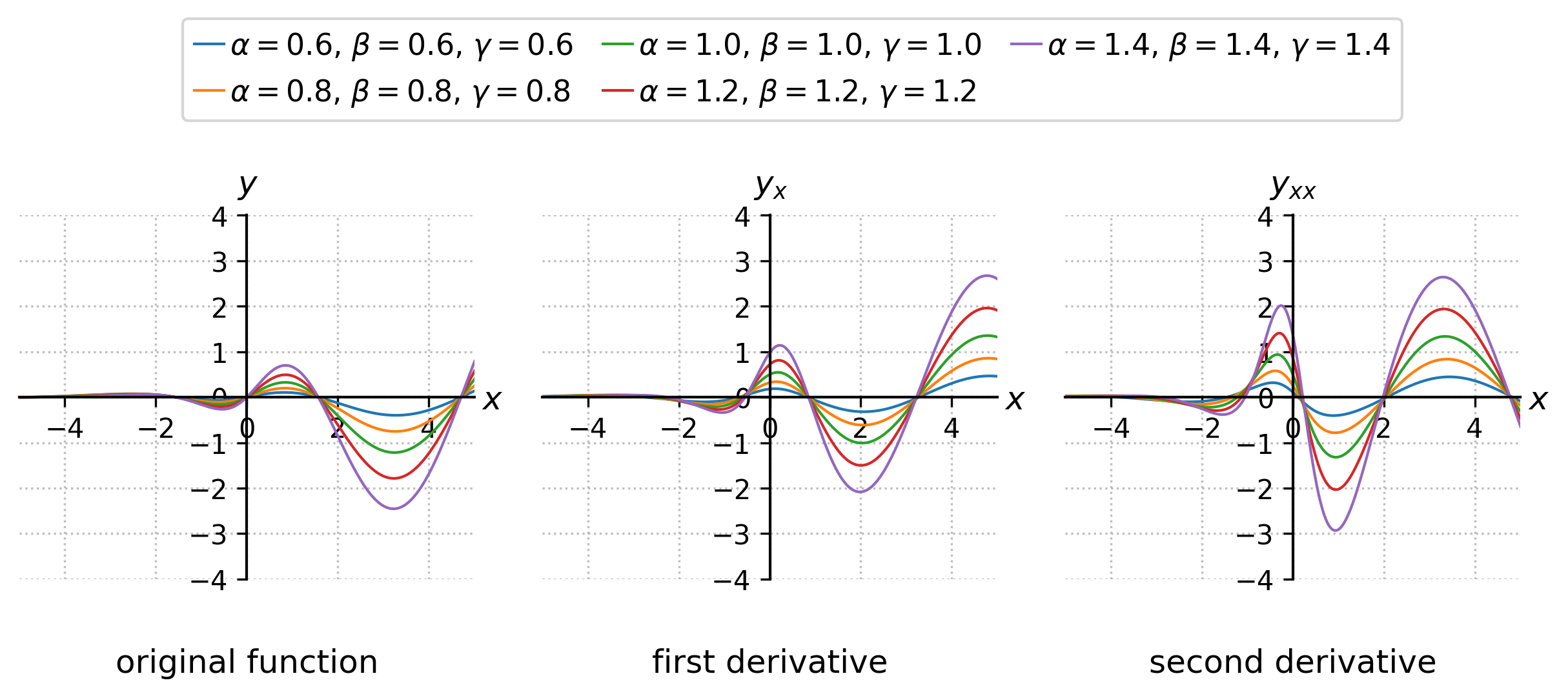}%
\label{fig:cos_X_atan_X_sigmoid_L0L4L7}}
\vspace{-0.3mm}
\subfloat[sin($\alpha \cdot x$) $\cdot$ sigmoid($x$)]{\includegraphics[width=3.4in]{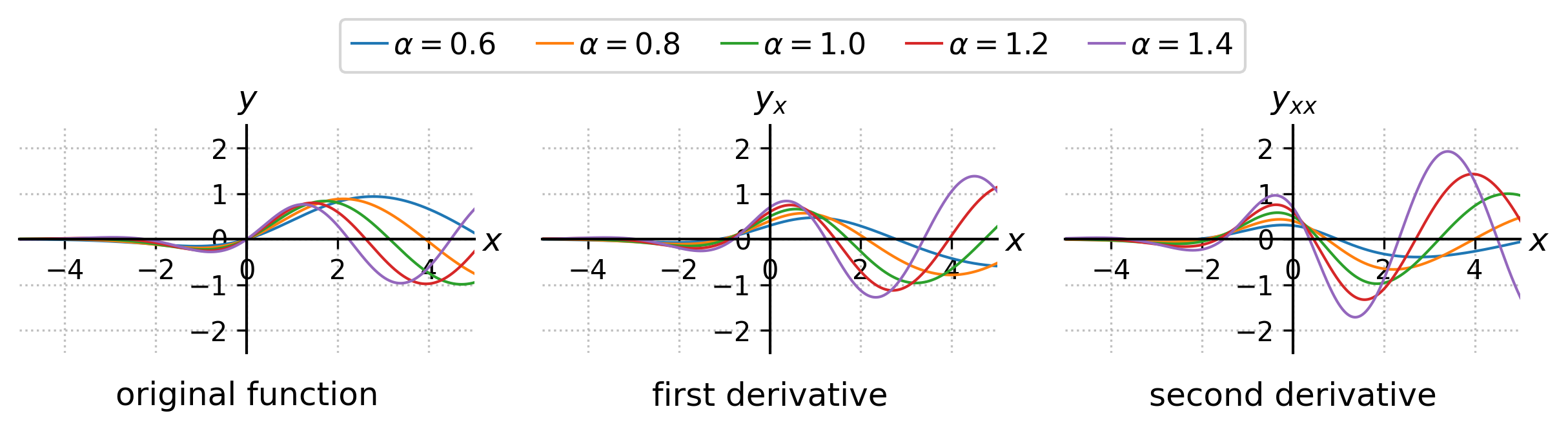}%
\label{fig:sin_X_sigmoid_L2}}
\hspace{0.05mm}
\subfloat[$\alpha \cdot$ sigmoid($\beta \cdot x$)]{\includegraphics[width=3.4in]{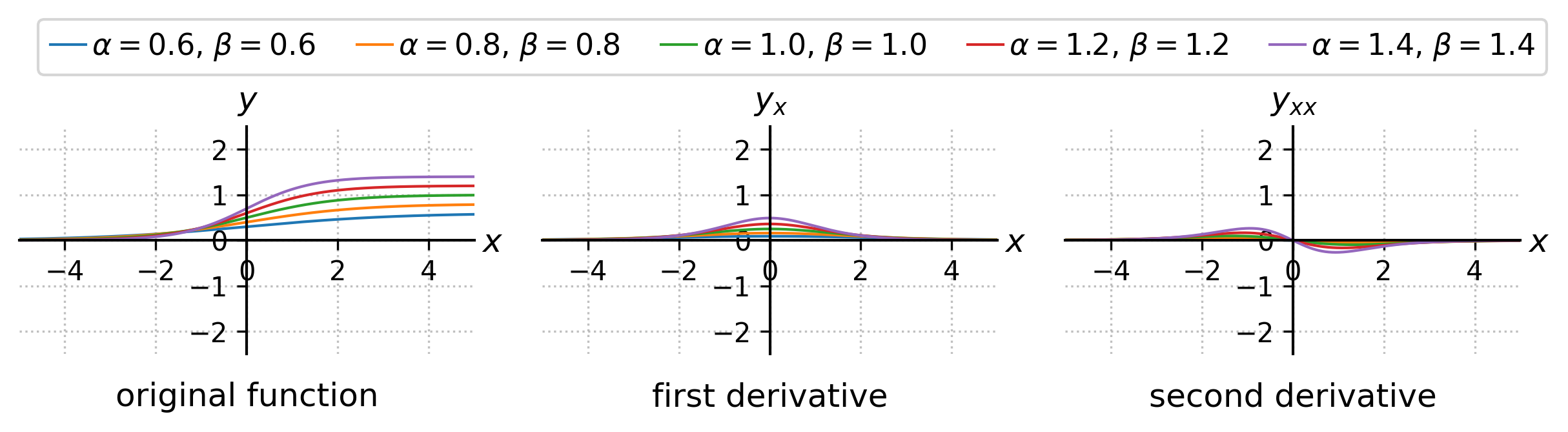}%
\label{fig:sigmoid_L0L1}}
\vspace{-0.3mm}
\subfloat[$\alpha \cdot$ asinh($\beta \cdot x $ $\cdot$ sigmoid($\beta \cdot x$))]{\includegraphics[width=3.4in]{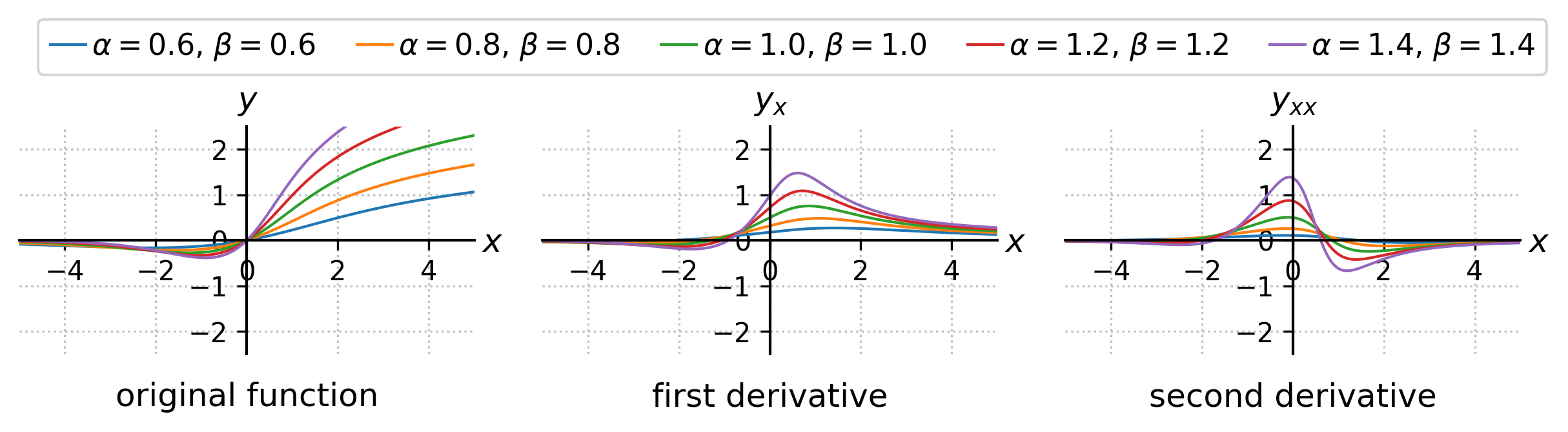}%
\label{fig:asinh_O_fixedswish_L0L2}}
\hspace{0.05mm}
\subfloat[$x$/($e^{x}+e^{-x}$)]{\includegraphics[width=3.4in]{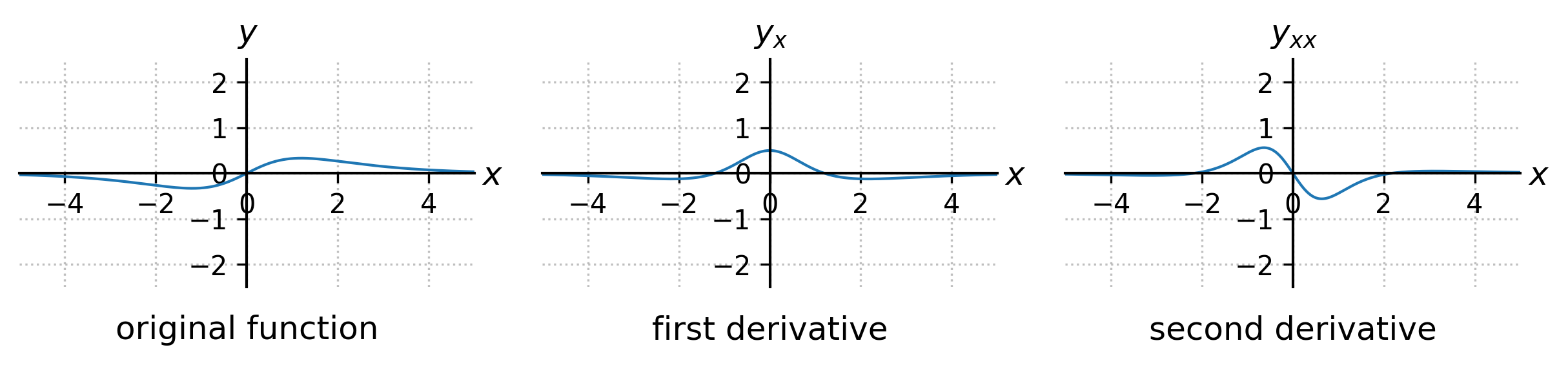}%
\label{fig:x_D_exp_p_expn}}
\vspace{-0.3mm}
\subfloat[sin(tanh($\alpha \cdot x$))]{\includegraphics[width=3.4in]{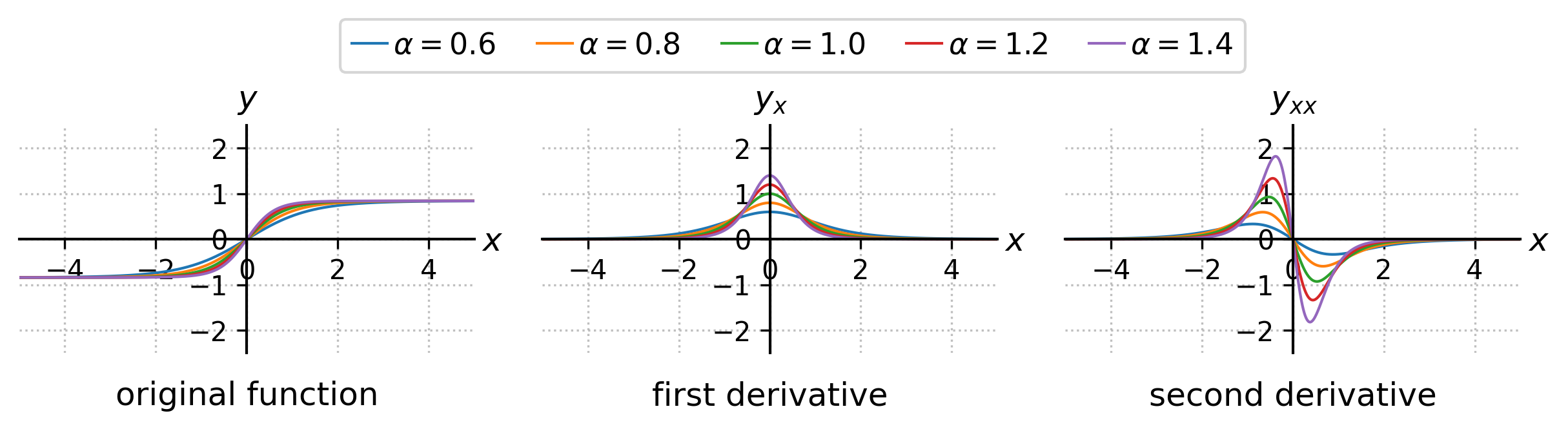}%
\label{fig:sin_O_tanh_L2}}
\hspace{0.05mm}
\subfloat[atan($\alpha \cdot$ $\beta \cdot x \cdot$ sigmoid($\beta \cdot x$))]{\includegraphics[width=3.4in]{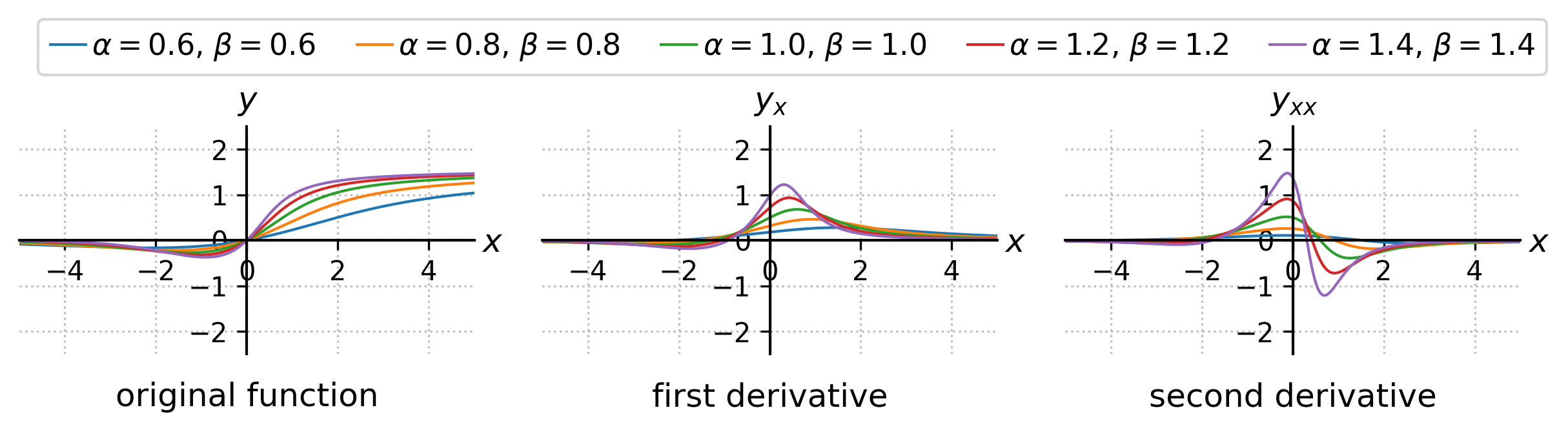}%
\label{fig:atan_O_fixedswish_L1L2}}
\caption{The common activation functions and the novel ones discovered by evo-w/-DPSTE.}
\label{fig:discovered_activation}
\vspace{0mm}
\end{figure*}

\section{Appendix: The details of each discovered model}
\label{sec:Appendix_discovered_model}
The relative $L_2$ error (mean and standard deviation), structure and activation function of each discovered model in the three experiments are displayed in Table \ref{tab:Klein-Gordon_appendix} (Klein-Gordon equation), Table \ref{tab:Burgers_appendix} (Burgers equation), and Table \ref{tab:Lame_appendix} (Lamé equations).
\vspace{0mm}

\section{Appendix: Discovered model structures}
\label{sec:Appendix_discovered_structure}
The model structures discovered though evo-w/-DPSTE in the three experiments are visualized in Fig.~\ref{fig:discovered_str}.
\vspace{0mm}

\section{Appendix: Discovered activation functions}
\label{sec:Appendix_discovered_activation}
The novel activation functions discovered by evo-w/-DPSTE in the three experiments and their first-order and second-order derivatives are illustrated in Fig.~\ref{fig:discovered_activation}.
\vspace{0mm}

%\FloatBarrier

\bibliographystyle{cas-model2-names}

\bibliography{cas-refs}

\bio{}
\endbio

\endbio

\end{document}